\documentclass[a4paper]{amsart}
\usepackage{bbm} 
\usepackage{graphicx}
\usepackage{subcaption}
\usepackage{amssymb}
\usepackage[utf8]{inputenc}
\usepackage{amsfonts}
\usepackage{amsmath}
\usepackage{amssymb}
\usepackage{amsthm}
\usepackage{url}
\usepackage[colorlinks,citecolor=blue]{hyperref}
\usepackage{mathabx}
\usepackage{wrapfig}
\usepackage{dsfont}
\usepackage{resizegather}
\usepackage{yfonts}
\usepackage{geometry}
\usepackage[dvipsnames]{xcolor}
\usepackage{algpseudocode}
\usepackage{algorithmicx}
\usepackage{algorithm}

\DeclareMathOperator*{\argmin}{arg\,min}

\providecommand{\U}[1]{\protect\rule{.1in}{.1in}}

\usepackage{xcolor}

\newtheorem{prop}{Proposition}[section]
\newtheorem{cor}[prop]{Corollary}
\newtheorem{defi}[prop]{Definition}

\newtheorem{rmk}[prop]{Remark}
\newtheorem{lem}[prop]{Lemma}

\newtheorem{theo}[prop]{Theorem}
\newtheorem{examp}[prop]{Example}

\newcommand{\tr}{\mbox{\rm Tr}}

\newcommand{\EE}{\mathbb{E}}

\newcommand{\PP}{\mathbb{P}}

\newcommand{\RR}{\mathbb{R}}
\newcommand{\SB}{\mathbb{S}}

\newcommand{\UU}{\mathbb{U}}
\newcommand{\VV}{\mathbb{V}}
\newcommand{\XX}{\mathbb{X}}
\newcommand{\WW}{\mathbb{W}}
\newcommand{\YY}{\mathbb{Y}}

\newcommand{\GG}{\mathbb{G}}

\newcommand{\Ba}{ {\mathcal B }}
\newcommand{\Ca}{ {\mathcal C }}

\newcommand{\La}{ {\mathcal L }}

\newcommand{\Ka}{ {\mathcal K }}

\newcommand{\Ea}{ {\mathcal E }}
\newcommand{\Sa}{ {\mathcal S }}
\newcommand{\Ra}{ {\mathcal R }}
\newcommand{\Va}{ {\mathcal V }}
\newcommand{\Ua}{ {\mathcal U }}

\newcommand{\Ga}{ {\mathcal G }}
\newcommand{\Qa}{ {\mathcal Q }}
\newcommand{\Oa}{ {\mathcal O }}

\newcommand{\Xa}{ {\mathcal X }}
\newcommand{\Ma}{ {\mathcal M }}

\newcommand{\Pa}{ {\mathcal P }}
\newcommand{\Za}{ {\mathcal Z }}
\newcommand{\Ya}{ {\mathcal Y }}

\newcommand{\Qb}{ {\bf Q}}

\newcommand{\Pb}{ {\bf P }}
\newcommand{\Cb}{ {\bf C }}
\newcommand{\Xb}{ {\bf X }}

\newcommand{\point}{\mbox{\LARGE .}}
\newcommand{\cqfd}{\hfill\blbx \\}
\def\blbx{\hbox{\vrule height 5pt width 5pt depth 0pt}\medskip}

\def \PP{\mathbb{P}}
\def \RR{\mathbb{R}}
\def \SS{\mathbb{S}}
\def \EE{\mathbb{E}}

\def \WW{\mathbb{W}}
\def \BB{\mathbb{B}}
\newcommand{\cchi}{\protect\raisebox{2pt}{$\chi$}}

\usepackage[textwidth=2.5cm, textsize=scriptsize]{todonotes}

\newcommand{\vertiii}[1]{{\left\vert\kern-0.25ex\left\vert\kern-0.25ex\left\vert #1
    \right\vert\kern-0.25ex\right\vert\kern-0.25ex\right\vert}}

\usepackage{amsopn}

\numberwithin{equation}{section}

\title[Gaussian entropic OT: Schr\"odinger bridges \& the Sinkhorn algorithm]{Gaussian entropic optimal transport: Schr\"odinger bridges and the Sinkhorn algorithm}

\author{O. Deniz Akyildiz}
\address{Department of Mathematics, Imperial College London, UK}
\email{\textcolor{blue}{\footnotesize \texttt{deniz.akyildiz@imperial.ac.uk}}}

\author{Pierre Del Moral}
\address{Centre de Recherche Inria Bordeaux Sud-Ouest, Talence, France}
\email{\textcolor{blue}{\footnotesize \texttt{pierre.del-moral@inria.fr}}}

\author{Joaqu\'in Miguez}
\address{Department of Signal Theory \& Communications, Universidad Carlos III de Madrid, Spain}
\email{\textcolor{blue}{\footnotesize \texttt{joaquin.miguez@uc3m.es}}}


\begin{document}

\maketitle
\begin{abstract} Entropic optimal transport problems are regularized versions of optimal transport problems. These models play an increasingly important role in machine learning and generative modelling. For finite spaces, these problems are commonly solved using Sinkhorn algorithm (a.k.a. iterative proportional fitting procedure). However, in more general settings the Sinkhorn iterations are based on nonlinear conditional/conjugate transformations and exact finite-dimensional solutions cannot be computed.

This article presents a finite-dimensional recursive formulation of the iterative proportional fitting procedure for general Gaussian multivariate models. As expected, this recursive formulation is closely related to the celebrated Kalman filter and related Riccati matrix difference equations, and it yields algorithms that can be implemented in practical settings without further approximations. We extend this filtering methodology to develop a refined and self-contained convergence analysis of Gaussian Sinkhorn algorithms, including closed form expressions of entropic transport maps and Schr\"odinger bridges.
\end{abstract}



\section{Introduction}

\subsection{Transport problems}

{The optimal transport problem consists in finding the most efficient way of transforming one given probability measure into another one selected as a target. To be specific,} let $\Ca(\eta,\mu)$ be the set of probability measures $\Pa(d(x,y))$ on the product space $(\RR^{d}\times\RR^d)$ for some $d\geq 1$, with prescribed first and {second} coordinate marginals $(\eta,\mu)$ {and} densities $(e^{-U},e^{-V})$ on $\RR^d$. {Additionally, let $Q(x,dy)=q(x,y)~dy$ be a Markov transition kernel} on $\RR^d$ with density $q(x,y)$ {with respect to (w.r.t.)} the Lebesgue measure $dy$ on $\RR^d$. The (regularized) entropic transport problem associated with these mathematical objects is formulated as \cite{cuturi,nutz}
\begin{equation}\label{def-entropy-pb}
\argmin_{\Pa\,\in\, \Ca(\eta,\mu)}\left(-\int~\log{q(x,y)}~\Pa(d(x,y))+\mbox{\rm Ent}(\Pa~|~\eta\otimes \mu)\right),
\end{equation}
{where} $\mbox{\rm Ent}(\Pa~|~\eta\otimes \mu)$ is the relative entropy of $\Pa$ w.r.t. the product measure $\eta\otimes \mu$ (we refer to Section~\ref{def-div-sec} for the description of the relative entropy).
In the optimal transport literature, the  function $c(x,y)=-\log{q(x,y)}$ is sometimes called the cost function. {As shown in Section~\ref{conditional-dist-sec} dedicated to conditioning principles, the function $c(x,y)$ can also be interpreted as a log-likelihood function.} The quadratic cost defined by
\begin{equation}\label{gibbs-cost}
c(x,y)={c_t(x,y):=}\frac{1}{2t}~\Vert x-y\Vert^2+\frac{d}{2}\log{(2\pi t)},
\end{equation}
for some given $t>0$ corresponds to Gaussian densities and the heat equation semigroup {(see for instance Remark~\ref{ref-general-models})}. The optimal transport problem corresponds to the case $t=0$. {Indeed, up to a rescaling, when $t=0$ solving (\ref{def-entropy-pb}) is equivalent to solving the optimal transport problem} 
$$
\argmin_{\Pa\,\in\, \Ca(\eta,\mu)}\int~~\Vert x-y\Vert^2~\Pa(d(x,y)).
$$ 
In this context, the parameter $t>0$ is seen as a regularization parameter. For a more thorough discussion on these entropically regularized optimal transport problems, we refer to the pioneering article by Cuturi~\cite{cuturi} (see also~\cite{genevay-cuturi,genevay-cuturi-2}).

 Choose $\Pa\,\in\, \Ca(\eta,\mu)$ of the form
 $$
 \Pa(d(x,y))=\eta(dx)~\Ka(x,dy)
 $$
 {and set
 $$
 \Pa_0(d(x,y)):=\eta(dx)~\Qa(x,dy),
 $$
where $\Ka(x,dy)$ and $\Qa(x,dy)$ are Markov kernels.} In this context, using the decomposition
 $$
 \frac{d\Pa}{d\Pa_0}= \frac{d\Pa}{d(\eta\otimes \mu)}~\frac{d(\eta\otimes \mu)}{d\Pa_0},$$
 we readily {obtain} the entropic cost formula
 $$
 \mbox{\rm Ent}(\Pa~|~\Pa_0)+ \int~\mu(dy)~V(y) =-\int~\log{q(x,y)}~\Pa(d(x,y))+\mbox{\rm Ent}(\Pa~|~\eta\otimes \mu).
 $$
In other words, the solution of the entropic transport problem (\ref{def-entropy-pb}) coincides with the solution of the (static) Schr\"odinger bridge from $\eta$ to $\mu$ w.r.t. the reference measure $\Pa_0$, {which is} defined by
\begin{equation}\label{def-entropy-pb-v2}
 \argmin_{\Pa\,\in\, \Ca(\eta,\mu)}\mbox{\rm Ent}(\Pa~|~\Pa_0).
\end{equation}
It is implicitly  assumed there exists some $\Pa\in \Ca(\eta,\mu)$ such that 
$\mbox{\rm Ent}(\Pa~|~\Pa_0)<\infty$.
This condition ensures  the existence of a Schr\" odinger bridge distribution $\Pa$
 that solves (\ref{def-entropy-pb-v2})  (cf. the seminal article by Csisz\'ar~\cite{csiszar-2}, as well as Section 6 in the Lecture Notes by Nutz~\cite{nutz}, see also the survey article by L\'eonard~\cite{leonard} and references therein).

Schr\" odinger bridges can rarely be solved analytically. {However, solutions can be approximated} efficiently using {the Sinkhorn algorithm, also referred to as the 
iterative proportional fitting procedure~\cite{cuturi,sinkhorn-2,sinkhorn-3}.} Let $\Ca_X(\eta)$ be the set of probability measures $\Pa(d(x,y))$ with {marginal $\Pa^X(dx)=\eta(dx)$ w.r.t. the $x$-coordinate and let $\Ca_Y(\eta)$ be the set of probability measures $\Pa(d(x,y))$ with marginal $\Pa^Y(dx)=\mu(dy)$ w.r.t. the $y$-coordinate.} In this notation, the Sinkhorn algorithm starts from $\Pa_0$ and solves sequentially the following entropic transport problems
\begin{equation}\label{sinhorn-entropy-form}
\Pa_{2n+1}:= \argmin_{\Pa\in \Ca_Y(\mu)}\mbox{\rm Ent}(\Pa~|~\Pa_{2n})\quad \mbox{\rm and}\quad
\Pa_{2(n+1)}:= \argmin_{\Pa\in  \Ca_X(\eta)}\mbox{\rm Ent}(\Pa~|~\Pa_{2n+1}).
\end{equation}
When $n\rightarrow\infty$, Sinkhorn bridges $\Pa_{n}$ converge towards the Schr\"odinger bridge  from $\eta$ to $\mu$.

In the dual formulation, these distributions are often written {as}
\begin{equation}\label{sinhorn-entropy-form-Sch}
\Pa_{n}(d(x,y))=e^{-U_{n}(x)}~q(x,y)~e^{-V_{n}(y)}~dxdy
\end{equation}
for a pair of Schr\" odinger potentials $(U_n,V_n)$ satisfying a system of integral relations starting from $(U_0,V_0)=(U,0)$ (see Proposition~\ref{prop-schp}).  The limiting 
Schr\" odinger {potentials $(\UU(x),\VV(x)):=\lim_{n\rightarrow\infty}(U_{n}(x),V_n(x))$ yield the bridge distribution $\Pa$ that solves problem (\ref{def-entropy-pb-v2}), i.e.,}
\begin{equation}\label{sinhorn-entropy-form-Sch-lim}
\lim_{n\to\infty} \Pa_{n}(d(x,y)) = \Pa(d(x,y))=e^{-\UU(x)}~q(x,y)~e^{-\VV(y)}~dxdy.
\end{equation}
 {While Sinkhorn iterations as presented in the recursions} \eqref{sinhorn-entropy-form} may look appealing {and easy to implement}, one should note that they are based on nonlinear conditional/conjugate transformations with generally no finite-dimensional recursive solutions and, therefore, they do not lead to a practical algorithm. In this paper, we present a self-contained analysis of Schr\" odinger bridges and a tractable Sinkhorn algorithm for a general class of Gaussian models. We provide closed form expression of  Schr\" odinger bridges  $(\UU,\VV)$ as well as the description of the bridge distribution $\Pa$ {that solves problem} \eqref{def-entropy-pb-v2} in terms of transport maps. We also provide a refined convergence analysis with sharp exponential convergence rates for entropic transport distributions $\Pa_n$ and the dual Schr\" odinger potentials $(U_n,V_n)$ {in expression  \eqref{sinhorn-entropy-form-Sch}.}

\subsection{Gaussian models}
{Let $\Sa^0_d$ be set of positive semi-definite matrices in $\RR^{d \times d}$, and let $\Sa^+_d\subset \Sa^0_d$ be the subset of positive definite matrices.}
 Denote by $\nu_{m,\sigma}$ the Gaussian distribution on $\RR^d$ with mean $m\in\RR^d$ and covariance matrix $\sigma \in \Sa^+_d$. In addition, let $g_{\sigma}$ denote the probability density function (pdf) of the distribution $\nu_{0,\sigma}$, with covariance matrix $\sigma \in \Sa^+_d$. {Hereafter, we study} general Gaussian models of the form
 \begin{eqnarray}
  (\eta,\mu)&=&(\nu_{m,\sigma},\nu_{\overline{m},\overline{\sigma}})\label{eq_gauss_pair_intro}\\
   q_{\theta}(x,y)&=&g_{\tau}(y-(\alpha+\beta x))
\quad\mbox{\rm with}\quad
\theta=(\alpha,\beta,\tau)\in \Theta:= \left(\RR^{d}\times\Ga l_d\times \Sa^+_{d}\right)\label{ref-gauss-mod-intro}
\end{eqnarray}
for some given $(m,\overline{m})\in(\RR^d\times\RR^{d})$ and $(\sigma,\overline{\sigma})\in (\Sa^+_d\times \Sa^+_{d})$, {where $\Ga l_d$ denotes the general linear group of $(d\times d)$-invertible matrices (hence $\Sa^+_d\subset \Ga l_d$).}

{The practical application of the Sinkhorn algorithm requires a finite-dimensional description of the flow of distributions $\Pa_n$ generated by the iteration of \eqref{sinhorn-entropy-form}.} As expected, for the Gaussian models in \eqref{ref-gauss-mod-intro}, the entropic transport problem (\ref{sinhorn-entropy-form}) is indeed solved by a finite-dimensional family of Gaussian conditional/conjugate distributions. 
For instance, {if $\Pa_{2n}:=\mbox{\rm Law}(X_n,Y_n)$ then we have $\Pa_{2n+1}(d(x,y))=\mu(dy)~\PP(X_n\in dx~|~Y_n=y)$ and the conditional distribution is Gaussian and can be calculated using least squares and linear regression methods {(see for instance the conditioning principles described in Section~\ref{conditional-dist-sec} and in Appendix~\ref{appendix-entro} on page~\pageref{appendix-entro}}).} The conditional mean and covariance updates associated with these models coincide with the traditional Kalman update {that arises} in discrete generation and linear-Gaussian filtering models, see for example Section 9.9.6 in~\cite{dp-17}. 

{Hence, one of the main goals of this paper is to apply this filtering methodology to solve Schr\"odinger bridges and analyze the convergence of the Sinkhorn algorithm for Gaussian models.} In the theory of Kalman filtering, the flow of covariance matrices associated with the Sinkhorn algorithm also satisfies offline matrix difference Riccati equations. The stability analysis and the stationary matrices associated with Riccati matrix flows are well understood, see for instance~\cite{dh-22} and references therein. In Appendix~\ref{app-ricc} we provide a brief discussion on Riccati matrix flows in the context of the Sinhorn algorithm, including the Floquet-type theory developed in~\cite{dh-22}, as well as several Lipschitz type inequalities and exponential type decays to equilibrium for Riccati flows and their associated exponential semigroups.

\subsection{{Motivation and related work}}

Optimal transport and its regularized entropic version {have become state-of-the-art tools} in a variety of application domains, {including} generative modeling and machine learning~\cite{arjovsky,doucet-bortoli,kolouri,peyre}, statistical barycenter problems~\cite{agueh,andersen,bigot,cramer-2,cuturi-w,loubes}, economy~\cite{bojitov}, computer vision~\cite{dominitz,solomon}, control theory~\cite{chen,chen-2}, and many others.

 {Finding and rigorously understanding closed-form solutions for multivariate Gaussian entropic optimal transport is of fundamental importance. Exact recursions for entropic optimal transport in the Gaussian case can serve as a baseline for testing approximate Sinkhorn algorithms on multivariate models, much like the Kalman filter’s role in testing approximate filtering algorithms. They can also form the basis for developing novel entropic optimal transport methods for non-Gaussian distributions using well-established nonlinear Kalman filtering ideas. Furthermore,} the problem of finding Gaussian distributions on product spaces with prescribed multivariate marginals and conditional constraints is a surprisingly difficult problem arising in graphical models~\cite{andersen,cramer,lauritzen}. Gaussian  Schr\" odinger bridges related entropic transport problems also arise in solving matching problems as well as in optimal control theory~\cite{bojitov,chen,chen-2}. {These articles provide and utilize closed form expressions for some specific classes of Gaussian Schr\" odinger bridges.}

{Given its central importance, the convergence of the Sinkhorn algorithm for Gaussian models {has been} discussed in prior works. The earliest works to discuss the convergence of Sinkhorn algorithm} are~\cite{cramer-2,cramer-3}. {However, these works do not present any closed form solutions or any explicit results on the convergence rates.} The more recent article~\cite{doucet-bortoli} also discusses quantitative exponential decays for  Gaussian centered models {(where $m=\overline{m}=0$ and $\alpha=0$)} and scalar-type matrices ($\beta=b~I$, $\tau=I$ and $\sigma=t I=\overline{\sigma}$) for some  real numbers $t\in ]0,\infty[$ and $b\in\RR$. 
The {type of models, with $m=\overline{m}=0$, $(\alpha,\beta)=(0,I)$ and scalar-type covariance $\tau=tI$,} is also {studied} in~\cite{janadi}. In this context, the authors present a closed form expression of limiting Schr\"odinger potential functions in terms of the fixed point of a Riccati-type equation. 

{When $(\alpha,\beta)=(0,I)$ and for scalar-type covariance $\tau=tI$, similar fixed-point equations are also} {investigated} in the series of recent articles~\cite{agueh,bojitov,bunne2023schrodinger,loubes,janadi,mallasto}. These articles  discuss Gaussian bridges and entropic interpolations of the form in {Eq. \eqref{gibbs-cost}}. They also discuss the effect of the regularization parameter but they do not seek any finite dimensional description of the iterations in the Sinkhorn algorithm or their convergence rate.

The regularity properties of the optimal transport map between Gaussian distributions can also be deduced from Caffarelli's contraction theorem~\cite{caffarelli} on the Lipschitz's regularity properties of the optimal transport map between Gaussian and strongly log-concave distributions. 

Most of the literature on Sinkhorn iterates is concerned with
finite state spaces~\cite{borwein,sinkhorn-2,sinkhorn-3,soules} as well as compact state spaces or bounded cost functions using Hilbert projective metrics techniques~\cite{chen-2016,deligiannidis,franklin-1989,marino}. It is out of the scope of this article to review all the contributions in this field --we simply refer to the recent book~\cite{peyre} and the references therein. 

There are very few articles on the convergence of Sinkhorn iterates on non-compact spaces and unbounded cost functions that apply to Gaussian models {with the notable exception of  
two recent significant contributions~\cite{chiarini,durmus}}.
{More precisely,} the exponential convergence of the Sinkhorn iterations in \eqref{sinhorn-entropy-form} for cost functions of the form in Eq. \eqref{gibbs-cost} can {be} deduced from the recent article~\cite{durmus}, {which investigate} quantitative contraction rates for target marginal distributions $(\eta,\mu)$ with an asymptotically positive log-concavity profile and cost functions of the form in \eqref{gibbs-cost} associated with a sufficiently large regularization parameter. {These exponential decays have been recently refined to  apply to all values of the regularization parameter in the more recent article~\cite{chiarini}. The entropy estimates presented in Proposition 1.3 of~\cite{chiarini} also apply directly  to Gaussian models of the form \eqref{ref-gauss-mod-intro} when the cost function is symmetric and the parameters are $(\alpha,\beta)=(0,I)$ and $\tau=t~\Sigma$, for some {symmetric positive-definite matrix $\Sigma$}. These exponential decays presented in~\cite{durmus,chiarini} are closely related but differ from to the ones based on Floquet-type representation of Riccati flows  discussed in the present article (see for instance Theorem~\ref{theo-qs} the estimate (\ref{ref-cc-24}) and Remark~\ref{comp-rmk}).}

Extensions of our results to log-concave models have been developed in \cite{dm-25}. The recent article~\cite{adm-25} also develops a semigroup contraction analysis based on Lyapunov techniques to prove the exponential convergence of Sinkhorn algorithm on weighted Banach spaces. These Lyapunov approaches also apply to multivariate linear Gaussian models  for sufficiently large regularization parameter as well as statistical finite mixture models including Gaussian-kernel density estimation of complex data distributions arising in generative models.

{In the same context,} the convergence of Sinkhorn iterations can also be deduced from Theorem 6.15 in~\cite{nutz} under an exponential integrability condition \cite[condition (6.8)]{nutz} which is again only met for a sufficiently large regularization parameter. To the best of our knowledge, the weakest regularity conditions {that ensure the convergence of Sinkhorn iterations} are presented in the recent articles~\cite{promit-2022,nutz-wiesel}. These are mild integrability conditions of the cost function w.r.t. the target marginal measures $(\eta,\mu)$, which apply to general Gaussian models and any choice of the regularization parameter (see Remark~\ref{rmk-int-nutz}). Nevertheless, the article~\cite{nutz-wiesel} does not provide convergence results in relative entropy but in total variation, without any {explicit} rates, and the article~\cite{promit-2022} presents sub-linear relative entropy rates.

\subsection{Main contributions}

The aim of this paper is to provide a self-contained and refined analysis of the Sinkhorn algorithm and Schr\"odinger bridges for general Gaussian multivariate models. {For given Gaussian measures $(\eta,\mu):=(\nu_{m,\sigma}, \nu_{\bar m,\bar \sigma})$ as in \eqref{eq_gauss_pair_intro}, the Sinkhorn iteration yields a sequence
$$
\overbrace{\nu_{m,\sigma}\Ka_0}^{\nu_{m_0,\sigma_0}} \longrightarrow  \underbrace{\nu_{\bar m,\bar \sigma}\Ka_1}_{\nu_{m_1,\sigma_1}} \longrightarrow \cdots \longrightarrow  \overbrace{\nu_{m,\sigma}\Ka_{2n}}^{\nu_{m_{2n},\sigma_{2n}}} \longrightarrow \underbrace{\nu_{\bar m,\bar \sigma}\Ka_{2n+1}}_{\nu_{m_{2n+1},\sigma_{2n+1}}} \longrightarrow \cdots,
$$
where each $\nu_{m_n,\sigma_n}$ is a Gaussian distribution with mean $m_n$ and covariance matrix $\sigma_n$, and $\Ka_n$ is a linear and Gaussian Markov kernel. The Sinkhorn algorithm converges in the sense that 
$$
\lim_{n\to\infty} \nu_{m_{2n},\sigma_{2n}} = \nu_{m,\sigma} 
\quad\text{and}\quad 
\lim_{n\to\infty} \nu_{m_{2n+1},\sigma_{2n+1}} = \nu_{\bar m,\bar\sigma},
$$ 
while the bridges $\Pa_{2n}(d(x,y)) = \nu_{m,\sigma}(dx) \Ka_n(x,dy)$ and $\Pa_{2n+1}(d(x,y)) = \nu_{\bar m,\bar\sigma}(dy) \Ka_{2n+1}(y,dx)$ are also Gaussian distributions that correspond to the iteration in \eqref{sinhorn-entropy-form} and satisfy 
$$
\lim_{n\to\infty} \Pa_{2n} = \lim_{n\to\infty} \Pa_{2n+1} = \Pa,
$$ 
where $\Pa$ is the optimal Schr\"odinger bridge that solves problem \eqref{def-entropy-pb-v2}. The Sinkhorn iteration can also be expressed in terms of a sequence of Schr\"odinger potentials $(U_n,V_n)$ that determine the bridges $\Pa_n$, as given by Eq. \eqref{sinhorn-entropy-form-Sch}. These potentials also converge,
$$
\lim_{n\to\infty} (U_n,V_n) = (\UU,\VV),
$$ 
where $\UU$ and $\VV$ are the optimal Schr\"odinger potentials that characterize the solution $\Pa$ of \eqref{def-entropy-pb-v2}, i.e., $\Pa(d(x,y))=e^{-\UU(x)}q(x,y)e^{-\VV(y)}dxdy$.}

{In this paper, we obtain closed-form expressions for the Schr\"odinger potentials $(U_n,V_n)$ and the Gaussian Sinkhorn iterates $\nu_{m_n,\sigma_n}$, as well as sharp (non asymptotic) convergence rates for the Sinkhorn algorithm. In particular:} 

\begin{itemize}
\item {We construct explicit closed-form expressions for the distribution flow $\Pa_n$, as well as the corresponding Schr\"odinger potentials $U_n$ and $V_n$, for general Gaussian models of the form in \eqref{eq_gauss_pair_intro}-\eqref{ref-gauss-mod-intro}. A sequential formulation of the distributions $\Pa_n$ generated by the Gaussian Sinkhorn algorithm is provided in Section~\ref{sec-sinkhorn}. Then, we provide a complete description of the mathematical objects $(\Pa_n,U_n,V_n)$ in terms of Riccati matrix difference equations in Section \ref{diff-ricc-sec} (see Theorem~\ref{th-2}). 
Closed-form expressions of the Schr\"odinger potentials $(U_n,V_n)$ are constructed in Section \ref{sec-schrod} (see Theorem~\ref{theo-end}).} 

\item {We analyze the convergence of the Gaussian Sinkhorn algorithm towards the corresponding Schr\"odinger bridges.}

    \begin{itemize}
    \item {Gaussian bridge transport maps and Schr\"odinger potential functions can be explicitly described in terms of the reference parameter $\theta = (\alpha, \beta, \tau) \in \left(\RR^{d}\times\Ga l_d\times \Sa^+_{d}\right)$ in \eqref{ref-gauss-mod-intro}. If we let $\theta_0:=\theta$ then the initial Markov kernel $\Ka_0(x,dy) = q_{\theta_0}(x,y)dy$ can be denoted as $\Ka_0 = K_{\theta_0}$. The Sinkhorn iteration can then be interpreted as generating a sequence of parameters
$$
\theta_0 \longrightarrow \theta_1 \longrightarrow \cdots \longrightarrow \theta_{2n} \longrightarrow \theta_{2n+1} \longrightarrow \cdots    
$$
which determine the Markov kernels $\Ka_n = K_{\theta_n}$, the distributions $\Pa_n = P_{\theta_n}$, and the sequence $\nu_{m_n,\sigma_n}$, where $m_n$ and $\sigma_n$ are computed from $\theta_n$. For given $\theta$, Theorem~\ref{Th1} provides an explicit expression for the Schr\"odinger bridge map from $\nu_{m,\sigma}$ to $\nu_{\bar m,\bar \sigma}$ in terms of an optimal parameter $\SS(\theta)=(\iota_\theta, \kappa_\theta, \varsigma_\theta) \in \left(\RR^{d}\times\Ga l_d\times \Sa^+_{d}\right)$ (with the dual map from $\nu_{\bar m,\bar \sigma}$ to $\nu_{m,\sigma}$ described in Corollary~\ref{cor-dual-bm}). An explicit Sinkhorn iteration in terms of the parameters $\theta_n$ is obtained, via Riccati difference equations, in Section \ref{sec-gauss-sinkhorn} (see also Section \ref{sec:alg:gaussian-sinkhorn} for a self-contained outline of the iteration).}

    \item {We prove that $\lim_{n\to\infty} \theta_{2n} = \SS(\theta) = (\iota_\theta, \kappa_\theta, \varsigma_\theta)$, and provide explicit contraction estimates in terms of the fixed-points of Riccati matrix difference equations, in Section \ref{quant-sec}}. As shown in~\cite{dh-22} these exponential contraction rates based on Floquet-type representation of Riccati flows are sharp --see Theorem 1.3 in ~\cite{dh-22} and Proposition~\ref{prop-cv-app} herein.\footnote{Closed form solutions of Riccati flows for one-dimensional models are also developed in Section 4.2 of~\cite{dh-23} (see also Remark~\ref{one-d-cf} herein).} Quantitative exponential stability estimates for the Gaussian Sinkhorn algorithm are obtained in Theorem~\ref{theo-qs} and Corollary~\ref{cor-theta-even-cv}. {In particular, we prove that $\theta_{2n} \to \SS(\theta)$ exponentially fast, with contraction coefficients that are obtained explicitly from the Riccati equations. For example, Corollary~\ref{cor-theta-even-cv} states that 
$$
\Vert\theta_{2n}-\SS(\theta)\Vert\leq 
c~\rho_{\theta}^n~
\Vert \tau_{0}-\varsigma_{\theta}\Vert+\
c'~\overline{\rho}_{\theta_1}^{n/2}~\Vert m_0-\overline{m} \Vert,
$$
for some contraction coefficients $0<\rho_\theta, \bar \rho_{\theta_1} < 1$ and finite constants $c,c'<\infty$.} Relative entropy, total variation and Wasserstein distance non-asymptotic estimates are also given by Corollary~\ref{cor-entrop-cv} and Corollary~\ref{cor-was-sinkhorn}.

    \item In Section~\ref{gibbs-loop-sec} we analyze the stability properties of a class of Gibbs loop-type time-varying Markov chains associated with the Sinkhorn iterations for general (non-necessarily Gaussian) models. We present a rather elementary way to derive sub-linear rates. Sharp exponential convergence rates for Gaussian models are then presented in Corollary~\ref{theo-cor-qs}. 

    \end{itemize}

\item {Recall the reference parameter $\theta=(\alpha,\beta,\tau)$ that, in turn, defines the reference distribution $\Pa_0$ in the static Schr\"odinger bridge problem \eqref{def-entropy-pb-v2}.} For the class of Gaussian models \eqref{ref-gauss-mod-intro} where the covariance parameter has the form $\tau=tI$, {and we denote $\theta(t)=(\alpha,\beta,tI)$,} we carry out a refined analysis of the effects of the regularization parameter $t>0$.

    \begin{itemize}

    \item Convergence rates for the bridge transport maps, {$\SS(\theta(t))$, and Schr\"odinger potentials, $\VV_{\theta(t)}$ and $\UU_{\theta(t)}$}, towards independent Gaussians as $t\rightarrow\infty$ are presented, respectively, in Corollary~\ref{cor-monge-maps-ind} and Proposition~\ref{est-ct-m-intro}. The effect of this regularization on the Sinkhorn algorithm and its exponential convergence rates is also discussed in Section~\ref{sinkhorn-reg-effects}.

    \item {Convergence rates for the Gaussian bridge transport maps and Schr\"odinger bridge measures  towards Monge maps as $t\rightarrow 0$ are presented in Corollary~\ref{cor-monge-maps}. Quantitative bounds on the rate of convergence of regularized optimal transport costs to standard Wasserstein optimal transport are presented in Theorem~\ref{theo-entr-wass}.}

    \end{itemize}

\end{itemize}

\subsection{Outline of the paper}

We provide background material in Section \ref{sBackground}. Gaussian Schr\"odinger bridges and entropic transport maps, including regularized models, are analyzed in Section \ref{sec-bridges}. Section \ref{sec-gauss-sinkhorn} is devoted to the Sinkhorn scheme, including the closed-form, finite-dimensional Gaussian Sinkhorn algorithm and the Gibbs loop-type heterogeneous Markov chains associated to general Sinkhorn iterations. In Section \ref{quant-sec} we provide quantitative estimates for the iterates of the Gaussian Sinkhorn algorithm and Section \ref{sec-schrod} is devoted to the analysis of the convergence of the Schr\"odinger potentials along the Gaussian Sinkhorn iterations. {In Section \ref{sec:alg:gaussian-sinkhorn} we provide a pseudocode of the Gaussian Sinkhorn iterations, for fast reference. Then, we provide a numerical illustration of their exponentially-fast convergence towards the optimal Schr\"odinger bridge.} Finally, Section \ref{sec:Conclusions} contains some concluding remarks and a discussion of the main results obtained in this paper. Most of the proofs, as well as extended analyses and numerics, are provided in Appendices \ref{app-ricc} through \ref{sec-tech-proofs}.

\section{Background} \label{sBackground}

\subsection{Integral operators}\label{def-div-sec}

Let $\XX$ be a Banach space equipped with some norm $\Vert x\Vert_{\XX}$. Also, let $\Ma(\XX)$ be the set of nonnegative bounded measures on $\XX$ and let $\Ma_1(\XX)\subset \Ma(\XX)$ denote the convex subset of probability measures. Let $\Ba(\XX)$ be the set of bounded and measurable functions $f$ on $\XX$ equipped with the uniform norm $\Vert f\Vert:=\sup_{x\in \XX}|f(x)|$. 

\paragraph{Divergences between probability measures} We denote by $$
\eta(f):=\int_{\XX} f(x)~\eta(dx),
$$ the Lebesgue integral of a integrable function
$f\in \Ba(\XX)$ w.r.t. some $\eta\in \Ma(\XX)$. {We make use of several notions of divergence between pairs of probability measures:}
\begin{itemize}
\item The total variation distance on $\Ma_1(\XX)$ is defined for any $\eta_1,\eta_2\in \Ma_1(\XX)$ by
\begin{equation}
\Vert \eta_1-\eta_2\Vert_{\text{tv}}:=\sup\left\{(\eta_1-\eta_2)(f)~:~f\in \Ba(\XX)~\mbox{\rm s.t.}~~ \mbox{\rm osc}(f)\leq 1\right\}.
\label{eqdeftv}
\end{equation}
In \eqref{eqdeftv}, $\mbox{\rm osc}(f)$ stands for the oscillations of the function $f$, defined as
 $$
  \mbox{\rm osc}(f):=\sup_{(x_1,x_2)\in \XX^2}\vert f(x_1)-f(x_2)\vert.
 $$
 \item {The relative entropy (a.k.a. Kullback–Leibler divergence) between to measures $\eta_1\ll\eta_2$ is defined by
$$
\mbox{\rm Ent}\left(\eta_1~|~\eta_2\right)=\int~\log{\left(\frac{d\eta_1}{d\eta_2}(x)\right)}~\eta_1(dx).
$$ 
Notation $\eta_1\ll\eta_2$ indicates that $\eta_1\in \Ma(\XX)$ is absolutely continuous w.r.t. $\eta_2\in \Ma(\XX)$, i.e., $\eta_2(A)=0$ implies that $\eta_1(A)$ for any measurable subset $A\subset \XX$. {We also write $\eta_1\simeq\eta_2$ when the measures are equivalent in the sense that $\eta_1\ll\eta_2\ll\eta_1$. When $\eta_1\not\ll\eta_2$, we set $\mbox{\rm Ent}\left(\eta_1~|~\eta_2\right)=\infty$.}}
\item The $p$-th Wasserstein distance {between $\eta_1$ and $\eta_2$ is given by}
   $$
   \WW_{p}(\eta_1,\eta_2):= \inf_{\pi \in \Ca(\eta_1,\eta_2)}~\left(\int~\Vert x_1-x_2\Vert_{\XX}^{p}~\pi(d(x_1,x_2))\right)^{1/{p}}
   \quad \text{{for $p\ge 1$,}}
   $$ 
where $\Ca(\eta_1,\eta_2)$ stands for the convex subset of probability measures $\pi\in \Ma_1(\XX^2)$ with marginal $\eta_1$ w.r.t. the first coordinate and marginal $\eta_2$ w.r.t. the second coordinate. 
\end{itemize}

\paragraph{Markov transition kernels} Given a probability measure $\Pa\in\Ma_1(\XX^n)$ for some $n\geq 1$ we denote by $\Pa^{\flat}$ the probability measure defined by reversing the coordinate order, that is
$$
\Pa^{\flat}(d(x_1,x_2,\ldots,x_n)):=\Pa(d(x_n,x_{n-1},\ldots,x_1)).
$$
{In particular, for a} Markov transition $\Ka(x,dy)$ from $\XX$ into itself and
a measure $\mu \in\Ma(\XX)$ {we see that}
$$
(\mu\times \Ka)(d(x,y)):=\mu(dx) \Ka(x,dy)
$$
{implies}
$$
(\mu\times \Ka)^{\flat}(d(x,y))=\mu(dy) \Ka(y,dx). 
$$
For any pair of Markov transitions $\Ka_1,\Ka_2$ from $\XX$ {into itself} we may also write
$$
(\mu\times \Ka_1\times \Ka_2)(d(x_0,x_1,x_2))=\mu(dx_0)\Ka_1(x_0,dx_1)
\Ka_2(x_1,dx_2)
$$
{We also denote by $(\Ka_1\Ka_2)$ the Markov transition defined by the integral composition
$$
(\Ka_1\Ka_2)(x_0,dx_2):=\int\Ka_1(x_0,dx_1)
\Ka_2(x_1,dx_2)
$$}
{Given a} function $f \in\Ba(\XX)$, {any measure $\mu \in\Ma(\XX)$ and any bounded integral positive operator $\Ka(x,dy)$}. We denote by $\mu \Ka\in \Ma(\XX)$ and $\Ka(f)\in \Ba(\XX)$ the measure and the function defined by  
$$
(\mu \Ka)(dy):=\int~\mu(dx) \Ka(x,dy)\quad \mbox{\rm and}\quad \Ka(f)(x):=\int \Ka(x,dy) f(y),
$$
{respectively.}

\paragraph{Transport maps}
For a given $\pi\in \Ma_1(\XX)$ and a transport map 
$$
\begin{array}{lccc}
T: &\XX &\mapsto &\XX\\
&x &\leadsto &T(x)\\
\end{array}
$$ 
we denote by $T\star\pi$ the push forward of $\pi$ by $T$. {Specifically,} for any $f\in\Ba(\XX)$ we have
$$
(T\star\pi)(f):=
(\pi\circ T^{-1})(f):=\pi(f\circ T).
$$
For a given $\Pa\in\Ma_1(\XX^2)$, we denote by $\Pa^X\in\Ma_1(\XX)$ and $\Pa^Y \in\Ma_1(\YY)$ {the marginal probability measures}
 $$
\Pa^X(dx):=\int\Pa(d(x,y))\quad\mbox{\rm and}\quad
\Pa^Y(dy):=\int\Pa(d(x,y)),
$$
{respectively.}

\subsection{Matrix spaces and Riccati maps}

We denote by $\lambda_{\text{\rm min}}(v)$ and $\lambda_{\text{max}}(v)$ the minimal and the maximal eigenvalues, respectively, of a symmetric matrix $v\in\RR^{d\times d}$ for some $d\geq 1$.
The Frobenius matrix norm of a given matrix $v$ is defined by
$\left\Vert v\right\Vert_{F}^2=\tr(v^{\prime}v)$, 
with the trace operator $\tr(\cdot)$ and $v^{\prime}$ the transpose
of the matrix $v$. The spectral norm is defined by $\Vert v\Vert_2=\sqrt{\lambda_{\text{max}}(v^{\prime}v)}$. We sometimes use the L\" owner partial ordering notation $v_1\succeq v_2$ to mean that a symmetric matrix $v_1-v_2$ is positive semi-definite (equivalently, $v_2 - v_1$ is negative semi-definite), and $v_1\succ v_2$ when $v_1-v_2$ is positive definite (equivalently, $v_2 - v_1$ is negative definite). Given $v\in\Sa_d^+$ we denote by $v^{1/2}$  the principal (unique) symmetric square root.

 {For any $u,v\in\Sa^+_d$,} the Bures-Wasserstein distance \cite{bathia-2} on $\Sa^+_d$ is given by
\begin{equation}\label{bw-def}
 D_{bw}(u,v)^2:=\tr(u)+\tr(v)-\tr\left(\left( v^{1/2}~u~v^{1/2}\right)^{1/2}\right)
\end{equation}
{and} the geometric mean $u~\sharp~ v$ of {two} positive definite matrices $u,v\in\Sa^+_d$ is defined by
\begin{equation}\label{sym-sharp}
u~\sharp~v=v~\sharp~u:=v^{1/2}~ \left(v^{-1/2}~u~v^{-1/2}\right)^{1/2}~v^{1/2}.
\end{equation}
For completeness, a proof of the symmetric property is provided in Appendix \ref{sec-tech-proofs} (on page~\pageref{sym-sharp-proof}). The geometric symmetric mean is the unique solution of the Riccati equation
$$
(u~\sharp~v)~u^{-1}~(u~\sharp~v)=v, \quad \text{or, equivalently,} \quad (v~\sharp~u)~v^{-1}~(v~\sharp~u)=u.
$$

For any conformal matrices $(u,v)$, a direct application of Cauchy-Schwarz inequality yields
\begin{equation}\label{form-pq-ref}
\vert\tr(uv)\vert\leq \Vert u\Vert_F~\Vert v\Vert_F\quad \mbox{\rm and}\quad
\Vert uv\Vert_F=\sqrt{\tr(u vv^{\prime}u^{\prime})}\leq \Vert u\Vert_2~
\Vert v\Vert_F
\end{equation}
{We also recall} the norm equivalence
$$
\Vert u\Vert_2^2\leq \Vert u\Vert_F^2\leq d~\Vert u\Vert_2^2,
$$
{that holds for any square $(d \times d)$ matrix $u$. Moreover,} for any $u,v\in\Sa^0_d$ we have
\begin{equation}\label{f30}
 \tr\left(u^2\right)\leq \tr\left(u\right)^2\leq d~\tr\left(u^2\right)
\quad\mbox{\rm
and}\quad
\lambda_{\text{\rm min}}(u)~\tr\left(v\right)\leq \tr\left(uv\right)\leq \lambda_{\text{max}}(u)~\tr\left(v\right).
\end{equation}
{We note that \eqref{f30}} is also valid when $v$ is positive semi-definite and $u$ is symmetric. {This can be verified} using an orthogonal diagonalization of $u$ and recalling that $v$ remains positive semi-definite (thus with non negative diagonal entries).

We also quote the following estimate taken from~\cite{dt-18}
\begin{equation}\label{lem-tech-2}
\Vert u \Vert_F< \frac{1}{2}\Longrightarrow
\left\vert\log{\mbox{\rm det}\left(I-u\right)}\right\vert 
\leq \frac{3}{2}~\Vert u \Vert_F.~
\end{equation}
For any $u,v\in\Sa_{d}^+$ we have the Ando-Hemmen inequality
\begin{equation}\label{square-root-key-estimate}
\Vert u^{1/2}-v^{1/2}\Vert \leq \left[\lambda^{1/2}_{\rm min}(u)+\lambda^{1/2}_{\rm min}(v)\right]^{-1}~\Vert u-v\Vert
\end{equation}
that holds for any unitary invariant matrix norm $\Vert\cdot \Vert$, including the spectral and the Frobenius norms --see for instance Theorem 6.2 on page 135 in~\cite{higham}, as well as Proposition 3.2  in~\cite{hemmen}.

With a slight abuse of notation, we denote by $I$ the $(d\times d)$-identity matrix and by $0$ the null $(d\times d)$-matrix and the null $d$-dimensional vector, for any choice of the dimension $d\geq 1$. {We usually represent points $x\in\RR^d$ by $d$-dimensional column vectors and $1 \times d$ matrices.} In this notation, the Frobenius norm $\Vert x\Vert_F=\sqrt{x^{\prime}x}$ coincides with the Euclidean norm {and} we denote by $\WW_{p}$ the $p$-th Wasserstein distance on $\Ma_1(\RR^d)$ associated with the Euclidean norm. When there is no possible confusion,
 we use the notation $\Vert\cdot\Vert$ for any equivalent matrix or vector norm. 
 
 For any given  $m_1,m_2\in\RR^d$ and $\sigma_1,\sigma_2 \in \Sa^+_d$, we have
  \begin{equation}
\WW_2(\nu_{m_1,\sigma_1},\nu_{m_2,\sigma_2})^2=D_{bw}(\sigma_1,\sigma_2)^2+\Vert m_1-m_2\Vert_F^2. 
 \end{equation} 
 Also recall that the relative entropy of $\nu_{m_1,\sigma_1}$ w.r.t. $\nu_{m_2,\sigma_2}$ is given by the formula
 {
 \begin{equation}\label{KL-def}
\mbox{\rm Ent}\left(\nu_{m_1,\sigma_1}~|~\nu_{m_2,\sigma_2}\right)
 = \frac{1}{2}\left(
  D(\sigma_1~|~\sigma_2)+\Vert \sigma_2^{-1/2}\left(m_1-m_2\right)\Vert^2_F
\right)
 \end{equation}
 }
with the Burg (a.k.a. log-det) divergence
\begin{equation}\label{burg-def}
 D(\sigma_1~|~\sigma_2):=\tr\left(\sigma_1\sigma_2^{-1}-I\right)-\log{\mbox{det}\left(\sigma_1\sigma_2^{-1}\right)}.
\end{equation}

 We associate with some given $\varpi\in\Sa^+_d$  the increasing map $\mbox{\rm Ricc}_{\varpi}$ from $\Sa^0_d$ into  {$\Sa^+_d$} defined by
\begin{equation}\label{ricc-maps-def}
\begin{array}{lccl}
\mbox{\rm Ricc}_{\varpi}: &\Sa^0_d &\mapsto &\Sa^+_d\\
&v &\leadsto &\mbox{\rm Ricc}_{\varpi}(v):=(I+(\varpi+v)^{-1})^{-1}\\ 
\end{array}
\end{equation}
A refined stability analysis of Riccati matrix differences $v_{n+1}:=\mbox{\rm Ricc}_{\varpi}(v_n)$ and the limiting stationary matrices $r=\mbox{\rm Ricc}_{\varpi}(r)$ associated with these maps is provided in {Appendix \ref{app-ricc} (on page~\pageref{app-ricc})}.

These matrix equations belong to the class of discrete algebraic Riccati equations (DARE), {and} no analytical solutions are available for general models. We present a {novel} simple closed-form solution in terms of the matrix $\varpi$. As shown in Proposition~\ref{fix-p} (see also (\ref{ricc-monotone})) the unique positive definite fixed point {of the Riccati differences} is given by 
\begin{equation}\label{def-fix-ricc-1}
(I+\varpi^{-1})^{-1}\preceq r:=-\frac{\varpi}{2}+\left(\varpi+\left(\frac{\varpi}{2}\right)^2\right)^{1/2}\preceq I.
\end{equation}
 In addition, applying Proposition~\ref{prop-cv-app} there exists some constant $c_{\varpi}$
such that
\begin{equation}\label{cv-ricc-intro}
 \Vert v_{n}-r\Vert_2\leq c_{\varpi}~(1+\lambda_{\text{\rm min}}(\varpi+r))^{-2n}~
\Vert v_0-r\Vert_2
\end{equation}
{The contraction rates in \eqref{cv-ricc-intro} are based on Floquet-type representation of Riccati flows and they are sharp} (see Theorem 1.3 in ~\cite{dh-22} as well as {Remark~\ref{one-d-cf} and Proposition~\ref{prop-cv-app}} in the present article). For further discussion see Appendix \ref{app-ricc} (on page~\pageref{app-ricc}), the article~\cite{dh-22}, and the references therein.

\subsection{Conjugate Gaussian principles} \label{ss-conjugate-Gauss}

We associate with some $\theta=(\alpha,\beta,\tau)\in \Theta$ the Markov transition $K_{\theta}$ from $\RR^d$ into itself defined by
\begin{equation}\label{random-Z-map}
K_{\theta}(x,dy):=
\PP(Z_{\theta}(x)\in dy)
\quad\mbox{\rm and}\quad
Z_{\theta}(x):=\alpha+\beta x+\tau^{1/2}~G\in \RR^{d},
\end{equation}
{where $\alpha \in \RR^d$, $\beta \in \RR^{d \times d}$, $\tau \in \Sa_d^+$ and} $G$ stands for a $d$-dimensional centered Gaussian random variable (r.v.) with unit covariance. 

{Hereafter, let us assume that $(m,\overline{m})\in(\RR^d\times\RR^{d})$ and $(\sigma,\overline{\sigma})\in (\Sa^+_d\times \Sa^+_{d})$ are given fixed parameters. For a given parameter set $\theta\in \Theta$ and Gaussian measures $\nu_{m,\sigma}$ and $\nu_{\bar m,\bar \sigma}$, we define the probability measures}
 \begin{equation}\label{def-P-theta}
 P_{\theta}:= \nu_{m,\sigma}\times K_{\theta}  \quad\mbox{\rm and}\quad
  \overline{P}_{\theta}:= \nu_{\overline{m},\overline{\sigma}}\times K_{\theta}
 \end{equation}
and observe that
$$
  \nu_{m,\sigma}K_{\theta}=\nu_{h_{m,\sigma}(\theta)}
\quad\mbox{\rm with}\quad
h_{m,\sigma}(\theta)=(a_m(\theta),b_{\sigma}(\theta)):=\left(\alpha+\beta~m,\beta~\sigma~\beta^{\prime}+\tau\right).
$$

\begin{defi} \label{def-BB}
{To each pair of fixed parameters $(m,\sigma)\in(\RR^d\times \Sa^+_{d})$ we associate the map 
$$
\begin{array}{llcl}
\BB_{m,\sigma} : &\Theta &\mapsto &\Theta\\
&\theta=(\alpha,\beta,\tau) &\leadsto &\BB_{m,\sigma}(\theta)=(\iota,\kappa,\varsigma),\\
\end{array}
$$
where
\begin{equation}\label{ref-conjug-Z-par}
\kappa:=\sigma~\beta^{\prime}~b_{\sigma}(\theta)^{-1}, \quad
\iota = m-\kappa~a_m(\theta), \quad 
\mbox{and}\quad 
\varsigma^{-1}:=\sigma^{-1}+ \beta^{\prime}~\tau^{-1}\beta.
\end{equation}
}
\end{defi}

\begin{lem}
{The conjugate formula 
\begin{equation}\label{ref-conjug}
\left(\nu_{h_{m,\sigma}(\theta)}\times~K_{\,\BB_{m,\sigma}(\theta)}\right)^{\flat}=\nu_{m,\sigma}\times K_{\theta}
\end{equation}
holds for any parameter set $\theta\in \Theta$.}
\end{lem}

{The proof of \eqref{ref-conjug} follows readily from the construction of the maps $h_{m,\sigma}$ and $\BB_{m,\sigma}$.} In statistical theory the transformation (\ref{ref-conjug}) coincides with the  
Bayes updates of Gaussian distributions {--hence, we use} the terminology Bayes maps to refer to these transformations.

The (random) transport map (\ref{random-Z-map}) associated with $\BB_{m,\sigma}(\theta)$ has the form
\begin{equation}\label{ref-conjug-Z}
Z_{\BB_{m,\sigma}(\theta)}(x)=m+\kappa~(x-a_m(\theta))+\varsigma^{1/2}~G
\end{equation}
and, using the matrix inversion lemma, we can readily verify that
$$
\varsigma=\sigma-\kappa\beta\sigma=\sigma-\sigma\beta^{\prime}~b_{\sigma}(\theta)^{-1}~\beta\sigma
=\sigma-\kappa~b_{\sigma}(\theta)~\kappa^{\prime}
$$
or, equivalently,
$$
\kappa~b_{\sigma}(\theta)~\kappa^{\prime}+\varsigma=\sigma.
$$

{\begin{rmk}\label{ref-general-models}
We underline that the Gaussian transition (\ref{random-Z-map}) encapsulates all continuous time Gaussian models used in machine learning applications of Schr\"odinger bridges. Following~\cite{bishop2019stability} (see also~\cite{bunne2023schrodinger}),
let $\Ea_{s,t}(A)$ be the exponential semi-group (or the state transition matrix) associated with a smooth flow of matrices $A:t\in\RR_+\mapsto A_t\in \RR^{d\times d}$ defined for any $s\leq t$ by the forward and backward differential equations
\begin{equation*}
 \partial_t \,\Ea_{s,t}(A)=A_t\,\Ea_{s,t}(A)\quad\mbox{\rm and}\quad
\partial_s\, \Ea_{s,t}(A)=-\Ea_{s,t}(A)\,A_s,
\end{equation*}
respectively, where $\Ea_{s,s}(A)=\mathrm{Id}$. Equivalently in terms of the matrices
$\Ea_t(A):=\Ea_{0,t}(A)$ we have
$
\Ea_{s,t}(A)=\Ea_t(A)\Ea_s(A)^{-1}
$.  
We let $\Xa_t(x)$ be the linear diffusion process starting at $\Xa_0(x)=x$ defined by
\begin{equation}\label{stochastic-OU}
d\Xa_t(x)=\left(A_t~\Xa_t(x)+b_t\right)~dt+\Sigma_t^{1/2}~dW_t,
\end{equation}
where $W_t$ a $d$-dimensional Wiener process, $b:t\in\RR_+\mapsto b_t\in \RR^d$ and $\Sigma:t\in\RR_+\mapsto \Sigma_t\in \Sa^+_d$ a flow of positive definite matrices.
{Observe that the solution of \eqref{stochastic-OU} at some final time horizon $t$ is provided by the formula
\begin{equation}\label{solution-stochastic-OU}
\Xa_t(x)\stackrel{\text{law}}{=}\alpha[t]+\beta[t]~x+\tau[t]^{1/2}~ G
\end{equation}
with the parameters
$$
\alpha[t]:=
\int_0^t~\Ea_{s,t}(A)~b_s~ds,
\qquad\beta[t]:=\Ea_{t}(A)
\quad \mbox{\rm and}\quad
\tau[t]:=
\int_0^t~\Ea_{s,t}(A)~\Sigma_s~\Ea_{s,t}(A)^{\prime}~ds.
$$
}
{There are some relevant special cases:}
{
\begin{itemize}
\item When $A_t=0$, $b_t=0$ and $\Sigma_t=\Sigma$ we recover the heat equation transition semigroup
$$
\PP(\Xa_t(x)\in dy)=(2\pi t)^{-d/2}~\exp{\left(-\frac{1}{2t}~(y-x)^{\prime}\Sigma^{-1}(y-x)\right)}~dy.
$$
Note that in this case we have $(\alpha[t],\beta[t])=(0,I)$ and a linear growth variance
$$
\tau[t]=t~\Sigma\stackrel{t\to\infty}{\longrightarrow}\infty.
$$
\item When $\Sigma=I$ the above formula reduces to
$$
\PP(\Xa_t(x)\in dy)=\exp{(-c_t(x,y))}~dy,
$$
with the symmetric quadratic cost $c_t$ defined in \eqref{gibbs-cost}.
\item The Ornstein-Uhlenbeck diffusion corresponds to the case case $b_t=0$, $\Sigma_t=\Sigma\in \Sa_d^+$ and $A_t=A$ for some Hurwitz matrix $A$. It yields $\alpha[t]=0$, $\beta[t]=e^{tA}\stackrel{t\to\infty}{\longrightarrow}0$ and a uniformly bounded variance $\tau[t] \leq \int_0^{\infty}e^{sA}~\Sigma~e^{sA^{\prime}}~ds$.
\end{itemize}
}
\end{rmk}}

\subsection{{Conditional Gaussian distributions}}\label{conditional-dist-sec}

{For conciseness, let $X\sim\eta$ indicate that $X$ is a r.v. with probability distribution $\eta$ on some state space. We can interpret the Bayes' maps of Section \ref{ss-conjugate-Gauss} in terms of conditional Gaussian distributions. Specifically, assume that the r.v. $X\sim\nu_{m,\sigma}$ is observed by way of some linear-Gaussian transformation, namely,
$$
Y=\alpha+\beta X+\tau^{1/2}G.
$$
The mean and covariance of $Y$ can readily be written as
$$
\EE(Y)=a_m(\theta)\quad\mbox{\rm and}\quad \Sigma_{Y,Y}:=\EE((Y-\EE(Y))(Y-\EE(Y))^{\prime})=b_{\sigma}(\theta),
$$
where $\theta:=(\alpha,\beta,\tau)$ and $G\sim\nu_{0,I}$ denotes a centered Gaussian r.v. independent of $X$. The conditional distribution of the r.v. $X$ given an observation $Y=y$ is Gaussian, specifically, 
$$
\PP(X\in dx~|~Y=y)=\PP(
Z_{\BB_{m,\sigma}(\theta)}(y)\in dx),
$$
where $\BB_{m,\sigma}$ is the Bayes map in Definition \ref{def-BB}.}

In the Kalman filtering literature, the matrix $\kappa$ is often called  the (Kalman) gain matrix as it reflects the degree to which each observation $Y=y$ is incorporated into the estimation of the state $X$. The gain matrix is sometimes given in terms of covariances matrices by the formulae 
$$
\kappa= \Sigma_{X,Y}\Sigma_{Y,Y}^{-1},
$$
where
$$
\Sigma_{X,Y}:=\EE((X-\EE(X))(Y-\EE(Y))^{\prime})=\Sigma_{X,X}~ \beta^{\prime}
\quad\mbox{\rm and}\quad 
\Sigma_{X,X}:=\sigma.
$$
Note that
\begin{equation}\label{equiv-k}
Z_{\BB_{m,\sigma}(\theta)}(y)\stackrel{\text{law}}{=}X^Y(y):=X+\kappa~(y-Y)
\end{equation}
To check this claim, note that $$
\EE(X^Y(y))=m+\kappa~(y-(\alpha+\beta m))=\EE(Z_{\BB_{m,\sigma}(\theta)}(y))
$$
and
$$
\begin{array}{l}
X^Y(y)-\EE(X^Y(y))=(I-\kappa\beta)(X-m)-\kappa\tau^{1/2}G\\
\\
\Longrightarrow
\Sigma_{X^Y(y),X^Y(y)}=(I-\kappa\beta)\sigma(I-\kappa\beta)^{\prime}+\kappa\tau\kappa^{\prime}=\sigma+\kappa(\beta\sigma \beta^{\prime}+\tau)\kappa^{\prime}-\kappa\beta\sigma-\sigma\beta^{\prime}\kappa^{\prime}.
\end{array}
$$
On the other hand, by  (\ref{ref-conjug-Z-par}) we have $\kappa (\beta~\sigma~\beta^{\prime}+\tau)=\sigma~\beta^{\prime}$. This implies that
$$
\Sigma_{X^Y(y),X^Y(y)}=\sigma-\kappa\beta\sigma=(\sigma^{-1}+ \beta^{\prime}~\tau^{-1}\beta)^{-1}=\varsigma.
$$
\begin{rmk}\label{rmk-gaussian-Sigma}
{Gaussian models of the form} \eqref{def-P-theta} encapsulate general models of the form 
$
\Xa:=\left(\begin{array}{c}
X\\
Y
\end{array}\right)
$ 
with a mean prescribed mean $\EE(\Xa)=\left(\begin{array}{c}
\EE(X)\\
\EE(Y)
\end{array}\right)$ and given covariance matrix
$$
\Sigma_{\Xa,\Xa}:=\EE((\Xa-\EE(\Xa))(\Xa-\EE(\Xa))^{\prime})=\left(\begin{array}{cc}
\Sigma_{X,X}&\Sigma_{X,Y}\\
\Sigma_{Y,X}&\Sigma_{Y,Y}
\end{array}\right).
$$ 
In this context, we have $\EE(X)=m$ and $\Sigma_{X,X}=\sigma$ as well as
$$
\alpha:=\EE(Y)-\beta ~\EE(X)\quad\mbox{\rm and}\quad
 \beta=\Sigma_{Y,X}\Sigma_{X,X}^{-1}.
$$
We also have the Schur complement
  $$\tau=
 \Sigma_{Y,Y}-\Sigma_{Y,X}\Sigma_{X,X}^{-1}~\Sigma_{X,Y}\succ 0,
 $$
where $\tau \succ 0$ if, and only if, $\Sigma_{\Xa,\Xa} \succ 0$.
\end{rmk}

We observe that
$$
\theta_0=\theta:=(\alpha,\beta,\tau)
\quad\text{yields} 
\quad\pi_{0}:=\nu_{m,\sigma}K_{\theta_0}=\nu_{m_{0},\sigma_{0}},
$$
with the parameters
$$
(m_0,\sigma_0)=(\alpha+\beta~m,\beta~\sigma~\beta^{\prime}+\tau).
$$
In addition, we have the conjugate property
$$
\theta_1:= \BB_{m,\sigma}(\theta_0)=\left(\alpha_1,\beta_1,\tau_1\right)
$$
with  the parameters
\begin{equation}\label{inter-tau-1-i}
\alpha_1+\beta_1~m_0=m, \qquad \beta_1=\sigma\beta^{\prime}\sigma_0^{-1}\quad \mbox{\rm and}\quad \tau_1^{-1}=\sigma^{-1}+\beta^{\prime}\tau^{-1}\beta.
\end{equation}
This yields the Gaussian Markov transport formula
$$
\pi_{1}:=\nu_{\overline{m},\overline{\sigma}} K_{\theta_1}=\nu_{m_1,\sigma_1},
$$
with  the parameters
\begin{equation}\label{inter-m-sigma-1}
m_1=\alpha_1+\beta_1~\overline{m}=m+\beta_1(\overline{m}-m_0)\quad \mbox{\rm and}\quad\sigma_1=\beta_1\overline{\sigma}\beta^{\prime}_1+\tau_1.
\end{equation}
Combining (\ref{inter-tau-1-i}) with the matrix inversion lemma we readily see that
$$
\sigma_0^{-1}=\tau^{-1}-\tau^{-1}\beta~\tau_1~\beta^{\prime}\tau^{-1},
$$
which implies
\begin{equation}\label{inter-beta-01}
\beta_1=\sigma\beta^{\prime}\tau^{-1}-\sigma\left(\beta^{\prime}\tau^{-1}\beta\right)~\tau_1~\beta^{\prime}\tau^{-1}=\tau_1\beta^{\prime}\tau^{-1},
\quad\text{hence}\quad
\tau_1^{-1}\beta_1=\beta^{\prime}~\tau^{-1}.
\end{equation}

\subsection{The Gaussian bridge problem}
 
Consider some probability measures  $\eta,\mu\in\Ma_1(\RR^d)$  and some reference probability measure $\Pa_0 \in\Pa(\RR^{2d})$ of the form $\Pa_0=\eta\times\Ka_0$. Assume that the Markov transition $\Ka_0$ from $\RR^d$ into itself is chosen so that $\mu\ll\eta \Ka_0$. This condition is clearly met for the linear Gaussian model
 (\ref{def-P-theta}) with the target marginal measures
 $(\eta,\mu):=(\nu_{m,\sigma},\nu_{\overline{m},\overline{\sigma}})$, the product measure $\Pa=\eta\otimes\mu$ and the reference probability measure
$$
\Pa_0= P_{\theta}
\quad\text{for any given $\theta=(\alpha,\beta,\tau)\in \Theta$.}
$$ 
For a given distribution $P_{\theta}$ associated with some reference parameter $\theta\in\Theta$, {the (static) Schr\"odinger bridge problem \eqref{def-entropy-pb-v2} from $\nu_{m,\sigma}$ to $\nu_{\overline{m},\overline{\sigma}}$ is equivalent to the problem}
\begin{equation}\label{opt}
\SS(\theta):=\argmin_{\theta_1\in\Omega_{m,\sigma}(\overline{m},\overline{\sigma})}\mbox{\rm Ent}\left(P_{\theta_1}~|~P_{\theta}\right)\end{equation}
with the subset
$$
\Omega_{m,\sigma}(\overline{m},\overline{\sigma}):=\left\{\theta\in\Theta~:~h_{m,\sigma}(\theta)=(\overline{m},\overline{\sigma})\right\}.
$$
Note that there is no need to specify the first coordinate of the parameter $\theta\in 
\Omega_{m,\sigma}(\overline{m},\overline{\sigma})$ because
$$
(\alpha,\beta,\tau)\in \Omega_{m,\sigma}(\overline{m},\overline{\sigma})
\quad \text{implies that}
\quad
\alpha=\overline{m}-\beta m.
$$

Given a reference measure $P_{\theta}$, the measure $P_{\SS(\theta)}$ is the minimal entropy probability distribution with prescribed marginal $\nu_{\overline{m},\overline{\sigma}}$ w.r.t. the second coordinate. Note that $\nu_{m,\sigma}$ is the marginal of both measures $P_{\theta}$ and $P_{\SS(\theta)}$ w.r.t. the first coordinate.

For any parameters 
$$
\theta_0=(\alpha,\beta,\tau)\in \Theta\quad\mbox{\rm and}\quad
\theta_1=(\iota,\kappa,\varsigma)\in \Theta
$$
we have the Boltzmann  relative entropy (a.k.a. 
Kullback Leibler divergence) formula
\begin{equation}
 \begin{array}{l}
  \displaystyle \mbox{\rm Ent}\left(P_{\theta_1}~|~P_{\theta_0}\right):=\int~\log\frac{dP_{\theta_1}}{dP_{\theta_0}}~d
P_{\theta_1}
\\
\\
 \displaystyle=\frac{1}{2}~D(\varsigma~|~\tau)+\frac{1}{2}~\Vert\tau^{-1/2} ((\iota+\kappa m)-(\alpha+\beta m))\Vert_F^2+\frac{1}{2}~\Vert\tau^{-1/2}(\kappa-\beta)~\sigma^{1/2}\Vert^2_F. 
\end{array}
\label{BL-intro-def}
\end{equation}
The proof of equality \eqref{BL-intro-def} follows from elementary manipulations and it is provided in Appendix \ref{BL-intro-def-proof} (on page~\pageref{BL-intro-def-proof}). 
 We also quote the following estimate
\begin{equation}\label{burg}
 \displaystyle\Vert \varsigma-\tau\Vert_F~\Vert \tau^{-1}\Vert_F
\leq \frac{1}{2},
{\quad\text{which implies}\quad}
 D(\varsigma~|~\tau)\leq \frac{5}{2}
  \left\Vert \tau^{-1}\right\Vert_F~\left\Vert\varsigma-\tau\right\Vert_F.
\end{equation}
 {A detailed proof of expression \eqref{burg}} is provided in Appendix \ref{BL-intro-def-proof} (see page~\pageref{BL-intro-def-proof}) and Section 11 in~\cite{dt-18}). We also note that
\begin{eqnarray}
H\left(P_{\theta_1}~|~P_{\theta_0}\right)&:=& \mbox{\rm Ent}(P_{\theta_1}~|~P_{\theta_0})+ \int~(\eta K_{\theta_1})(dy)~V(y)\nonumber \\
&=&-\int~\log{q_{\theta_0}(x,y)}~P_{\theta_1}(d(x,y))+\mbox{\rm Ent}(P_{\theta_1}~|~\eta\otimes \mu).
\label{entropic-cost}
\end{eqnarray}
In addition, 
$$
\theta_1\in\Omega_{m,\sigma}(\overline{m},\overline{\sigma})
\quad\text{if, and only if,}\quad
\eta K_{\theta_1}=\nu_{m,\sigma}K_{\theta_1}=\nu_{\overline{m},\overline{\sigma}}=\mu,
$$
which shows that
\begin{equation}
\SS(\theta)=\argmin_{\theta_1\in\Omega_{m,\sigma}(\overline{m},\overline{\sigma})}
H\left(P_{\theta_1}~|~P_{\theta}\right).
\label{eqSolSH}
\end{equation}
Switching the role of the parameters $(\overline{m},\overline{\sigma})$ and $(m,\sigma)$, for a given distribution $\overline{P}_{\theta}$ associated with some reference parameter $\theta\in\Theta$, solving the Schr\"odinger bridge from  $\nu_{\overline{m},\overline{\sigma}}$  to $\nu_{m,\sigma}$ is equivalent to  solving the minimization problem 
\begin{equation}\label{over-S-def}
\overline{\SS}(\theta):=\argmin_{\theta_1\in\Omega_{\overline{m},\overline{\sigma}}(m,\sigma)}\mbox{\rm Ent}\left(\overline{P}_{\theta_1}~|~\overline{P}_{\theta}\right).
\end{equation}

\section{Bridges and transport maps}\label{sec-bridges}

\subsection{Entropic transport maps}
The solution of the minimization problem (\ref{opt}) clearly depends on the choice of the reference parameters $\theta=(\alpha,\beta,\tau)$, as well as on the parameters  $(\overline{m},\overline{\sigma})$ and $(m,\sigma)$ of the target marginal measures. To be precise, consider the  matrix
\begin{equation}\label{def-w-1}
 \varpi_{\theta}^{-1}:= \gamma_{\theta}\,\gamma_{\theta}^{\prime}\in\Sa^+_{d}
\quad\mbox{\rm with}\quad  \gamma_{\theta}:=\overline{\sigma}^{1/2}~\cchi_{\theta}~\sigma^{1/2}
\quad\mbox{\rm and}\quad
\cchi_{\theta}:=\tau^{-1}\beta,
\end{equation}
{and} denote by $r_{\theta}$ the (unique) {positive-definite fixed point of the Riccati map  (\ref{ricc-maps-def}) associated with $\varpi_{\theta}$, i.e.,}
\begin{equation}\label{def-r-t}
\mbox{\rm Ricc}_{\varpi_{\theta}}\left(r_{\theta}\right)=r_{\theta}=-\frac{\varpi_{\theta}}{2}+\left(\varpi_{\theta}+\left(\frac{\varpi_{\theta}}{2}\right)^2\right)^{1/2}.
\end{equation} 
{A closed form expression of $r_{\theta}$ is given by (\ref{def-fix-ricc-1}), simply replacing $\varpi$ by $\varpi_{\theta}$. A proof of the fixed-point formula (\ref{def-r-t}) is provided in Appendix \ref{gauss-sinhorn-details} (on page~\pageref{Sec-fixed-p}, see Eq. \eqref{fp-ricc}).}

We are now in position to state the first main result.
\begin{theo}\label{Th1}
The Schr\"odinger bridge map (\ref{opt}) is given by 
$\SS(\theta):=(\iota_{\theta},\kappa_{\theta},\varsigma_{\theta})\in  \Omega_{m,\sigma}(\overline{m},\overline{\sigma})$
with the parameters
\begin{equation}\label{def-Sa}
{
\varsigma_{\theta}:=\overline{\sigma}^{1/2}~r_{\theta}~\overline{\sigma}^{1/2}, \quad
 \kappa_{\theta}:=\varsigma_{\theta}~\cchi_{\theta}, \quad \text{and} \quad
\iota_{\theta} = \overline{m} - \kappa_{\theta}~m.
}
\end{equation}

\end{theo}

The bridge map given by \eqref{def-Sa} is the limiting value of 
Sinkhorn bridge maps (cf. for instance the exponential stability theorem stated in Section~\ref{quant-sec}). By the uniqueness of the Schr\"odinger bridge \eqref{opt}, Theorem \ref{Th1} is a also a direct consequence of the fixed point equation stated in Corollary~\ref{cor-commute-ss}. The transport property stems from the equivalences
\begin{eqnarray}
\kappa_{\theta}~\sigma~\kappa_{\theta}^{\prime}+\varsigma_{\theta}=\overline{\sigma}
&\Longleftrightarrow & \varsigma_{\theta}+\varsigma_{\theta}~\left(\overline{\sigma}^{1/2}~\varpi_{\theta}~\overline{\sigma}^{1/2}\right)^{-1}~\varsigma_{\theta}=\overline{\sigma}\nonumber\\
&\Longleftrightarrow & r_{\theta} ~\varpi_{\theta}^{-1}~r_{\theta}+ r_{\theta}=I\Longleftrightarrow
r_{\theta}=
\mbox{\rm Ricc}_{\varpi_{\theta}}\left(r_{\theta}\right), \label{ref-fix-point-intro}
\end{eqnarray}
where the first assertion comes from the fact that
$$
(\overline{\sigma}^{1/2}~\varpi_{\theta}~\overline{\sigma}^{1/2})^{-1}=
\tau^{-1}~\beta~\sigma\beta^{\prime}~\tau^{-1}=\cchi_{\theta}~\sigma~\cchi_{\theta}^{\prime}.
$$
and the last assertion is proved in Appendix \ref{app-ricc} (see Eq. \eqref{fp-ricc}). Theorem~\ref{Th1} can also be verified combining the dual formulae \eqref{ref-uv-infty-intro-eq} with a theorem by Nutz (Theorem 2.1 in~\cite{nutz}).

{We recall that Schr\"odinger bridges can also be written in terms of the entropic cost function $H(\cdot~|~P_{\theta})$ defined in \eqref{entropic-cost}. In particular, from Theorem \ref{Th1} and \eqref{eqSolSH} we have}
$$
H\left(P_{\SS(\theta)}~|~P_{\theta}\right)=
\min_{\theta_1\in\Omega_{m,\sigma}(\overline{m},\overline{\sigma})}
H\left(P_{\theta_1}~|~P_{\theta}\right).
$$
{We can also express the bridge transport map between the distributions $\nu_{m,\sigma}$ and $ \nu_{\overline{m},\overline{\sigma}}$ in terms of the random transformation \eqref{random-Z-map}, namely,}
\begin{equation}\label{bridge-map-i}
Z_{\SS(\theta)}(x)=\overline{m}+\kappa_{\theta}~\left(x-m\right)+\varsigma_{\theta}^{1/2}~G.
\end{equation}
{In particular, for the multivariate Gaussian models discussed in Remark~\ref{rmk-gaussian-Sigma}, the Schr\" odinger bridge $P_{\SS(\theta)}$ is the distribution of a random variable
$
\Xa:=\left(\begin{array}{c}
X\\
Y
\end{array}\right)
$ 
with mean $\EE(\Xa)=\left(\begin{array}{c}
m\\
\overline{m}
\end{array}\right)$ and covariance matrix
$$
\Sigma_{\Xa,\Xa}=\left(\begin{array}{cc}
\Sigma_{X,X}&\Sigma_{X,Y}\\
\Sigma_{Y,X}&\Sigma_{Y,Y}
\end{array}\right)=\left(\begin{array}{cc}
\sigma&\sigma~\kappa_{\theta}^{\prime}\\
\kappa_{\theta}~\sigma&\overline{\sigma}
\end{array}\right).
$$ 
When $\beta=I$ and $\tau=tI$ the above formula reduces to  formula (2) discussed in the recent article~\cite{bunne2023schrodinger}. 
Theorem~\ref{Th1} can be seen as a simplification and a generalization of 
Schr\" odinger bridge formulae recently presented in the series of articles~\cite{agueh,bojitov,bunne2023schrodinger,loubes,janadi,mallasto} when the drift matrix $\beta$ is arbitrary and $\tau$ is an arbitrary positive definite matrix. Following the discussion given in Remark~\ref{ref-general-models}, this formula also apply to all the continuous time Gaussian models used in machine learning applications of Schr\"odinger bridges. To the best of our knowledge this general formula is new.}
\begin{rmk}
{Note that $\SS(\theta)$ does not depend on $\alpha$.}
Also observe that
$$
\cchi_{\SS(\theta)}=\varsigma_{\theta}^{-1}\kappa_{\theta}=\cchi_{\theta}
\quad\text{implies that}\quad
 \varpi_{\SS(\theta)}= \varpi_{\theta}\quad \mbox{and}\quad
\SS^2:=\SS\circ \SS=\SS.
$$
\end{rmk}

\subsection{Dual bridge maps}
In this section, we discuss dual bridge maps and define several dual quantities which will be used throughout the paper.
\begin{theo}\label{theo-2-S-o-S}
The Schr\"odinger bridges $(\SS,\overline{\SS})$ defined in (\ref{opt}) and (\ref{over-S-def})
satisfy the commutation property
\begin{equation}\label{comm-intro}
\BB_{m,\sigma}\circ\SS=\overline{\SS}\circ\BB_{m,\sigma}\quad
\mbox{and}\quad
\BB_{\overline{m},\overline{\sigma}}\circ\overline{\SS}=\SS\circ
\BB_{\overline{m},\overline{\sigma}}
\end{equation} 
with the Bayes maps $\BB_{m,\sigma}$ and $\BB_{\overline{m},\overline{\sigma}}$ defined in (\ref{ref-conjug-Z-par}). 
\end{theo}
{A proof of Theorem \ref{theo-2-S-o-S} is provided in Appendix~\ref{sec-tech-proofs} (on page~\pageref{theo-2-S-o-S-proof}). The commutation property \eqref{comm-intro} also yields the straightforward corollary below.} 
\begin{cor}\label{cor-commute-ss}
We have the fixed point properties
\begin{eqnarray*}
(\BB_{\overline{m},\overline{\sigma}}\circ\BB_{m,\sigma})\circ \SS&=\SS=&
\SS\circ(\BB_{\overline{m},\overline{\sigma}}\circ\BB_{m,\sigma}).
\end{eqnarray*}
\end{cor}
For Gaussian models, {the iterations of the Sinkhorn algorithm coincide with the iterates of Bayes maps (see, e.g., \eqref{rec-bb}) and, in particular, Corollary~\ref{cor-theta-even-cv} shows that}
$$
\lim_{n\rightarrow\infty}(\BB_{\overline{m},\overline{\sigma}}\circ\BB_{m,\sigma})^n(\theta)=
\SS(\theta).
$$
By symmetry arguments we also have
$$
(\BB_{m,\sigma}\circ \BB_{\overline{m},\overline{\sigma}})\circ \overline{\SS}=\overline{\SS}=
\overline{\SS}\circ(\BB_{m,\sigma}\circ \BB_{\overline{m},\overline{\sigma}}).
$$

Theorem~\ref{theo-2-S-o-S} {implies that} the dual bridge parameter $\overline{\SS}(\theta_1)$ between the distributions  $ \nu_{\overline{m},\overline{\sigma}}$ and $\nu_{m,\sigma}$ and reference parameter $\theta_1:=\BB_{m,\sigma}(\theta)$ can be computed using the Bayes transform $\BB_{m,\sigma}(\SS(\theta))$ of the bridge parameter $\SS(\theta)$ between  $\nu_{m,\sigma}$ and   $ \nu_{\overline{m},\overline{\sigma}}$. {As a consequence}
$$
\BB_{m,\sigma}(\SS(\theta))=\left(m-\sigma~\kappa_{\theta}^{\prime}~\overline{\sigma}^{-1}~\overline{m},~\sigma~\kappa_{\theta}^{\prime}~\overline{\sigma}^{-1},(\sigma^{-1}+\kappa_{\theta}^{\prime}~\varsigma_{\theta}^{-1}\kappa_{\theta}
)^{-1}\right).
$$
 Theorem~\ref{Th1} also ensures that 
 the   Schr\"odinger bridge map $\overline{\SS}$ from  $\nu_{\overline{m},\overline{\sigma}}$  to $\nu_{m,\sigma}$  with reference parameter $\theta_1=(\alpha_1,\beta_1,\tau_1)$ is given by
$$
\overline{\SS}(\theta_1):=(\overline{\iota}_{\theta_1},\overline{\kappa}_{\theta_1},\overline{\varsigma}_{\theta_1})\quad \mbox{\rm with}\quad
\overline{\iota}_{\theta_1}:=m-\overline{\kappa}_{\theta_1}~\overline{m}
$$
and the parameters
\begin{equation}\label{def-over-Sa}
\overline{\kappa}_{\theta_1}:=\overline{\varsigma}_{\theta_1}~\cchi_{\theta_1}\quad
\mbox{and}\quad
\overline{\varsigma}_{\theta_1}:=\sigma^{1/2}~\overline{r}_{\theta_1}~\sigma^{1/2}
\quad\mbox{\rm where}\quad
\cchi_{\theta_1}:=\tau_1^{-1}\beta_1.
\end{equation}
In \eqref{def-over-Sa}, $\overline{r}_{\theta_1}$ stands for the positive definite fixed point of the Riccati map associated with the matrix 
\begin{equation}\label{def-wo-1}
\overline{\varpi}_{\theta_1}^{-1}:=\overline{\gamma}_{\theta_1}\overline{\gamma}_{\theta_1}^{\prime},\quad \mbox{\rm where}\quad
\overline{\gamma}_{\theta_1}:=
\sigma^{1/2}~\cchi_{\theta_1}~\overline{\sigma}^{1/2}.
\end{equation}
By \eqref{def-r-t} we have 
\begin{equation}\label{def-over-r}
\mbox{\rm Ricc}_{\overline{\varpi}_{\theta_1}}(\overline{r}_{\theta_1})=\overline{r}_{\theta_1}:=-\frac{\overline{\varpi}_{\theta_1}}{2}+\left(\overline{\varpi}_{\theta_1}+\left(\frac{\overline{\varpi}_{\theta_1}}{2}\right)^2\right)^{1/2}
\end{equation} 
{and \eqref{inter-beta-01} implies that}
\begin{equation}
\theta_1:=\BB_{m,\sigma}(\theta)\quad \mbox{\rm and}\quad
\theta=(\alpha,\beta,\tau)
\nonumber
\end{equation}
{together yield}
\begin{equation}
\cchi_{\theta_1}=\tau_1^{-1}\beta_1=\beta^{\prime}\tau^{-1}=\cchi_{\theta}^{\prime}, \quad
\overline{\gamma}_{\theta_1}=\gamma_{\theta}^{\prime}\quad \mbox{\rm and}\quad \overline{\varpi}_{\theta_1}^{-1}=\gamma_{\theta}^{\prime}\gamma_{\theta}.
\nonumber
\end{equation}
{
\begin{rmk}\label{rmq-one-dim-ricc}
Note that for one-dimensional models we have $\overline{\varpi}_{\theta_1}={\varpi}_{\theta}$ and, therefore, $\mbox{\rm Ricc}_{\overline{\varpi}_{\theta_1}}=\mbox{\rm Ricc}_{{\varpi}_{\theta}}$.
\end{rmk}
}
The  fixed point  matrices $(r_{\theta},\overline{r}_{\theta_1})$ defined in (\ref{def-r-t}) and (\ref{def-over-r}) are connected with the formulae
\begin{equation}\label{connect-fp}
r_{\theta}^{-1}=I+ \gamma_{\theta}~\overline{r}_{\theta_1}~  \gamma_{\theta}^{\prime}\quad \mbox{and}\quad
\overline{r}_{\theta_1}^{-1}=
I+  \gamma_{\theta}^{\prime}~r_{\theta}~ \gamma_{\theta}
\end{equation}
The proof of (\ref{connect-fp}) is rather technical, thus it is provided in Appendix~\ref{app-ricc} (on page~\pageref{stat-Ricc-sec}). In terms of the rescaled fixed points $(\varsigma_{\theta},\overline{\varsigma}_{\theta_1})$ formulae \eqref{connect-fp} take the following form
\begin{equation}\label{connect-fp-v2}
\varsigma_{\theta}^{-1}=\overline{\sigma}^{-1}+\cchi_{\theta}~\overline{\varsigma}_{\theta_1}~\cchi_{\theta}^{\prime}\quad \mbox{\rm and}\quad \overline{\varsigma}_{\theta_1}^{-1}=\sigma^{-1}+ \cchi_{\theta}^{\prime}~
\varsigma_{\theta}~\cchi_{\theta}.
\end{equation}

Finally, {we can rewrite Theorem~\ref{theo-2-S-o-S} in terms of transport maps as shown below.}
\begin{cor}\label{cor-dual-bm}
The dual transport map between the distributions  $\nu_{\overline{m},\overline{\sigma}}$ and $\nu_{m,\sigma}$ and reference parameter $
\theta_1:=\BB_{m,\sigma}(\theta)
$ {is} given by
\begin{equation}\label{dbridge-map-i}
Z_{\overline{\SS}(\theta_1)}(y):=m+\overline{\kappa}_{\theta_1}\left(y-\overline{m}\right)+\overline{\varsigma}_{\theta_1}^{1/2}~G
\end{equation}
with the parameters
\begin{equation}\label{fix-intro}
\overline{\kappa}_{\theta_1}=\sigma~\kappa_{\theta}^{\prime}~\overline{\sigma}^{-1}=\overline{\varsigma}_{\theta_1}~\cchi_{\theta_1}\quad \mbox{and}\quad
\overline{\varsigma}_{\theta_1}=(\sigma^{-1}+\kappa_{\theta}^{\prime}~\varsigma_{\theta}^{-1}\kappa_{\theta}
)^{-1}=\sigma^{1/2}~\overline{r}_{\theta_1}~\sigma^{1/2}.
\end{equation}
\end{cor}

\subsection{Dynamic and static bridges}\label{sec-OU-bridges}

Consider the linear diffusion process $(\Xa_t)_{t\in [0,T]}$ defined on the time interval $[0,T]$ by the stochastic differential equation (\ref{stochastic-OU}) starting from some random variable $\Xa_0$ with distribution $\eta$. Specifically, we have
$$
d\Xa_t=\left(A_t~\Xa_t+b_t\right)~dt+\Sigma_t^{1/2}~dW_t,\quad \text{with}\quad
\mbox{\rm Prob}(\Xa_0\in dx)=\eta(dx).
$$
We fix a terminal time horizon $T>0$ and let  $\Pb$ be the distribution of  the random path $\Xa:=(\Xa_t)_{t\in [0,T]}$ on the space of $\RR^d$-valued continuous functions on the time interval $[0,T]$, denoted by $\Ca([0,T],\RR^d)$.  The distribution of the diffusion $\Xa$ conditioned on $\Xa_0=x$ and $\Xa_T=y$ is given by
$$
   \Pb^{x,y}(d\omega):=\Pb( d\omega~|~(\omega_0,\omega_T)=(x,y)).
$$
Moreover, if we let $p_{t,T}(x,y)$ denote the density of $\Xa_T$ conditional on $\Xa_t=x$, i.e.,
 $$
 p_{t,T}(x,y)dy:=\PP(\Xa_T\in dy~|~\Xa_t=x)
 $$
 then we obtain
 $$
 \nabla_{x} \log p_{t,T}(x,y)=\Ea_{t,T}(A)^{\prime}~  \Sigma_{t,T}^{-1}~\left(y-\left(\Ea_{t,T}(A)~x+\int_t^T \Ea_{s,T}(A)~b_s~ ds\right)\right),
 $$
 where $\Ea_{t,T}(A)$ is the exponential semigroup associated to the flow of matrices $A_t$ (see Remark \ref{ref-general-models}) and 
 $$
 \Sigma_{t,T}:=
 \int_t^T~\Ea_{s,T}(A)~\Sigma_s~\Ea_{s,T}(A)^{\prime}~ds
 $$
is the conditional covariance matrix. Thus, as shown in~\cite{delyon},    $  \Pb^{x,y}$  is the distribution of the pinned random path $\Xb^{x,y}:=(\Xb^{x,y}_t)_{t\in [0,T]}$ stating at $\Xb^{x,y}_0=x$ en ending at $\Xb^{x,y}_T=y$ satisfying the stochastic differential equation
\begin{eqnarray}
d\Xb^{x,y}_t &=&(A_t~\Xb^{x,y}_t+b_t)dt\nonumber\\
&&+\Sigma_t~\left(\Ea_{t,T}(A)^{\prime}~  \Sigma_{t,T}^{-1}~\left(y-\left(\Ea_{t,T}(A)~\Xb^{x,y}_t+\int_t^T \Ea_{s,T}(A)~b_s~ ds\right)\right)\right)dt \nonumber\\
&&+ \Sigma_t^{1/2}dW_t.\nonumber
\end{eqnarray}
{As underlined in~\cite{delyon}, in terms of the stochastic flow $\Xa_t(x)$ starting at $\Xa_0(x)=x$ defined in (\ref{stochastic-OU}) using (\ref{equiv-k}) we check that
the distribution of the diffusion $\Xb^{x,y}_t$ coincides with the distribution of the process
\begin{equation}\label{equiv-k-proc}
\Xa_t(x)+C_{t,T}~C_{T,T}^{-1}~(y-\Xa_T(x))\stackrel{\text{law}}{=}\Xb^{x,y}_t
\end{equation}
with the covariance matrices
$$
C_{t,T}:=\EE((\Xa_t(x)-\EE(\Xa_t(x)))(\Xa_T(x)-\EE(\Xa_T(x)))^{\prime})=\int_0^{t}\Ea_{s,t}(A)~\Sigma_s~\Ea_{s,T}(A)^{\prime}~ds.
$$}

Consider now the $\Pb$-marginal distribution  of the random states $(\Xa_0,\Xa_T)$ defined by
 $$
  \Pa(d(x,y)):=
  \eta(dx)~\Ka(x,dy)\quad
  \mbox{\rm with}\quad
  \Ka(x,dy):=\PP(\omega_T\in dy~|~\omega_0=x).
 $$
 In this notation, we have the disintegration formula 
$$
\Pb(d\omega):=\int_{\RR^d\times\RR^d}~ \Pb^{x,y}(d\omega)~
\Pa(d(x,y)).
$$

 \begin{rmk}
 The static Schr\" odinger bridge with reference measure $\Pa=\eta\times\Ka$ is given by
 $$
 \Pa_{\eta,\mu}:=\argmin_{\Qa\,\in\, \Ca(\eta,\mu)}
\mbox{\rm Ent}(\Qa~|~\eta\times\Ka).
 $$
 Choosing $ (\eta,\mu):=(\nu_{m,\sigma},\nu_{\overline{m},\overline{\sigma}})$
 and $\theta=(\alpha[T],\beta[T],\tau[T])$ with the parameters $(\alpha[T],\beta[T],\tau[T])$ as in (\ref{solution-stochastic-OU}) we have
 $\Ka=K_{\theta}
 $ and $ \Pa_{\eta,\mu}=\eta\times K_{\SS(\theta)}$, with the Schr\"odinger bridge map $\SS(\theta)$ defined in Theorem~\ref{Th1}.
 \end{rmk}

 Arguing as above, any probability measure $\Qb\ll \Pb$ on  $\Ca([0,T],\RR^d)$ with marginal density $\eta(dx)$ at time $t=0$ can be disintegrated with respect to the initial and final conditions $(\omega_0,\omega_T)=(x,y)$, namely,
$$
\Qb(d\omega):=\int_{\RR^d\times\RR^d}~ \Qb^{x,y}(d\omega)~
\Qa(d(x,y)),
$$
where
   \begin{eqnarray*}
  \Qb^{x,y}(d\omega)&:=&\Qb( d\omega~|~(\omega_0,\omega_T)=(x,y)), \quad \text{and}\\
  \Qa(d(x,y))&:=&
  \eta(dx)~\La(x,dy)\quad
  \mbox{\rm with}\quad
  \La(x,dy):=\Qb(\omega_T\in dy~|~\omega_0=x).
  \end{eqnarray*} 
This yields the entropy factorization
$$
\mbox{\rm Ent}(\Qb~|~\Pb)=
\mbox{\rm Ent}(\Qa~|~\Pa)+\int~\mbox{\rm Ent}(~\Qb^{x,y}~|~\Pb^{x,y})~\Qa(d(x,y)).
$$
Let $ \Cb(\eta,\mu)$ be the set of probability measures $\Qb$ on path space  $\Ca([0,T],\RR^d)$ with marginals $\eta$ and $\mu$ at time $t=0$ and $t=T$. {The measure on path space $\Ca([0,T],\RR^d)$ obtained as
\begin{equation}
\Pb_{\eta,\mu} = \arg\min_{\Qb \in \Ca(\eta,\mu)} {\rm Ent}(\Qb~|~\Pb)
\label{eq_def_dynSB}
\end{equation}
is the usually termed the \textit{dynamic} Schr\"odinger bridge between $\eta$ and $\mu$. We can readily connect the static \eqref{def-entropy-pb-v2} and dynamic \eqref{eq_def_dynSB} Schr\"odinger bridge problems by choosing $\Qb^{x,y}=\Pb^{x,y}$, which yields}
$$
\inf_{\Qb\,\in\, \Cb(\eta,\mu)} \mbox{\rm Ent}(\Qb~|~\Pb)=\inf_{\Qa\,\in\, \Ca(\eta,\mu)}
\mbox{\rm Ent}(\Qa~|~\Pa).
$$
In addition (cf.~\cite{chen-phd,essid,follmer,leonard}), the static and dynamic Schr\" odinger bridges are connected by the formulae
$$
\Pb_{\eta,\mu}(d\omega):=\int \Pb^{x,y}(d\omega) \Pa_{\eta,\mu}(d(x,y))~~
\Longrightarrow~~
\Pb_{\eta,\mu}=\argmin_{\Qb\,\in\, \Cb(\eta,\mu)} \mbox{\rm Ent}(\Qb~|~\Pb).
$$

\subsection{Schr\"odinger potential functions}\label{schrod-intro}

In the context of Gaussian models,  the bridge distribution  $\Pa=P_{\SS(\theta)}$ discussed in (\ref{sinhorn-entropy-form-Sch-lim}) and (\ref{opt}) can be expressed  in terms of Schr\" odinger potential functions $(\UU_{\theta},\VV_{\theta})$ that depend on the reference parameter $\theta\in\Theta$.  These potential functions satisfy the bridge equation
\begin{equation}\label{ref-uv-infty-intro-eq}
\begin{array}{rcl}
P_{\SS(\theta)}(d(x,y))&=& \PP_{\theta}(d(x,y)):=
e^{-\UU_{\theta}(x)}~q_{\theta}(x,y)~e^{-\VV_{\theta}(y)}~dxdy.
\end{array}
\end{equation}
{Note that the potentials $(\UU_{\theta},\VV_{\theta})$ in \eqref{ref-uv-infty-intro-eq} are unique up to} an additive constant.
Choosing $(x,y)=(m,\overline{m})$ and setting $m_0:=(\alpha+\beta m)$ we readily check that potential functions satisfy the identity
\begin{equation}\label{ref-uv-infty}
\VV_{\theta}(\overline{m})+\UU_{\theta}(m)=\frac{1}{2}~\log{\mbox{\rm det}(\sigma)}+
\frac{1}{2}~\log{\mbox{\rm det}(\varsigma_{\theta}\tau^{-1})}-\frac{1}{2}~\left(m_0-\overline{m}\right)^{\prime}\tau^{-1}
\left(m_0-\overline{m}\right).
\end{equation}
As in (\ref{sinhorn-entropy-form-Sch}), the potential functions $(\UU_{\theta},\VV_{\theta})$ can be estimated using the Sinkhorn algorithm. We refer to Section~\ref{gauss-pot-sect} for a refined analysis of these approximations. 

\begin{theo}\label{theo-s-pot}
For any $\theta=(\alpha,\beta,\tau)\in\Theta$ we have
\begin{eqnarray*}
\VV_{\theta}(y+\overline{m})-\VV_{\theta}(\overline{m})&=&
y^{\prime}~\tau^{-1}~\left(m_{0}-\overline{m}\right)+\frac{1}{2}~
y^{\prime}~\left(\varsigma_{\theta}^{-1}-\tau^{-1}\right)~y, \quad\text{and}\\
\UU_{\theta}(x+m)-\UU_{\theta}(m)&=&x^{\prime}~\beta^{\prime}\tau^{-1}(\overline{m}-m_0)
+\frac{1}{2}~x^{\prime}\left(\overline{\varsigma}_{\theta_1}^{-1}-\beta^{\prime}\tau^{-1}\beta\right)
~x,
\end{eqnarray*}
with  $\theta_1=\BB_{m,\sigma}(\theta)$, $m_0$ as in (\ref{ref-uv-infty})  and  $(\varsigma_{\theta},\overline{\varsigma}_{\theta_1})$ defined in (\ref{def-Sa}) and (\ref{def-over-Sa}).
\end{theo}
\proof A detailed proof of the theorem is provided in Appendix~\ref{sch-appendix} on page~\pageref{theo-s-pot-proof}, see also Corollary~\ref{end-cor}. Next, we sketch an elementary and direct proof based on the identification of the quadratic forms involved in (\ref{ref-uv-infty-intro-eq}).
For instance, using (\ref{ref-uv-infty-intro-eq}) we have
$$
\begin{array}{l}
\displaystyle
\frac{\exp{((\UU_{\theta}-U)(x))}}{\sqrt{\mbox{det}(2\pi\varsigma_{\theta})}}~\exp{\left(-\frac{1}{2}~\Vert \varsigma_{\theta}^{-1/2}
\left((y-\overline{m})-\kappa_{\theta} (x-m)\right)\Vert_F^2\right)}\\
\\
\displaystyle=\frac{\exp{(-\VV_{\theta}(y))}}{\sqrt{\mbox{det}(2\pi\tau)}}~\exp{\left(-\frac{1}{2}\Vert \tau^{-1/2} (y-\overline{m})-\tau^{-1/2}\left((m_0-\overline{m})+\beta (x-m)\right)\Vert_F^2\right)}
\end{array}
$$
The proof of the first assertion simply rely on the identification of the terms of the quadratic function w.r.t. coordinate $(y-\overline{m})$.
On the other hand, using (\ref{ref-uv-infty-intro-eq}) and the commutation Theorem \ref{theo-2-S-o-S} we also have the conjugate formulae
$$
P_{\SS(\theta)}(d(x,y))=e^{-V(y)}~dy~K_{\overline{\SS}(\theta_1)}(y,dx)=
e^{-\VV_{\theta}(y)}~q_{\theta}(x,y)~e^{-\UU_{\theta}(x)}~dxdy.
$$
This yields
$$
\begin{array}{l}
\displaystyle
\frac{\exp{((\VV_{\theta}-V)(y))}}{\sqrt{\mbox{det}(2\pi\overline{\varsigma}_{\theta_1})}}~\exp{\left(-\frac{1}{2}~\Vert \overline{\varsigma}_{\theta_1}^{-1/2}
\left((x-m)-\overline{\kappa}_{\theta_1} (y-\overline{m})\right)\Vert_F^2\right)}\\
\\
\displaystyle=\frac{\exp{(-\UU_{\theta}(x))}}{\sqrt{\mbox{det}(2\pi\tau)}}~
~\exp{\left(-\frac{1}{2}\Vert \tau^{-1/2} (y-\overline{m})-\tau^{-1/2}\left((m_0-\overline{m})+\beta (x-m)\right)\Vert_F^2\right)}
\end{array}
$$
The proof of the second assertion simply relies on the identification of the terms of the quadratic function w.r.t. coordinate $(x-m)$. \cqfd

\begin{rmk}Using  (\ref{inter-beta-01}) we have
$$
\tau_1^{-1}=\sigma^{-1}+\beta^{\prime}\tau^{-1}\beta
\quad\mbox{\rm and}\quad
\tau^{-1}_1(m_1-m)=
\tau^{-1}_1\beta_1(\overline{m}-m_0)=\beta^{\prime}\tau^{-1}(\overline{m}-m_0)
$$
Combining the above formula with (\ref{inter-tau-1-i}) we check that
\begin{equation}\label{alter-UU}
\UU_{\theta}(x+m)-\UU_{\theta}(m)=(U(x+m)-U(m))+x^{\prime}~\tau_1^{-1}~\left(m_{1}-m\right)
+\frac{1}{2}~x^{\prime}\left(\overline{\varsigma}_{\theta_1}^{-1}-\tau_{1}^{-1}\right)
~x.
\end{equation}
\end{rmk}

\subsection{Entropic regularization}
\subsubsection{Bridge transport maps}
Consider the reference parameter 
\begin{equation}\label{theta-t-def}
\theta(t):=(\alpha,\beta,tI)\quad \mbox{\rm for some $t>0$.}
\end{equation}
The bridge transport map (\ref{bridge-map-i}) associated with the reference parameter $\theta(t)$ takes the form
$$
Z_{\SS(\theta(t))}(x)=\overline{m}+\kappa_{\theta(t)}~\left(x-m\right)+\varsigma_{\theta(t)}^{1/2}~G
$$
with the matrices
$$
\kappa_{\theta(t)}:=\frac{\varsigma_{\theta(t)}}{t}~\beta\quad \mbox{\rm and}\quad
\frac{\varsigma_{\theta(t)}}{t}=\overline{\sigma}^{1/2}~\frac{r_{\theta(t)}}{t}~\overline{\sigma}^{1/2}.
$$
If we now consider the conjugate parameter
$$
\theta_1(t):=\BB_{m,\sigma}(\theta(t))=(\alpha_1(t),\beta_1(t),\tau_1(t))\quad \mbox{\rm and}\quad
\nu_{\overline{m},\overline{\sigma}} K_{\theta_1(t)}=\nu_{m_1(t),\sigma_1(t)}
$$
{then the bridge transport map \eqref{dbridge-map-i}} associated with $\theta_1(t)$ takes the form
$$
Z_{\overline{\SS}(\theta_1(t))}(y):=m+\overline{\kappa}_{\theta_1(t)}~\left(y-\overline{m}\right)+\overline{\varsigma}_{\theta_1(t)}^{1/2}~G
$$
with the parameters
$$
\overline{\kappa}_{\theta_1(t)}=\frac{\overline{\varsigma}_{\theta_1(t)}}{t}~\beta^{\prime}\quad \mbox{and}\quad
\frac{\overline{\varsigma}_{\theta_1(t)}}{t}=\sigma^{1/2}~\frac{\overline{r}_{\theta_1(t)}}{t}~\sigma^{1/2}.
$$
Also, we have
\begin{eqnarray}
 \varpi_{\theta(t)}&=&t^{2}~ \omega    \quad\mbox{\rm with}\quad
   \omega:= 
 \overline{\sigma}^{-1/2}~\sigma_{\beta}^{-1}~\overline{\sigma}^{-1/2}
 \quad\mbox{and}\quad\sigma_{\beta}:=\beta\sigma
 \beta^{\prime}, \quad\text{and}\nonumber\\
   \overline{\varpi}_{\theta_1(t)}&=&t^2~\omega_1
 \quad\mbox{\rm with}\quad    \omega_1:=
~\sigma^{-1/2}~ \overline{\sigma}_{\beta^{\prime}}^{-1}~\sigma^{-1/2}
 \quad\mbox{and}\quad \overline{\sigma}_{\beta^{\prime}}:=\beta^{\prime}\overline{\sigma}\beta\label{rmk-theta-t-1}.
\end{eqnarray}

\subsubsection{Independence property}
Next result reflects the independence properties of the bridge maps when the regularization parameter $t\rightarrow\infty$. By (\ref{def-fix-ricc-1}), {choosing the parameter $\theta(t)$ defined in (\ref{theta-t-def})}, we have
\begin{equation}\label{def-fix-t-large}
(I+t^{-2}~\varpi^{-1})^{-1}\preceq r_{\theta(t)}\preceq I\quad \mbox{\rm and}\quad
(I+t^{-2}~\varpi_1^{-1})^{-1}\preceq \overline{r}_{\theta_1(t)}\preceq I.
\end{equation}
{
\begin{rmk}\label{bridge-gain-unstable}
Note that the bridge gain matrix $\kappa_{\theta(t)}$ may be unstable.
For instance, for one dimensional models with $\beta=1$, we have $\omega^{-1}:= 
 \overline{\sigma}\sigma$ and using (\ref{def-fix-t-large}) we arrive at
$$
\kappa_{\theta(t)}=\frac{\overline{\sigma}}{t}~r_{\theta(t)}\geq \frac{\overline{\sigma}}{t}~\left(1+\frac{\overline{\sigma}}{t}~\frac{\sigma}{t}\right)^{-1}.
$$
Choosing $\overline{\sigma}$ sufficiently large,  we have
$$
\frac{\overline{\sigma}}{t}\left(1-\frac{\sigma}{t}\right)>1 \quad \text{which implies that} \quad \kappa_{\theta(t)}>1.
$$
Nevertheless, choosing $t>\overline{\sigma}$ sufficiently large we ensure $\kappa_{\theta(t)}\leq {\overline{\sigma}}/{t}<1$. 
\end{rmk}}
{The corollary below is a rather direct} consequence of Theorem~\ref{Th1} and the closed form expression of the fixed point (\ref{def-fix-ricc-1}). A detailed proof is provided in Appendix \ref{cor-monge-maps-ind-proof} (on page~\pageref{cor-monge-maps-ind-proof}).

\begin{cor}\label{cor-monge-maps-ind}
 There exists some constant $c<\infty$ such that, for any $t>0$,
\begin{eqnarray}
  \Vert\SS(\theta(t))-\left(\overline{m},0,\overline{\sigma}\right)\Vert\vee\Vert r_{\theta(t)}-I\Vert&\leq& c/t, \quad \text{and}\label{monge-maps-ind}\\
  \Vert \overline{\SS}(\theta_1(t))-(m,0,\sigma)\Vert\vee \Vert \overline{r}_{\theta_1(t)}-I\Vert&\leq& c/t. \label{monge-maps-ind-2}
 \end{eqnarray}
 \end{cor}

Combining \eqref{ref-uv-infty-intro-eq} with the estimate \eqref{monge-maps-ind}, for any $x,y\in\RR^d$ we have
$$
\begin{array}{l}
\displaystyle e^{-\UU_{\theta(t)}(x)}~q_{\theta(t)}(x,y)~e^{-\VV_{\theta(t)}(y)}=\\
\\
\displaystyle e^{-U(x)}~
\frac{1}{\sqrt{\mbox{det}(2\pi\varsigma_{\theta(t)})}}~\exp{\left(-\frac{1}{2}~\Vert \varsigma_{\theta(t)}^{-1/2}
\left((y-\overline{m})-\kappa_{\theta(t)} (x-m)\right)\Vert_F^2\right)}
~\stackrel{t\rightarrow\infty}{\longrightarrow}~ e^{-U(x)}~e^{-V(y)},
\end{array}
$$
{i.e., the two marginal distributions become independent.}

{Similarly, with the regularized reference parameter $\theta(t)$,} condition \eqref{ref-uv-infty} takes the form
$$
\begin{array}{l}
\displaystyle \VV_{\theta(t)}(\overline{m})+\UU_{\theta(t)}(m)+\frac{1}{2}~\log{\mbox{\rm det}(
tI)}
\\
\\
\displaystyle=\frac{1}{2}~\log{\mbox{\rm det}(\sigma~\overline{\sigma})}-
\frac{1}{2}~\log{\mbox{\rm det}(\varsigma_{\theta(t)}^{-1}~\overline{\sigma})}-\frac{1}{2t}~\Vert m_0-\overline{m}\Vert_F^2,
\end{array}
$$
and using \eqref{connect-fp-v2} and Theorem~\ref{theo-s-pot} we see that
\begin{eqnarray*}
\nabla^2\VV_{\theta(t)}(y)&=&\overline{\sigma}^{-1}+\frac{1}{t}~\left(\beta~\sigma^{1/2}~\frac{\overline{r}_{\theta_1(t)}}{t}~\sigma^{1/2}~\beta^{\prime}-I\right)\stackrel{t\rightarrow\infty}{\longrightarrow}~\overline{\sigma}^{-1}, \quad\text{and}\\
\nabla^2\UU_{\theta(t)}(x)&=& \sigma^{-1}+\frac{1}{t}~
\beta^{\prime}~\left(\overline{\sigma}^{1/2}~\frac{r_{\theta(t)}}{t}~\overline{\sigma}^{1/2}-I\right)~\beta\stackrel{t\rightarrow\infty}{\longrightarrow}~\sigma^{-1}.
\end{eqnarray*}
{Moreover, we can obtain some explicit regularization rates, as shown by the proposition below.}  
\begin{prop}\label{est-ct-m-intro}
There exists some constant $c_0$ and some $t_0$ such that for any $t\geq t_0$  we have the estimate
$$
\begin{array}{l}
\displaystyle \Vert 2^{-1}\log{\mbox{\rm det}(
tI)}
+\VV_{\theta(t)}(\overline{m})+\UU_{\theta(t)}(m)-2^{-1}\log{\mbox{\rm det}(\sigma~\overline{\sigma})}\Vert\leq c_0/t.
\end{array}
$$
In addition, there exist some constants $c_1,c_2$ such that for any $t>0$ we have the estimates
\begin{eqnarray*}
\Vert \UU_{\theta(t)}(x+m)-\UU_{\theta(t)}(m)-2^{-1}x^{\prime}~\sigma^{-1}~x\Vert&\leq& c_1~\Vert x\Vert~(1+ \Vert x\Vert)/t, \quad \text{and}\\
\Vert \VV_{\theta(t)}(y+\overline{m})-\VV_{\theta(t)}(\overline{m})-2^{-1}y^{\prime}~\overline{\sigma}^{-1}~y\Vert&\leq& c_2~\Vert y\Vert~(1+ \Vert y\Vert)/t.
\end{eqnarray*}
\end{prop}
The proof of Proposition \ref{est-ct-m-intro} is provided in Appendix~\ref{sec-tech-proofs} (see page~\pageref{est-ct-m-intro-proof}). 

The estimates in \eqref{def-fix-t-large} yield, for any $t\geq 1$, the rather crude {lower} bounds
\begin{eqnarray*}
\nabla^2\VV_{\theta(t)}(y)&\succeq &\overline{\sigma}^{-1}+\frac{1}{t}~\left(\beta~\sigma^{1/2}~\frac{\left(I+\varpi_1^{-1}\right)^{-1}}{t}~\sigma^{1/2}~\beta^{\prime}-I\right), \quad \text{and}\\
\nabla^2\UU_{\theta(t)}(x)&\succeq & \sigma^{-1}+\frac{1}{t}~
\beta^{\prime}~\left(\overline{\sigma}^{1/2}~\frac{\left(I+\varpi^{-1}\right)^{-1}}{t}~\overline{\sigma}^{1/2}-I\right)~\beta,
\end{eqnarray*}
{which imply that} there exists some $t_0$ sufficiently large such that for any $t\geq t_0$ both potentials are strongly convex.{ The parameter associated with the Ornstein-Uhlenbeck diffusion discussed in Remark~\ref{ref-general-models}
is given by
$$
\theta[t]=\left(0,\beta[t],\tau[t]\right)\quad \mbox{\rm with}\quad \beta[t]=e^{tA}
\quad \mbox{\rm and}\quad
\tau[t]:=
\int_0^t~e^{sA}~\Sigma~e^{sA^{\prime}}~ds
$$
for some $\Sigma\in \Sa_d^+$ and some Hurwitz matrix 
$A$. In this context, there exist some $c_1,c_2>0$ such that for any $t\geq 0$ we have
$$
\Vert\sigma_{\beta[t]}\Vert=\Vert e^{tA}~\sigma~e^{tA^{\prime}}\Vert\leq c_1~e^{-c_{2}t}.
$$
Thus, for any $t_0$ there exists some constant $c_{3,t_0}>0$ such that, for any $t\geq t_0>0$,
$$
 \varpi_{\theta[t]}^{-1}=
  \overline{\sigma}^{1/2}~\tau[t]^{-1}~\sigma_{\beta[t]}~\tau[t]^{-1}\overline{\sigma}^{1/2}\Longrightarrow
  \Vert \varpi_{\theta[t]}^{-1}\Vert\leq c_{3,t_0}~e^{-c_2t}.
$$
Using (\ref{def-fix-ricc-1}), this yields, for any $t\geq t_0>0$, the estimate
$$
\Vert r_{\theta[t]}-I\Vert\leq c_{3,t_0}~e^{-c_2t}\quad \mbox{\rm and, therefore,}\quad
\Vert\varsigma_{\theta[t]}-\overline{\sigma}\Vert\vee \Vert\kappa_{\theta[t]}\Vert \leq c_{4,t_0}~e^{-c_2t}
$$
for some constant $c_{4,t_0}>0$. As a consequence, for any $t\geq t_0>0$ we obtain the exponential decays
$$
  \Vert\SS(\theta[t])-\left(\overline{m},0,\overline{\sigma}\right)\Vert\vee\Vert r_{\theta[t]}-I\Vert\leq c_{5,t_0}~e^{-c_2t}
$$
for some constant $c_{5,t_0}>0$.}

\subsubsection{Monge maps}
Using \eqref{rmk-theta-t-1} we find the identities
\begin{eqnarray*}
\overline{\sigma}^{1/2}\omega^{1/2}~\overline{\sigma}^{1/2}
&=&\sigma_{\beta}^{-1}~\sharp~\overline{\sigma}=\overline{\sigma}^{1/2}~ \left(\overline{\sigma}^{-1/2}~\sigma_{\beta}^{-1}~\overline{\sigma}^{-1/2}\right)^{1/2}~\overline{\sigma}^{1/2},\\
&=&\overline{\sigma}~\sharp~\sigma_{\beta}^{-1}=\sigma_{\beta}^{-1/2}~ \left(\sigma_{\beta}^{1/2}~\overline{\sigma}~\sigma_{\beta}^{1/2}\right)^{1/2}~\sigma_{\beta}^{-1/2},
\end{eqnarray*}
and also note that
$$
(\sigma_{\beta}^{-1}~\sharp~ \overline{\sigma})~\beta\sigma\beta^{\prime}
~(\sigma_{\beta}^{-1}~\sharp~ \overline{\sigma})=
(\sigma_{\beta}^{-1}~\sharp~ \overline{\sigma})~\sigma_{\beta}~(\sigma_{\beta}^{-1}~\sharp~ \overline{\sigma})=\overline{\sigma}.
$$
For any $t>0$, we also have the decompositions
$$
\frac{\varsigma_{\theta(t)}}{t}-(\sigma_{\beta}^{-1}~\sharp~ \overline{\sigma})
=\overline{\sigma}^{1/2}~\left(\frac{r_{\theta(t)}}{t}-\omega^{1/2}\right)~\overline{\sigma}^{1/2}
$$
and
$$
\kappa_{\theta(t)}-(\sigma_{\beta}^{-1}~\sharp~ \overline{\sigma})~\beta=\overline{\sigma}^{1/2}~\left(\frac{r_{\theta(t)}}{t}-\omega^{1/2}\right)~\overline{\sigma}^{1/2}~\beta.
$$

On the other hand, using \eqref{rmk-theta-t-1} we readily check that
$$
\frac{r_{\theta(t)}}{t}-\omega^{1/2}=\left(\omega+\left(\frac{t\omega}{2}\right)^2\right)^{1/2}-\omega^{1/2}-\frac{t\omega}{2},
$$
and the Ando-Hemmen inequality \eqref{square-root-key-estimate} readily yields
$$
\Vert  t^{-1}~  r_{\theta(t)}-\omega^{1/2}\Vert_2\leq \frac{t}{2}~\Vert \omega\Vert_2+\frac{t^2}{\lambda_{\text{\rm min}}(\omega)^{1/2}}~\left(\frac{\Vert \omega^2\Vert}{2}\right)
$$
and, therefore,
$$
\Vert  t^{-1}~  r_{\theta(t)}\Vert_2\leq \Vert \omega^{1/2}\Vert_2+\frac{t}{2}~\Vert \omega\Vert_2+\frac{t^2}{\lambda_{\text{\rm min}}(\omega)^{1/2}}~\left(\frac{\Vert \omega^2\Vert}{2}\right).
$$
The above estimates readily imply the following {regularization rates}.
\begin{cor}\label{cor-monge-maps}
 There exists some constant $c<\infty$ such that for any $t\in [0,1]$ we have
\begin{equation}\label{monge-maps}
\Vert
\kappa_{\theta(t)}-(\sigma_{\beta}^{-1}~\sharp~ \overline{\sigma})~\beta \Vert\vee
\Vert {\varsigma_{\theta(t)}}/{t}-(\sigma_{\beta}^{-1}~\sharp~ \overline{\sigma})\Vert\vee    \Vert r_{\theta(t)}/t-\omega^{1/2}\Vert\leq c~t.
 \end{equation}

 \end{cor}
 
Note that the limiting transport map from $\nu_{m,\sigma}$ to $\nu_{\overline{m},\overline{\sigma}}$ is given by
$$
\lim_{t\rightarrow 0}Z_{\SS(\theta(t))}(x)=
T_{\beta}(x):=\overline{m}+(\sigma_{\beta}^{-1}~\sharp~ \overline{\sigma})~\beta ~\left(x-m\right).
$$

 \begin{theo}\label{theo-entr-wass}
 For any $t>0$ we have
 \begin{equation}\label{decomp-ww}
 \begin{array}{l}
  \displaystyle   t~H\left(P_{\SS(\theta(t))}~|~P_{\theta(t)}\right)-\frac{1}{2}~\WW_2\left(\nu_{\overline{m},\overline{\sigma}},\nu_{(\alpha+\beta m),\sigma_{\beta}}\right)^2
\\
\\
=  \displaystyle 
\tr\left(\left((\overline{\sigma}~\sharp~\sigma_{\beta}^{-1})-\frac{\varsigma_{\theta(t)}}{t}\right)~\sigma_{\beta}\right)
+\frac{t}{2}~\left(d\log{(2\pi)}-\log{\mbox{det}\left(\frac{r_{\theta(t)}}{t}
\right)}\right).
\end{array}
\end{equation}
In addition, there exists some constant $c<\infty$ and some $t_0$ such that for any $0< t\leq t_0$ we have
 $$
\left\vert t~H\left(P_{\SS(\theta(t))}~|~P_{\theta(t)}\right)-\frac{1}{2}~\WW^2_2\left(\nu_{\overline{m},\overline{\sigma}},\nu_{(\alpha+\beta m),\sigma_{\beta}}\right)\right\vert\leq c~t
 $$
 with the rescaled relative entropy $H$ defined in (\ref{entropic-cost}).
 \end{theo}

The proof of Theorem \ref{theo-entr-wass} is provided in the Appendix~\ref{appendix-entro} (on page~\pageref{theo-entr-wass-proof}). The proof of the latter estimate in Theorem~\ref{theo-entr-wass} utilizes Corollary~\ref{cor-monge-maps}.

When  $\beta=I$ we recover the well known Monge map $\Ya_I$ between Gaussian distributions. In addition, when $(\alpha,\beta)=(0,I)$ we have
$$
T_I=\argmin_{T~:~(T\star \nu_{m,\sigma})=\nu_{\overline{m},\overline{\sigma}}} \WW_2(\nu_{m,\sigma},T\star \nu_{m,\sigma}).
$$
To the best of our knowledge, the formula and the non asymptotic estimates presented in Theorem~\ref{theo-entr-wass} for general Gaussian models are new.
{A related result can be found in} Theorem 1 in~\cite{genevay}, which provides quantitative bounds on the rate of convergence of  regularized optimal transport costs to standard optimal transport when the cost function $c(x,y)=-\log{q(x,y)}$ in (\ref{def-entropy-pb}) is Lipschitz and the measures  $(\eta,\mu)$  have bounded support.

\section{Sinkhorn algorithm}\label{sec-gauss-sinkhorn}

Consider some probability measures  $\eta$ and  $\mu$ on $\RR^d$  as well as some Markov transition $\Ka_0(x,dy)$ from $\RR^d$ into itself such that $\eta \Ka_0\simeq \mu$. 
The Sinkhorn iterations are defined sequentially for any $n\geq 0$ by a collection of probability distributions
\begin{equation}\label{def-Pa-n}
 \Pa_{2n}=\eta\times\Ka_{2n}
\quad \mbox{\rm and}\quad
 \Pa_{2n+1}=(\mu\times\Ka_{2n+1})^{\flat}
\end{equation}
starting from $\Pa_0$ at rank $n=0$.
{For $n \ge 1$, the  Markov transitions} $\Ka_n$ in (\ref{def-Pa-n}) {are defined} sequentially by the conditioning formulae
\begin{equation}
\left\{\begin{array}{l}
(\pi_{2n}\times\Ka_{2n+1})^{\flat}=\eta\times\Ka_{2n}\quad \mbox{\rm and}\quad
\pi_{2n+1}\times\Ka_{2(n+1)}=(\mu\times\Ka_{2n+1})^{\flat}\\
\\
\mbox{\rm with the distributions}\quad \pi_{2n}:=\eta \Ka_{2n}\quad\mbox{\rm and}\quad \pi_{2n+1}:=\mu \Ka_{2n+1}.
\end{array}\right.
\label{s-2}
\end{equation}
{The equivalence between \eqref{sinhorn-entropy-form} and the formulae \eqref{s-2} is rather well known \cite{peyre, nutz}. For completeness, a sketch of a proof is provided in Appendix~\ref{appendix-entro} (on page~\pageref{appendix-entro}), see also Section~\ref{sec-extended}.}

\subsection{Gaussian Sinkhorn equations}\label{sec-sinkhorn}

{For the linear Gaussian model where
\begin{equation}
(\eta,\mu):=(\nu_{m,\sigma},\nu_{\overline{m},\overline{\sigma}})\quad \mbox{\rm and}\quad \Ka_0:=K_{\theta_0}\quad \mbox{\rm with} \quad\theta_0=(\alpha,\beta,\tau)\in\Theta
\label{gauss-model}
\end{equation}
one readily obtains that
\begin{equation}
\pi_{0}:=\nu_{m,\sigma}\Ka_0=\nu_{m_{0},\sigma_{0}}\quad\mbox{\rm with}\quad
(m_0,\sigma_0):=h_{m,\sigma}(\theta_0).
\label{gauss-model-2}
\end{equation}
}
{Then, by  conjugacy arguments,} we also have
 \begin{equation}
 \left\{\begin{array}{l}
\pi_{n}=\nu_{m_{n},\sigma_{n}}
\quad \mbox{\rm and}\quad
  \Ka_{n}=K_{\theta_n}\\
  \\
  \mbox{\rm for some parameters}\quad (m_{n},\sigma_n)\in(\RR^d\times\Sa^+_d)\quad\mbox{\rm and}\quad
\theta_n=(\alpha_n,\beta_n,\tau_n)\in\Theta.
  \end{array}\right.
  \label{ref-mean-cov-intro}
\end{equation}
To identify the parameters $\theta_n$, first we use (\ref{s-2}) to verify that
$$
(m_{2n},\sigma_{2n})=h_{m,\sigma}(\theta_{2n})\quad \mbox{\rm and}\quad
(m_{2n+1},\sigma_{2n+1})=h_{\overline{m},\overline{\sigma}}(\theta_{2n+1}),
$$
with the functions $h_{m,\sigma}$ and $h_{\overline{m},\overline{\sigma}}$ defined in (\ref{def-P-theta}).
In terms of the probability measures $P_{\theta}$ and $\overline{P}_{\theta}$ defined in (\ref{def-P-theta}), the conjugate formula (\ref{ref-conjug}) applied to (\ref{s-2}) also shows that
\begin{equation}\label{rec-bb}
\left\{\begin{array}{l}
\Pa_{2n}=P_{\theta_{2n}}\quad \mbox{\rm and}\quad
\Pa_{2n+1}=\overline{P}^{\,\flat}_{\theta_{2n+1}},\\
\\
  \mbox{\rm with}\quad \theta_{2n+1}=\BB_{m,\sigma}(\theta_{2n})
\quad \mbox{\rm and}\quad
\theta_{2(n+1)}=\BB_{\overline{m},\overline{\sigma}}(\theta_{2n+1})
\end{array}\right.
\end{equation}
and the Bayes'  maps $\BB_{m,\sigma}$ and $\BB_{\overline{m},\overline{\sigma}}$ defined in (\ref{ref-conjug-Z-par}). 
This yields for any $n\geq 0$ the {mean values}
\begin{equation}\label{def-m-s}
m_{2n+1}=m+
\beta_{2n+1} (\overline{m}-m_{2n})
\quad\mbox{\rm and}\quad
m_{2(n+1)}=\overline{m}+\beta_{2(n+1)} ~(m-m_{2n+1}),
\end{equation}
{which are easily found} using the conjugate random map \eqref{ref-conjug-Z} and \eqref{s-2}. A more detailed description of these parameters and the corresponding random maps is provided in Appendix \ref{gauss-sinhorn-details}, on page~\pageref{gauss-sinhorn-details} (see for instance (\ref{beta-s-2-i}) and (\ref{beta-s-2-ii}) as well as (\ref{Xa-1}) and (\ref{Ya-1})). 

{The correction matrices $\beta_n$ are called gain matrices, in analogy to Kalman filtering theory. They allow to adjust the mean values of the target marginal measures. As in the Kalman filter, they are also used to sequentially adjust the covariances.}

\subsection{Riccati difference equations}\label{diff-ricc-sec}

Next technical lemma is pivotal: it provides a complete description of the gain matrices in terms of the reference parameter $\theta_0=(\alpha,\beta,\tau)$ and the flow of covariance matrices $\tau_n$. 

\begin{lem}\label{lem-pivot}
For any $n\geq 0$ we have
\begin{equation}\label{hom}
\beta_{2n}=\tau_{2n}~\tau^{-1}\beta\quad \mbox{and}\quad
\beta_{2n+1}=\tau_{2n+1}~\beta^{\prime}\tau^{-1}.
\end{equation}
\end{lem}
{The proof of Lemma \ref{lem-pivot} is rather technical; it is provided in Appendix \ref{lem-pivot-proof}, on page~\pageref{lem-pivot-proof}. Lemma \ref{lem-pivot} shows that the analysis of Sinkhorn algorithm reduces to that of the flow of covariance matrices $\tau_{n}$.}

The Bayes' map recursions in \eqref{rec-bb} show that formulae involving the ordered sequence $(\theta_n)_{n\geq 0}$ coincide with formulae involving the ordered sequence $(\theta_n)_{n\geq 1}$ by changing $(m,\sigma)$ by $(\overline{m},\overline{\sigma})$ (and the initial parameters $\theta_0$ by $\theta_1$). 

Our next objective is to show that the flow of covariance matrices $\tau_n$ can be computed offline by solving a time-homogeneous Riccati equation. {To this end, we first introduce a sequence of suitably rescaled matrices.}
\begin{defi}
Let $\upsilon_{n}$ be the rescaled covariance matrices defined for any $n\geq 0$ by
\begin{equation}
\upsilon_{2n}:=\overline{\sigma}^{-1/2}\tau_{2n}~\overline{\sigma}^{-1/2}
\quad\mbox{and}\quad
\upsilon_{2n+1}:=\sigma^{-1/2}\tau_{2n+1}\sigma^{-1/2}.
\label{resc-M}
\end{equation}
\end{defi}

{The theorem below is the second key result in this paper. It yields an offline description of the flow of covariance matrices of the Sinkhorn algorithm in terms of the dual Riccati maps} $\mbox{\rm Ricc}_{\varpi_{\theta}}$ and $\mbox{\rm Ricc}_{ \overline{\varpi}_{\theta_1}}$ defined in  (\ref{ricc-maps-def}), with the {positive-definite} matrices $(\varpi_{\theta}, \overline{\varpi}_{\theta_1})$ defined in (\ref{def-w-1}) and (\ref{def-wo-1}), {respectively}.

\begin{theo}\label{th-2}
For any $n\geq 0$, we have the recursions
\begin{equation}\label{k-ricc-0}
\upsilon_{2(n+1)}^{-1}=I+ \gamma_{\theta}~\upsilon_{2n+1}~  \gamma_{\theta}^{\prime}\quad \mbox{and}\quad
\upsilon_{2n+1}^{-1}=
I+  \gamma_{\theta}^{\prime}~\upsilon_{2n}~ \gamma_{\theta},
\end{equation}
{together with} the matrix Riccati difference equations
\begin{equation}\label{k-ricc}
\upsilon_{2(n+1)}=\mbox{\rm Ricc}_{\varpi_{\theta}}\left(\upsilon_{2n}\right)\quad
\mbox{and}\quad
\upsilon_{2n+1}=\mbox{\rm Ricc}_{ \overline{\varpi}_{\theta_1}}\left(\upsilon_{2n-1}\right).
\end{equation}
\end{theo}
The proof of Theorem~\ref{th-2}~is provided in Appendix~\ref{app-ricc} on page~\pageref{th-2-proof}. 
\begin{rmk}\label{one-d-cf}
Following Remark~\ref{rmq-one-dim-ricc}, the matrix difference equations (\ref{k-ricc}) coincide for one-dimensional models. In this context, Lemma 4.3 in~\cite{dh-23} provides closed-form solutions of Riccati difference equations. For instance, for even indices we find the equation
$$
(\upsilon_{2n}-r_{\theta})=(\upsilon_{0}-r_{\theta})~\frac{(\varpi_{\theta}+2r_{\theta})~\rho_{\theta}^n}{(\upsilon_{0}+\varpi_{\theta}+r_{\theta})(1-\rho_{\theta}^n)+(\varpi_{\theta}+2r_{\theta})~\rho_{\theta}^n}
$$
with the positive fixed point $r_{\theta}$ defined in (\ref{def-r-t}) and the exponential decay parameter
$$
\rho_{\theta}:=(1+r_{\theta}+\varpi_{\theta})^{-2}<1.
$$
\end{rmk}

The monotone properties of Riccati maps (see for instance (\ref{ricc-monotone}) in Appendix~\ref{app-ricc}) yield the following estimates {for the covariance matrices $\tau_n$ and $\sigma_n$, and the gain matrices $\beta_n$.}

\begin{cor}\label{cor-beta-est}
For any $n\geq 1$ we have the uniform estimates 
$$
\overline{\sigma}^{1/2}(I+\varpi_{\theta}^{-1})^{-1}\overline{\sigma}^{1/2}\preceq \tau_{2n}\preceq \overline{\sigma}\quad\mbox{and}\quad
\sigma^{1/2}(I+\overline{\varpi}_{\theta_1}^{-1})^{-1}\sigma^{1/2}\preceq \tau_{2n+1}\preceq \sigma.
$$
In addition, we have
$$
\Vert\beta_{2n}\Vert_2 \leq \Vert\overline{\sigma}\Vert_2~\Vert\tau^{-1}\beta\Vert_2\quad \mbox{and}\quad
\Vert\beta_{2n+1}\Vert_2 \leq \Vert\sigma\Vert_2~\Vert\tau^{-1}\beta\Vert_2,
$$
as well as
$$
\sigma_{2n}\succeq \overline{\sigma}^{1/2}(I+\varpi_{\theta}^{-1})^{-1}\overline{\sigma}^{1/2}\quad \mbox{and}\quad
\sigma_{2n+1}\succeq \sigma^{1/2}(I+\overline{\varpi}_{\theta_1}^{-1})^{-1}\sigma^{1/2}.
$$
\end{cor}
The last assertion in Corollary \ref{cor-beta-est} comes from the fact that
$$
\sigma_{2n}\succeq \tau_{2n}\quad \mbox{and}\quad
\sigma_{2n+1}\succeq \tau_{2n+1}.
$$
\begin{rmk}\label{rmk-stabi}
Following Remark~\ref{bridge-gain-unstable},
when $\lambda_{\text{\rm min}}(\tau)$ is sufficiently large or when $\Vert\beta\Vert_2$ is sufficiently small the gain matrices $\beta_n$ are stable. For instance,
$$
\Vert\tau^{-1}\Vert_2~
\Vert\beta\Vert_2<\Vert \sigma\Vert_2^{-1}\wedge\Vert \overline{\sigma}\Vert_2^{-1} \quad \text{implies} \quad \sup_{n\geq 1}\Vert\beta_{n}\Vert_2<1.
$$
Nevertheless, for one dimensional models with $\beta=1$, we have $\omega^{-1}:= 
 \overline{\sigma}\sigma$ and by Corollary~\ref{cor-beta-est} we obtain
$$
\beta_{2n}=\frac{\tau_{2n}}{t}\geq \frac{\overline{\sigma}}{t}~\left(1+\frac{\overline{\sigma}}{t}~\frac{\sigma}{t}\right)^{-1}.
$$
Choosing $\overline{\sigma}$ sufficiently large, we have
$$
\frac{\overline{\sigma}}{t}\left(1-\frac{\sigma}{t}\right)>1 \quad \text{which implies that} \quad
\beta_{2n}>1.
$$
\end{rmk}
A more detailed discussion on these inequalities {is provided in Appendix \ref{gauss-sinhorn-details} (see \eqref{ref-var-2n} and \eqref{ref-var-2n1}).} More refined estimates can be obtained using the monotone properties of Riccati maps. For instance, using \eqref{estimate-app-ricc-im}  we easily check the following result.
\begin{prop}\label{prop-ref-t}
For any $n\geq 2$ we have the uniform estimates
$$
\begin{array}{rclclcl}
 \overline{\sigma}^{-1} +\overline{\sigma}^{-1/2} (\varpi_{\theta}+I)^{-1}\overline{\sigma}^{-1/2}&\preceq &\tau_{2n}^{-1}&\preceq &
 \overline{\sigma}^{-1/2}\left(I+\varpi_{\theta}^{-1}\right)\overline{\sigma}^{-1/2},\\
  \sigma^{-1}+ \sigma^{-1/2}(\overline{\varpi}_{\theta_1}+I)^{-1}\sigma^{-1/2}&\preceq & \tau_{2n+1}^{-1}&\preceq &
  \sigma^{-1/2}\left(I+\overline{\varpi}_{\theta_1}^{-1}\right)\sigma^{-1/2}. 
\end{array}$$
\end{prop}

\subsection{A Gibbs loop process}\label{gibbs-loop-sec}
For any $n\geq 0$ we have the reversibility properties
\begin{equation}
\begin{array}[b]{rcl}
\eta\times\Ka_{2n}\times \Ka_{2n+1}&=&(\eta\times\Ka_{2n}\times \Ka_{2n+1})^{\flat}\\
\mu\times \Ka_{2n+1}\times\Ka_{2(n+1)}&=&(\mu\times \Ka_{2n+1}\times\Ka_{2(n+1)})^{\flat}
\end{array}\label{rev-s-1}
\end{equation}
which, in turn, yield the fixed-point equations
\begin{equation}\label{fixed-points-gibbs}
\left\{
    \begin{array}{rcl}
    \eta \Ka^{\circ}_{2n+1}&=&\eta\\
    \mu \Ka^{\circ}_{2(n+1)}&=&\mu\\
    \end{array}
\right\}
\quad\mbox{\rm with}\quad
 \Ka^{\circ}_{2n+1}:=\Ka_{2n}\Ka_{2n+1}\quad\text{and}\quad
\Ka^{\circ}_{2(n+1)}:=\Ka_{2n+1}\Ka_{2(n+1)}.
\end{equation}
{The kernels $\Ka^\circ_n$ can be used to construct, for $n \ge 1$, the Markov evolutions}
\begin{equation}\label{gibbs-tv}
\pi_{2n}=\pi_{2(n-1)}\Ka^{\circ}_{2n}\quad\mbox{\rm and}\quad
\pi_{2n+1}=\pi_{2n-1}\Ka^{\circ}_{2n+1}.
\end{equation}
The properties of the random maps associated with the Gibbs-type transitions $\Ka^{\circ}_n$ are discussed in Appendix~\ref{gauss-sinhorn-details}. 

The Markov evolution equations in \eqref{gibbs-tv} ensure the decreasing properties of the relative entropies~\cite{dm-03},
$$
\mbox{\rm Ent}\left(\mu~|~\pi_{2n}\right)=
\mbox{\rm Ent}\left(\mu \Ka^{\circ}_{2n}~|~\pi_{2(n-1)}\Ka^{\circ}_{2n}\right)\leq
\mbox{\rm Ent}\left(\mu~|~\pi_{2(n-1)}\right)
$$
and, in the same vein,
$$
\mbox{\rm Ent}\left(\eta~|~\pi_{2n+1}\right)\leq
\mbox{\rm Ent}\left(\eta~|~\pi_{2n-1}\right).
$$
On the other hand, using (\ref{s-2}) for any $p<q$ we readily see that
\begin{eqnarray*}
\frac{d\Pa_{2q}}{d\Pa_{2p}}(x,y)&=&
\left[\prod_{p\leq l< q}\frac{d\Pa_{2l+1}}{d\Pa_{2l}}(x,y)\right]~\left[\prod_{p\leq l< q}\frac{d\Pa_{2(l+1)}}{d\Pa_{2l+1}}(x,y)\right]\\
&=&\left[\prod_{p\leq l< q}\frac{d\mu}{d\pi_{2l}}(y)\right]~\left[\prod_{p\leq l< q}\frac{d\eta}{d\pi_{2l+1}}(x)\right].
\end{eqnarray*}
Thus, for any given $\Pa\,\in\, \Ca(\eta,\mu)$ and $q>p$ we have the decomposition
\begin{eqnarray}
\mbox{\rm Ent}(\Pa~|~\Pa_{2p})&\geq&\mbox{\rm Ent}(\Pa~|~\Pa_{2p})-\mbox{\rm Ent}(\Pa~|~\Pa_{2q})\nonumber\\
&=& \Pa\left(\log{\frac{d\Pa_{2q}}{d\Pa_{2p}}}\right)=\sum_{p\leq l< q}(\mbox{\rm Ent}\left(\mu~|~\pi_{2l}\right)+\mbox{\rm Ent}\left(\eta~|~\pi_{2l+1}\right))\label{pyth}\\
&\geq&(q-p)~(\mbox{\rm Ent}\left(\mu~|~\pi_{2(q-1)}\right)+\mbox{\rm Ent}\left(\eta~|~\pi_{2q-1}\right))\nonumber
\end{eqnarray}
and choosing $p=0$ in \eqref{pyth} yields the following theorem.
\begin{theo}\label{th-gibbs-loop-entrop}
Assume there exists some $\Pa\in \Ca(\eta,\mu)$ such that 
$\mbox{\rm Ent}(\Pa~|~\Pa_0)<\infty$. Then, 
for any $n\geq 1$ we have
\begin{eqnarray*}
 \mbox{\rm Ent}\left(\mu~|~\pi_{2n}\right)\vee\mbox{\rm Ent}\left(\eta~|~\pi_{2n+1}\right) \leq \frac{1}{n}~\mbox{\rm Ent}(\Pa~|~\Pa_0)
\end{eqnarray*}
and, in addition,
$$
\lim_{n\rightarrow \infty}n~ \mbox{\rm Ent}\left(\mu~|~\pi_{2n}\right)=
\lim_{n\rightarrow \infty} n~\mbox{\rm Ent}\left(\eta~|~\pi_{2n+1}\right)=0.
$$
\end{theo}

The last assertion is a direct consequence of the convergence of the series (\ref{pyth}). For instance, for any $\epsilon>0$ there exists some $n_{\epsilon}\geq 1$ such that for every
$n\geq n_{\epsilon}$ we have
$$
n~\mbox{\rm Ent}(\eta~|~\pi_{2n+1})\leq 
\sum_{n\leq p\leq  2n}\mbox{\rm Ent}(\eta~|~\pi_{2p+1})\leq \epsilon.
$$
Also by \eqref{pyth}, for any $p<q$ and  $\Pa\,\in\, \Ca(\eta,\mu)$  we have the monotone properties
\begin{eqnarray*}
\mbox{\rm Ent}(\Pa~|~\Pa_{2q})&=&\mbox{\rm Ent}(\Pa~|~\Pa_{2p})-\sum_{p\leq l< q}(\mbox{\rm Ent}\left(\mu~|~\pi_{2l}\right)+\mbox{\rm Ent}\left(\eta~|~\pi_{2l+1}\right))\\
&=&\mbox{\rm Ent}(\Pa~|~\Pa_{2q-1})-\mbox{\rm Ent}\left(\eta~|~\pi_{2q-1}\right).
 \end{eqnarray*}
The above formulae are not new, they are sometimes called the Pythagorean law for the relative entropy~\cite{csiszar-2,ruschen}  (see also Proposition 6.5 in~\cite{nutz}).
{They show that the sequence $\mbox{\rm Ent}\left(\Pa~|~\Pa_{n}\right)$ is decreasing} and we have
\begin{eqnarray*}
\lim_{n\rightarrow\infty}\mbox{\rm Ent}\left(\Pa~|~\Pa_{n}\right)&=&\inf_{n\geq 0}\mbox{\rm Ent}\left(\Pa~|~\Pa_{n}\right)\\
&=&\mbox{\rm Ent}(\Pa~|~\Pa_{0})-\sum_{p\geq 0}\left(\mbox{\rm Ent}\left(\mu~|~\pi_{2p}\right)+\mbox{\rm Ent}\left(\eta~|~\pi_{2p+1}\right)\right).
\end{eqnarray*}

Sublinear rates have been developed in the articles~\cite{alts-2017,chak-2018,dvu-2017}.
In the context of finite state spaces, the above linear rates can be deduced from the exponential rates presented in the pioneering article by Fienberg~\cite{fienberg-1970} using Hilbert projective metrics, further developed by Franklin and Lorenz in~\cite{franklin-1989}. The extension of these Hilbert projective techniques to general compact space models are developed in~\cite{chen-2016}.
Linear rates with the robust constant $\mbox{\rm Ent}(\Pa~|~\Pa_0)$ and $\pi$ solving the minimum entropy problem on non-necessarily compact spaces were first obtained by L\'eger in~\cite{leger} using elegant gradient descent and Bregman divergence
 techniques, see also the recent articles~\cite{doucet-bortoli,karimi-2024}. A refined convergence rate at least one order faster has also been developed in~\cite{promit-2022}.

For Gaussian models of the form in \eqref{ref-mean-cov-intro}, the mean and covariance parameters of the Gaussian distributions $\pi_{n}=\nu_{m_{n},\sigma_{n}}$ are computed sequentially.
By (\ref{def-m-s}), for any $n\geq 1$ we have
\begin{eqnarray}
m_{2n}-\overline{m}&=&\beta^{\circ}_{2n} ~(m_{2(n-1)}-\overline{m})
\quad\mbox{\rm with}\quad
\beta^{\circ}_{2n}:=\beta_{2n}\beta_{2n-1}, \quad \text{and}\label{ref-m-intro}\\
m_{2n+1}-m&=&
\beta_{2n+1}^{\circ} (m_{2n-1}-m)\quad\mbox{\rm with}\quad\beta_{2n+1}^{\circ}:=\beta_{2n+1}\beta_{2n}.\label{ref-m-o-intro}
\end{eqnarray}
Consider the directed matrix products
\begin{equation}\label{directed-prod}
\beta^{\circ}_{2n,0}:=\beta^{\circ}_{2n}\beta^{\circ}_{2(n-1)}\ldots \beta^{\circ}_{2}\quad
\quad\mbox{\rm and}\quad
\beta^{\circ}_{2n+1,1}:=\beta^{\circ}_{2n+1}\beta^{\circ}_{2n-1}\ldots \beta^{\circ}_{3}.
\end{equation}
In this notation, we have
\begin{equation}\label{directed-prod-sig}
\sigma_{2n}-\overline{\sigma}=\beta^{\circ}_{2n,0}(\sigma_0-\overline{\sigma})\left(\beta^{\circ}_{2n,0}\right)^{\prime}
\quad\mbox{\rm and}\quad\sigma_{2n+1}-\sigma=\beta^{\circ}_{2n+1,1}(\sigma_1-\sigma)\left(\beta^{\circ}_{2n+1,1}\right)^{\prime}.
\end{equation}
The proof of the above covariance formulae is provided in Appendix \ref{directed-prod-sig-proof} (on page~\pageref{directed-prod-sig-proof}).

{We finish this section with a technical lemma that is key to the construction of quantitative estimates in Section \ref{quant-sec}.} It yields a description of the gain matrices $\beta^{\circ}_{n}$ of the Gibbs-loop process in terms of matrix Riccati difference equations \eqref{k-ricc}. 

\begin{lem}\label{prop-beta-o}
For any $n\geq 0$ we have
\begin{eqnarray}
\overline{\sigma}^{-1/2}~\beta^{\circ}_{2(n+1)}~\overline{\sigma}^{1/2}&=&\overline{\gamma}_{\theta_1}^{\prime}~\left(\overline{\varpi}_{\theta_1}^{-1}+\upsilon_{2n+1}^{-1}
\right)^{-1}~\overline{\gamma}_{\theta_1}~{=I-\upsilon_{2(n+1)}} \quad \text{and}
\label{beta-2-tau-1}\\
\sigma^{-1/2}~\beta_{2n+1}^{\circ}~\sigma^{1/2}
&=& \gamma_{\theta}^{\prime}~\left( \varpi_{\theta}^{-1}+\upsilon_{2n}^{-1}\right)^{-1}~ \gamma_{\theta}~{=I-\upsilon_{2n+1}},
\label{beta-2-tau-2}
\end{eqnarray}
with the matrices $ (\varpi_{\theta}, \gamma_{\theta})$ and $(\overline{\varpi}_{\theta_1},\overline{\gamma}_{\theta_1})$ defined in (\ref{def-w-1}) and (\ref{def-wo-1}).
\end{lem}
The proof is provided in Appendix \ref{prop-beta-o-proof}, on page~\pageref{prop-beta-o-proof}.

\section{Quantitative estimates}\label{quant-sec}

\subsection{An exponential stability theorem}
The exponential stability properties of the matrix Riccati difference equations
in \eqref{k-ricc} are well understood. 
For instance, the Ando-Hemmen inequality (\ref{square-root-key-estimate}) and
the stability estimates  stated in Proposition~\ref{prop-cv-app}  readily yield the following estimates.

\begin{theo}\label{theo-qs}
There exists some  $c_{\theta}<\infty$ such that for any $n\geq 0$ we have
\begin{eqnarray*}
\Vert \tau_{2n}-\varsigma_{\theta}\Vert\vee \Vert  \tau_{2n}^{1/2}-\varsigma_{\theta}^{1/2}\Vert\vee \Vert \beta_{2n}-\kappa_{\theta}\Vert&\leq &c_{\theta}~\rho_{\theta}^n~
\Vert \tau_{0}-\varsigma_{\theta}\Vert
\end{eqnarray*}
with the parameter
\begin{equation}\label{def-rho-t}
\rho_{\theta}:=(1+\lambda_{\text{\rm min}}(r_{\theta}+\varpi_{\theta}))^{-2}<1.
\end{equation}

\end{theo}

The recursions in Theorem \ref{th-2} (see the formulae in expression \eqref{k-ricc-0}) also show that the fixed point matrices $(r_{\theta},\overline{r}_{\theta_1})$ and their rescaled versions defined in \eqref{def-Sa} and \eqref{def-over-Sa} are connected by the formulae
\begin{equation}\label{def-cchi}
\varsigma_{\theta}^{-1}=\overline{\sigma}^{-1}+\cchi_{\theta}~\overline{\varsigma}_{\theta_1}~\cchi_{\theta}^{\prime}~\quad \mbox{and}\quad
\overline{\varsigma}_{\theta_1}^{-1}=\sigma^{-1}
+\cchi_{\theta}^{\prime}~~\varsigma_{\theta}~\cchi_{\theta}
\end{equation}
with the parameter $\cchi_{\theta}$ defined in (\ref{def-w-1}).

Lemma~\ref{prop-beta-o} expresses the matrices $\beta^{\circ}_{n}$ in terms of $\upsilon_{n}$. {The stability properties of these Riccati matrices are discussed in Appendix~\ref{app-ricc} (see for instance (\ref{v2u}) as well as (\ref{En-prod-uv}) and Theorem~\ref{theo-floquet}). There exists some  $c_{\theta}<\infty$ such that for any $n\geq 1$ we have} the inequality
\begin{equation}\label{def-rho-1}
\Vert\overline{\sigma}^{-1/2}~\beta^{\circ}_{2n,0}~\overline{\sigma}^{1/2}\Vert\leq  c_{\theta}~\rho^{n/2}_{\theta},
\end{equation}
which is a consequence of Lemma~\ref{prop-beta-o}. A detailed proof is provided in Appendix \ref{gauss-sinhorn-details} (see page~\pageref{def-rho-12-proof}).
\begin{rmk}In contrast with the possible instability properties of the gain matrices $\beta_{2n}$ discussed in Remark~\ref{rmk-stabi}, the matrix product semigroup $\beta^{\circ}_{2n,0}$ is stable for any values of the parameters $(\tau,\beta)$. For instance, the exponential decay  estimates (\ref{def-rho-1}) apply to the linear diffusions discussed in Remark~\ref{ref-general-models} for non necessarily stable drift matrices $A_t$.\end{rmk}

\begin{rmk}
Matrices  $(\tau_{2n+1}, \beta_{2n+1})$ as well as $\sigma^{-1/2}\beta_{2n+1,1}^{\circ}\sigma^{1/2}$ satisfy the same inequalities as in Theorem~\ref{theo-qs} and in expression \eqref{def-rho-1} for some parameter {$\overline{\rho}_{\theta_1}$}. These inequalities (and the {parameter $\overline{\rho}_{\theta_1}$}) are defined as above by replacing the parameters $(\varsigma_{\theta},\kappa_{\theta},r_{\theta},\varpi_{\theta})$
by $(\overline{\varsigma}_{\theta_1},\overline{\kappa}_{\theta_1},\overline{r}_{\theta_1},\overline{\varpi}_{\theta_1})$. For instance, we have
$$
\overline{\rho}_{\theta_1}:=(1+\lambda_{\text{\rm min}}(\overline{r}_{\theta_1}+\overline{\varpi}_{\theta_1}))^{-2}<1
\quad\mbox{
with $(\overline{r}_{\theta_1},\overline{\varpi}_{\theta_1})$ defined in (\ref{def-wo-1}) and (\ref{def-over-r}).}
$$
\end{rmk}

This yields the following corollary.
\begin{cor}\label{theo-cor-qs} 
There exists some  $c_{\theta}<\infty$such that for any $n\geq 1$ we have the exponential estimates
$$
\Vert m_{2n}-\overline{m}\Vert\leq  c_{\theta}~\rho_{\theta}^{n/2}~\Vert m_{0}-\overline{m}\Vert\quad
\mbox{and}\quad
\Vert \sigma_{2n}-\overline{\sigma}\Vert\leq  c_{\theta}~\rho_{\theta}^{n}~\Vert \sigma_{0}-\overline{\sigma}\Vert.
$$
Parameters $(m_{2n-1}, \sigma_{2n-1})$ with odd indices satisfy the same inequalities as above
by replacing $(\overline{m},\overline{\sigma},\rho_{\theta})$ by $(m,\sigma,\overline{\rho}_{\theta_1})$ and the initial parameters $(m_{0},\sigma_{0})$
by $(m_{1},\sigma_{1})$.
\end{cor}

{ 
\begin{rmk}\label{ref-rmk-en-marg}
Using the Gaussian entropy formula \eqref{KL-def} and the estimates stated in Corollary~\ref{cor-beta-est} and Lemma~\ref{lem-detlog} there exists some $n_0\geq 1$ and some constant $c_{\theta}>0$ such that for any $n\geq n_0$ we have
\begin{eqnarray*}
\mbox{\rm Ent}\left(\nu_{\overline{m},\overline{\sigma}}~|~\nu_{m_{2n},\sigma_{2n}}\right)\vee
\mbox{\rm Ent}\left(\nu_{m_{2n},\sigma_{2n}}~|~\nu_{\overline{m},\overline{\sigma}}\right)&\leq& c_{\theta}~\rho_{\theta}^{n} \quad \text{and}
\\
\mbox{\rm Ent}\left(\nu_{m,\sigma}~|~\nu_{m_{2n+1},\sigma_{2n+1}}\right)\vee
\mbox{\rm Ent}\left(\nu_{m_{2n+1},\sigma_{2n+1}}~|~\nu_{m,\sigma}\right)&\leq& c_{\theta}~\overline{\rho}_{\theta_1}^{n}.
\end{eqnarray*}
\end{rmk} }

Applying Corollary~\ref{cor-beta-est} and Theorem~\ref{theo-qs}, there exists some  $c_{\theta}$  such that, for any $n\geq 1$, the equality
$$
\alpha_{2n}-\iota_{\theta}=(\kappa_{\theta}-\beta_{2n})~m+\beta_{2n}~(m-m_{2n-1})
$$
implies that
$$
\Vert \alpha_{2n}-\iota_{\theta}\Vert\leq c_{\theta}~\rho_{\theta}^n~
\Vert \tau_{0}-\varsigma_{\theta}\Vert+\
c_{\theta}~\overline{\rho}_{\theta_1}^{n/2}~\Vert m_0-\overline{m} \Vert.
$$
Combining the above estimates with Theorem~\ref{theo-qs} we readily obtain the following result.
\begin{cor}\label{cor-theta-even-cv}
There exists some  $c_{\theta}$ and $c_{1,\theta}<\infty$ such that
for any $n\geq 1$ we have the exponential estimates
$$
\Vert\theta_{2n}-\SS(\theta)\Vert\leq 
c_{\theta}~\rho_{\theta}^n~
\Vert \tau_{0}-\varsigma_{\theta}\Vert+\
c_{1,\theta}~\overline{\rho}_{\theta_1}^{n/2}~\Vert m_0-\overline{m} \Vert,
$$
where $\SS$ stands for the Schr\"odinger bridge map defined in (\ref{def-Sa}).
\end{cor}

\subsection{Relative entropy estimates}
Theorem~\ref{theo-qs} and the estimates stated in Corollary \ref{theo-cor-qs}  can be used to derive a variety of quantitative estimates. For instance, we have the relative entropy formula
\begin{equation}\label{ent-intro-even}
\mbox{\rm Ent}\left(P_{\theta_{2n}}~|~P_{\SS(\theta)}\right) 
=  \frac{1}{2}~D(\tau_{2n}~|~\varsigma_{\theta})+\Vert\varsigma_{\theta}^{-1/2}\left(m_{2n}-\overline{m}\right)\Vert_F^2+\frac{1}{2}~\Vert\varsigma_{\theta}^{-1/2}
(\beta_{2n}-\kappa_{\theta})~\sigma^{1/2}\Vert_F^2
\end{equation}
with the Burg distance $D$ defined in (\ref{burg-def}) and the Schr\"odinger bridge map $\SS$ defined in (\ref{def-Sa}). The above formula is a direct consequence of (\ref{BL-intro-def}). A detailed proof is provided in Appendix \ref{appendix-entro} (on page~\pageref{ent-intro-even-proof}).

Choose $n_0\geq 1$ such that
$$
\rho_{\theta}^{n_0}~
\Vert \tau_{0}-\varsigma_{\theta}\Vert_F
\leq 1\wedge \frac{1}{2c_{\theta} \Vert \varsigma_{\theta}^{-1}\Vert_F},
$$
{where the constant $c_\theta$ and the parameter $\rho_\theta$ are the same as in Theorem \ref{theo-qs}. The following estimates can be readily obtained.}

\begin{cor}\label{cor-entrop-cv}
There exists some finite $c_{\theta}<\infty$ such that for any $n\geq n_0$ we have the entropy estimates
$$
\mbox{\rm Ent}\left(P_{\theta_{2n}}~|~P_{\SS(\theta)}\right) \leq c_{\theta}~\rho_{\theta}^n~\left(
\Vert \tau_{0}-\varsigma_{\theta}\Vert+\Vert m_{0}-\overline{m}\Vert^2\right).
$$
\end{cor}
Applying Pinsker's inequality we also easily deduce the total variation estimate 
$$
\Vert P_{\theta_{2n}}-P_{\SS(\theta)}\Vert_{\text{tv}}\leq 
c^{1/2}_{\theta}~\rho_{\theta}^{n/2}~\left(
\Vert \tau_{0}-\varsigma_{\theta}\Vert^{1/2}+\Vert m_{0}-\overline{m}\Vert\right).
$$

In terms of the random maps (\ref{random-Z-map})  and  (\ref{bridge-map-i}) we have
\begin{eqnarray*}
Z_{\theta_{2n}}(x)-Z_{\SS(\theta)}(x)&=&(m_{2n}-\overline{m})+(\beta_{2n}-\kappa_{\theta})(x-m)+(\tau_{2n}^{1/2}-\varsigma_{\theta}^{1/2})~G,
\end{eqnarray*}
which yields, for any $p\geq 1$, the Wasserstein distance estimate
$$
\WW_p\left(P_{\theta_{2n}},P_{\SS(\theta)}\right)
\leq \left(
\Vert m_{2n}-\overline{m}\Vert_F+e_d(p)~\Vert \sigma^{1/2}\Vert_F~\Vert \beta_{2n}-\kappa_{\theta}\Vert_F\right)+e_{d}(p)~\Vert\tau_{2n}^{1/2}-\varsigma_{\theta}^{1/2}\Vert_F
$$
with the parameter 
$$
e_d(p):=\EE\left(\Vert G\Vert_F^p\right)^{1/p}.
$$
Applying Theorem~\ref{theo-qs} and Corollary~\ref{theo-cor-qs}  to the Frobenius norm, we readily prove the following estimates.
\begin{cor}\label{cor-was-sinkhorn}
 For any $p\geq 1$ there exists some finite $c_{1,\theta}(p),c_{2,\theta}<\infty$ such that for any $n\geq 0$ we have
$$
\begin{array}{l}
\WW_p\left(P_{\theta_{2n}},P_{\SS(\theta)}\right)
\leq 
c_{1,\theta}(p)~\rho_{\theta}^n~
\Vert \tau_{0}-\varsigma_{\theta}\Vert_F+c_{2,\theta}~\rho_{\theta}^{n/2}~\Vert m_{0}-\overline{m}\Vert_F
\end{array}
$$
\end{cor}
{Exactly the same analysis can be applied to the random transport maps, to arrive at the equation} 
\begin{eqnarray*}
Z_{\theta_{2n+1}}(y)-Z_{\overline{\SS}(\theta_1)}(y)&=&
(m_{2n+1}-m)+(\beta_{2n+1}-\overline{\kappa}_{\theta_1})~(y-\overline{m})+(\tau_{2n+1}^{1/2}-\overline{\varsigma}_{\theta_1}^{1/2})~G
\end{eqnarray*}
with the function $\overline{\SS}$ defined in (\ref{over-S-def}).

\subsection{Regularization effects}\label{sinkhorn-reg-effects}

Denote by $\upsilon_n(t)$ the solution of the matrix Riccati difference equations (\ref{k-ricc}) associated with the parameter $\theta(t)$ defined in (\ref{theta-t-def}). In this case, we have 
$$
 \upsilon_0(t)=t~\overline{\sigma}^{-1}
    \quad\mbox{\rm and}\quad \tau_0(t)=t~I.
$$

Let $\theta_n(t)=(\alpha_n(t),\beta_n(t),\tau_n(t))$ be the flow of  Sinkhorn parameters associated with the initial 
parameter $\theta_0=\theta(t)$. In this notation, for any $n\geq 0$ we have
$$
\tau_{2n}(t):=\overline{\sigma}^{1/2}~\upsilon_{2n}(t)~\overline{\sigma}^{1/2}
\quad\mbox{and}\quad
\tau_{2n+1}(t):=\sigma^{1/2}~\upsilon_{2n+1}(t)~\sigma^{1/2}.
$$
Combining (\ref{rmk-theta-t-1}) with Proposition~\ref{prop-ref-t} for any $n\geq 2$ we readily obtain the estimates
$$
\begin{array}{rclclcl}
 \overline{\sigma}^{-1} +\overline{\sigma}^{-1/2} (t^2\omega+I)^{-1}\overline{\sigma}^{-1/2}&\preceq &\tau_{2n}(t)^{-1}&\preceq &
 \overline{\sigma}^{-1/2}\left(I+t^{-2}\omega^{-1}\right)\overline{\sigma}^{-1/2} \quad \text{and}\\
  \sigma^{-1}+ \sigma^{-1/2}(t^2\omega_1+I)^{-1}\sigma^{-1/2}&\preceq & \tau_{2n+1}(t)^{-1}&\preceq &
  \sigma^{-1/2}\left(I+t^{-2}\omega_1^{-1}\right)\sigma^{-1/2}, 
\end{array}$$
{with the matrices $(  \omega,  \omega_1)$ defined in (\ref{rmk-theta-t-1}).}

Equivalently, {following \eqref{estimate-app-ricc-2} in Lemma \ref{lem-estimate-app-ricc-2}} we also have  the following result.
\begin{prop}\label{prop-t-infty}
For any $t\geq 0$ and $n\geq 2$ we have the uniform estimates
$$
\begin{array}{rclclcl}
 \overline{\sigma}^{1/2}(I+(t^2\omega+I))^{-1} \overline{\sigma}^{1/2}&\preceq&   \overline{\sigma}-\tau_{2n}(t)&\preceq& \overline{\sigma}^{1/2}(I+t^2\omega)^{-1}
\overline{\sigma}^{1/2} \quad \text{and}\\
\sigma^{1/2} (I+(t^2\omega_1+I))^{-1} \sigma^{1/2} &\preceq& \sigma- \tau_{2n+1}(t)&\preceq& \sigma^{1/2} (I+t^2\omega_{1})^{-1}\sigma^{1/2}.
\end{array}$$
\end{prop}
This shows that the flow of covariance matrices $(\tau_{2n}(t), \tau_{2n+1}(t))$ converges towards $(\overline{\sigma},\sigma)$ as $t\rightarrow\infty$ uniformly w.r.t. the parameter $n\geq 0$.

Combining (\ref{hom}) with Proposition~\ref{prop-t-infty} we also see that
$$
\Vert \beta_{2n}(t)\Vert\leq \frac{1}{t}~\Vert\beta\Vert~\Vert \overline{\sigma}\Vert\quad \mbox{\rm and}\quad
\Vert \beta_{2n+1}(t)\Vert\leq \frac{1}{t}~\Vert\beta\Vert~\Vert \sigma\Vert.
$$

In this context, we also have
\begin{equation} \label{eq_tw_est}
t~\omega^{1/2}\preceq 
r_{\theta(t)}+  \varpi_{\theta(t)}=t~\left(
\frac{t\omega}{2}+\left(\omega+\left(\frac{t\omega}{2}\right)^2\right)^{1/2}\right)
\end{equation}
As expected {from (\ref{def-rho-t})}, $\rho_{\theta(t)} \to 0$ as $t \to \infty$, while $\rho_{\theta(t)} \to 1$ as $t\to 0$. We also have the exponential estimate
$$
\rho_{\theta(t)}\leq (1+t~\lambda_{\text{\rm min}}(\omega)^{1/2})^{-2}\leq \exp{\left(-2t\lambda_{\text{\rm min}}(\omega)^{1/2}\right)}.
$$
The estimation constant $c_{\theta(t)}$ in Theorem~\ref{theo-qs} {can be estimated using Propositions \ref{prop-Ea-u} and \ref{prop-cv-app} in Appendix \ref{app-ricc} (see Eqs. \eqref{def-psi} and \eqref{cv-ricc}, respectively). We also note that in the case $\beta=I$ we readily have
\begin{equation}
r_{\theta(t)}+  \varpi_{\theta(t)}=t~\left(
\frac{t\omega}{2}+\left(\omega+\left(\frac{t\omega}{2}\right)^2\right)^{1/2}\right),
\nonumber
\end{equation}
where 
$
\omega:= 
 \overline{\sigma}^{-1/2}~\sigma^{-1}~\overline{\sigma}^{-1/2},
$ 
and the identity \eqref{eq_tw_est} yields the lower bound
$$
r_{\theta(t)}+  \varpi_{\theta(t)} \succ
t~\left(\sigma^{-1/2}~ \overline{\sigma}^{-1}~\sigma^{-1/2}\right)^{1/2}
$$
and, in turn, the (simpler) estimate
\begin{equation}
\rho_{\theta(t)}=(1+\lambda_{\text{\rm min}}(r_{\theta(t)}+\varpi_{\theta(t)}))^{-2}< \left(1+t~\lambda_{\text{\rm min}}\left(\sigma^{-1/2}~ \overline{\sigma}^{-1}~\sigma^{-1/2}\right)^{1/2}\right)^{-2}.
\label{ref-cc-24}
\end{equation}
}
\begin{rmk}\label{comp-rmk}
In the special case of one-dimensional Gaussian models, the r.h.s. estimate in (\ref{ref-cc-24}) is the square of the entropy exponential decay rate presented in Proposition 1.3, part (i), of Ref.~\cite{chiarini} for general log-concave marginal models and sufficiently small values of $t$. As shown above, we also have $\rho_{\theta(t)} \to 0$ as $t \to \infty$. When applied to Gaussian models, for large values of $t$, the estimate (i) stated in Theorem 1.2 of Ref.~\cite{chiarini} also yields an entropy exponential decay rate of $1/2$.
\end{rmk}

\section{Schr\"odinger potential functions}\label{sec-schrod}

\subsection{{Integral recursive formulations}}

The iterative proportional fitting procedure can be defined in terms of Schr\"odinger potential functions as long as $\eta$ and $\mu$ are defined by some Gibbs measures, namely,
$$
\eta(dx)=e^{-U(x)}~dx\quad \mbox{\rm and}\quad\mu(dy)=e^{-V(y)}dy
$$
for some potential functions $U$ and $V$ on $\RR^d$. We also assume that
the Markov transition $\Ka_0(x,dy)$ is defined by a positive operator $Q(x,dy)$ with density $q(x,y)$, in the sense that
$$
\Ka_0(x,dy)=Q(x,dy):=q(x,y)~dy$$
and we set
$$
R(y,dx):=r(y,x)~dx\quad\mbox{\rm with}\quad q(x,y)=r(y,x).
$$
These conditions are clearly met for the Gaussian model (\ref{gauss-model}) with
\begin{eqnarray}
q(x,y)&=&r(y,x)=g_{\tau}(y-(\alpha+\beta x)) \quad \text{and}\nonumber\\
U(x)&=&-\log{g_{\sigma}(x-m)}
\quad \mbox{\rm and}\quad V(y):=-\log{g_{\overline{\sigma}}(y-\overline{m})}.
\label{gauss-model-6-1}
\end{eqnarray}

\begin{prop}\label{prop-schp}
For {every} $n\geq 0$ we have {the distributions $\Pa_n$ described in} \eqref{sinhorn-entropy-form-Sch} with the initial potential functions $(U_0,V_0)=(U,0)$ and the recursions
\begin{eqnarray*}
U_{2n+1}&=&~U_{2n}~:=U+\log{Q(e^{-V_{2n}})}, \text{and}\\
V_{2(n+1)}&=&V_{2n+1}:=V+\log{R(e^{-U_{2n+1}})}.
\end{eqnarray*}
Equivalently, we have
$$
\Ka_{2n}(x,dy)=\frac{Q(x,dy)e^{-V_{2n}(y)}}{Q(e^{-V_{2n}})(x)}\quad \mbox{and}\quad
\Ka_{2n+1}(y,dx)=\frac{R(y,dx)~e^{-U_{2n+1}(x)}}{R(e^{-U_{2n+1}})(y)}.
$$
\end{prop}
The above proposition is rather well known. For completeness, we provide a detailed proof in Appendix \ref{sch-appendix} (see page~\pageref{prop-schp-proof}).

In the literature, Schr\" odinger potential functions are sometimes written in terms of the functions
$$
\Ua_{n}=U_{n}-U\quad \mbox{\rm and}\quad
\Va_{n}=V_{n}-V
$$
and the integral operators
$$
\Qa(x,dy):=q(x,y)~\mu(dy)\quad \mbox{and}\quad
\Ra(y,dx):=r(y,x)~\eta(dx)
$$
In this context, the recursive formulae stated in Proposition~\ref{prop-schp} take the form
$$
\Ua_{2n+1}=\Ua_{2n}=
\log{\Qa(e^{-\Va_{2n}})}
\quad \mbox{and}\quad
\Va_{2(n+1)}=\Va_{2n+1}=
\log{\Ra(e^{-\Ua_{2n+1}})}
$$

\begin{rmk}
  In terms of the potential functions $(\Ua_n,\Va_n)$, the probability measures in (\ref{sinhorn-entropy-form-Sch}) can be rewritten as
$$
\Pa_{n}(d(x,y))=e^{-(\Ua_{n}(x)+\Va_{n}(y))}~\Oa(d(x,y))
$$
with $(\Ua_0,\Va_0)=(0,-V)$ and the reference bounded measure
$$
\Oa(d(x,y)):=q(x,y)~(\eta\otimes \mu)(d(x,y))=e^{-V(y)}~\Pa_{0}(d(x,y)).
$$ 
 For any $\Pa\ll \Oa$, we have
$$
 \mbox{\rm Ent}(\Pa~|~\Oa)=-\int~\log{q(x,y)}~\Pa(d(x,y))+ \mbox{\rm Ent}(\Pa~|~\eta\otimes \mu).
$$
 In this case, the entropic transport problem with reference measure $\Oa$ consists in finding $\Pa\in\Ca(\eta,\mu)$ with minimal entropy $ \mbox{\rm Ent}(\Pa~|~\Oa)$. 
For the Gaussian model (\ref{gauss-model-6-1}), in terms of the parameter $\theta=(\alpha,\beta,\tau)$, we have
 $$
 \Oa(d(x,y))=e^{-V(y)}~P_{\theta}(d(x,y)).
 $$
Choosing $\Pa=P_{\theta_1}\in\Ca(\eta,\mu)$ yields
$$
\int~V(y)~P_{\theta_1}(d(x,y))=\mu(V)
 $$
and we obtain
 $$
  \mbox{\rm Ent}(\Pa~|~\Oa)=\mu(V)+  \mbox{\rm Ent}(P_{\theta_1}~|~P_{\theta}).
 $$
 This shows that the entropic transport problem is equivalent to the (static) Schr\"odinger bridge with reference measure $P_{\theta}$, as shown in \eqref{opt}. 
 \end{rmk}

Proposition~\ref{prop-schp} combined with {the Sinkhorn iterations as defined in} \eqref{def-Pa-n} and \eqref{s-2} yields the formulae
$$
\frac{ d\Pa_{2n}}{ d\Pa_{2n+1}}(x,y)=\frac{d\pi_{2n}}{d\mu}(y)=\exp{\left(V_{2n+1}(y)-V_{2n}(y)\right)}.
$$
In the same vein, using  \eqref{s-2} we verify that
$$
\frac{ d\Pa_{2n+1}}{ d\Pa_{2(n+1)}}(x,y)=\frac{d\pi_{2n+1}}{d\eta}(x)=\exp{\left(U_{2(n+1)}(x)-U_{2n+1}(x)\right)}.
$$
We summarize the above discussion with the following proposition.
\begin{prop}\label{prop-series}
For every $n\geq 0$, we obtain
$$
V_{2n}=V_0+\sum_{0\leq p< n}
\log\frac{ d\pi_{2p}}{d\mu}
\quad \mbox{and}\quad
U_{2n}=U_0+\sum_{0\leq p< n}\log{\frac{d \pi_{2p+1}}{d\eta}}.
$$
In addition, we have the monotone properties
$$
\begin{array}{rcccccl}
\mu(V_{2(n+1)})&\leq&
\mu(V_{2n})&=&\mu(V_{2(n+1)})+\mbox{\rm Ent}(\mu~|~\pi_{2n})&\leq& \mu(V_0)=0, \quad \text{and}\\
\eta(U_{2(n+1)})&\leq &\eta(U_{2n})&=&\eta(U_{2(n+1)})+\mbox{\rm Ent}(\eta~|~\pi_{2n+1})&\leq& \eta(U_0)=\eta(U).
\end{array}
$$
\end{prop}

\begin{examp}\label{examp-gauss-sc}
Consider the cost function $c(x,y)=-\log{q(x,y)}$ and set
$$
c_{\eta}(y):=\int~\eta(dx)~c(x,y)\quad\mbox{\rm with}\quad
c^{\mu}(x):=\int~\mu(dy)~c(x,y).
$$
For the Gaussian model in \eqref{gauss-model-6-1} we readily verify that
\begin{equation}\label{c-eta}
c_{\eta}(y)=\frac{1}{2}\left\Vert \tau^{-1/2}(y-m_0)\right\Vert_F^2+\frac{1}{2}~\tr(\tau^{-1}\sigma_{\beta})+\frac{1}{2}\log{\mbox{\rm det}(2\pi \tau)},
\end{equation}
with $m_0:=(\alpha+\beta m)$ and $\sigma_{\beta}:=(\beta\sigma\beta^{\prime})$, as well as
\begin{equation}\label{c-mu}
c^{\mu}(x)=\frac{1}{2}\left\Vert \tau^{-1/2}(\overline{m}-(\alpha+\beta x))\right\Vert_F^2+\frac{1}{2}~\tr(\tau^{-1}\overline{\sigma})+\frac{1}{2}\log{\mbox{\rm det}(2\pi \tau)}.
\end{equation}
Equations \eqref{c-eta} and \eqref{c-mu} are obtained by way of elementary calculations. We sketch proofs in Appendix~\ref{sec-tech-proofs} (see page~\pageref{c-mu-proofs}).
\end{examp}
The proposition below applies to Gaussian models of the form \eqref{gauss-model-6-1}. It provides some rather crude uniform estimates --more refined estimates that also apply to Gaussian models are presented in Section 5 of \cite{promit-2022}.
\begin{prop}\label{prop-uint}
Assume that, for any $z\in \RR^d$, the inequalities
$$
\Qa(\exp{(c_{\eta})})(z)<\infty\quad \mbox{and}\quad\Ra(\exp{(c^{\mu})})(z)<\infty
$$
are satisfied. Then, the uniform estimates
$$
\begin{array}{rcccl}
-c^{\mu}+\mu(V)&\leq &
U_{2n}-U&\leq& \displaystyle\log{\Qa(\exp{(c_{\eta})})} \\
-c_{\eta}&\leq &V_{2n}-V&\leq &\displaystyle-\mu(V)+\log{\Ra(\exp{(c^{\mu})})}
\end{array}$$
hold for every $n \ge 1$.
\end{prop}
\proof
 Applying Jensen's inequality we have
\begin{eqnarray}
\log{Q(\exp{(-V_{2n})})(x)}&=&\log{\int~\mu(dy)~\exp{(-c(x,y)+V(y)-V_{2n}(y))}}\nonumber\\
&\geq&
-c^{\mu}(x)+\mu(V)-\mu(V_{2n})\nonumber\\
&\geq& -c^{\mu}(x)+\mu(V),
\label{eq_A}
\end{eqnarray}
and, in a similar manner,
\begin{eqnarray}
\log{R(\exp{(-U_{2n})})(y)}&=&\log{\int~\eta(dx)~\exp{(-c(x,y)+U(x)-U_{2n}(x))}}\nonumber\\
&\geq& -c_{\eta}(y)+\eta(U)-\eta(U_{2n})\nonumber\\
&\geq& -c_{\eta}(y),
\label{eq_B}
\end{eqnarray}
{where the second inequality follows from Proposition \ref{prop-series}. Hence, combining Proposition \ref{prop-schp} with \eqref{eq_A} and \eqref{eq_B} above we arrive at}
$$
U_{2n}\geq U-c^{\mu}+\mu(V)
\quad\mbox{\rm and}\quad
V_{2n} \geq V-c_{\eta},
$$
which imply that
$$
\begin{array}{rcccl}
-c^{\mu}+\mu(V)&\leq &
U_{2n}-U&\leq& \log{Q(\exp{(c_{\eta}-V)})} \\
-c_{\eta}&\leq &V_{2n}-V&\leq &-\mu(V)+\log{R(\exp{(c^{\mu}-U)})}.
\end{array}$$
Finally, we observe that
\begin{eqnarray*}
Q(\exp{(c_{\eta}-V)})(x)&=&\int~\mu(dy)~\exp{(c_{\eta}(y)-c(x,y))}=\Qa(\exp{(c_{\eta})})(x) \quad \text{and}\\
R(\exp{(c^{\mu}-U)})(y)&=&\int~\eta(dx)~\exp{(c^{\mu}(x)-c(x,y))}=\Ra(\exp{(c^{\mu})})(y).
\end{eqnarray*}
\cqfd

\begin{examp}
Consider the Gaussian model (\ref{gauss-model-6-1}) and the integrated cost functions
$(c_{\eta},c^{\mu})$ defined in (\ref{c-eta}) and (\ref{c-mu}). We have
\begin{equation}\label{c-eta-2}
\begin{array}{l}
\displaystyle\log{\Qa(\exp{(c_{\eta})})(x)}-\frac{1}{2}~\tr(\tau^{-1} \sigma_{\beta})\\
\\
\displaystyle=
\frac{1}{2}~(\cchi_{\theta}(x-m))^{\prime}
(\overline{\sigma}-\tau)~(\cchi_{\theta}~(x-m))+(\overline{m}-m_0)^{\prime}\cchi_{\theta}(x-m),
\end{array}
\end{equation}
with $\cchi_{\theta}$ defined in \eqref{def-w-1}, as well as
\begin{equation}\label{c-mu-2}
\begin{array}{l}
\displaystyle\log{\Ra(\exp{(c^{\mu})})(y)}-\frac{1}{2}~\tr(\tau^{-1}\overline{\sigma})\\
\\
\displaystyle=
\frac{1}{2}
(y-\overline{m})^{\prime}\left(\cchi_{\theta}\,\sigma\,\cchi_{\theta}^{\prime}-\tau^{-1}\right)
(y-\overline{m})-(y-\overline{m})^{\prime}\tau^{-1}(\overline{m}-m_0).
\end{array}
\end{equation}
Equations \eqref{c-eta-2} and \eqref{c-mu-2} follow elementary calculations.  Proofs are sketched in Appendix~\ref{sec-tech-proofs} (see page~\pageref{c-mu-2-proofs}).
\end{examp}
We further assume that the series in Proposition~\ref{prop-series} 
converge almost everywhere, i.e., for almost every $x$ and $y\in\RR^d$ {and $q\geq 0$} we have
\begin{eqnarray}
\lim_{n\rightarrow\infty}U_{2n}(x)=\UU(x)&:=&U_{{2q}}(x)+\sum_{p\geq {q}}\log{\frac{d \pi_{2p+1}}{d\eta}(x)} \quad \text{and}
\nonumber\\
\lim_{n\rightarrow\infty}
V_{2n}(y)=\VV(y)&:=&V_{{2q}}(y)+\sum_{p\geq {q}}
\log{\frac{ d\pi_{2p}}{d\mu}(y)}.
\label{hyp-cv}
\end{eqnarray}
For Gaussian models, the convergence of the above series can be easily verified following the arguments provided in Remark~\ref{ref-rmk-en-marg}.
We refer to Section~\ref{gauss-pot-sect} for more refined convergence rates of these series in the context of Gaussian models.

  Following the discussion in Section 6 of~\cite{nutz}, the uniform estimates presented in Proposition~\ref{prop-uint} can also be used to check the boundedness property of the sequences of potentials $U_{2n}$ and $V_{2n}$ in Lebesgue spaces. By the uniqueness property of the Schr\" odinger bridge $\Pa$, all the extracted convergent sub-sequences converge to $\UU$ and $\VV$.
In this context, the bridge distribution  the $\Pa$ has the form (\ref{sinhorn-entropy-form-Sch-lim}) and we have
\begin{eqnarray*}
  \mbox{\rm Ent}(\Pa~|~\Pa_{2n})  &=&\eta(U_{2n}-\UU)+\mu(V_{2n}-\VV)\\
  &=&\mu\left(\sum_{p\geq n}\log{\frac{d\mu}{d\pi_{2p}}}\right)+\eta\left(\sum_{p\geq n}\log{\frac{d\eta}{d\pi_{2p+1}}}\right)=\Pa\left(\log{\frac{d\Pa}{d\Pa_{2n}}}\right).
\end{eqnarray*}
{\begin{rmk}
As shown in Corollary~\ref{cor-commute-ss} in the context of Gaussian models,  (\ref{hyp-cv}) also implies that the bridge distribution  $\Pa$ in (\ref{sinhorn-entropy-form-Sch-lim}) can be computed  at any level of Sinkhorn iterations, in the sense that for any $q\geq 0$ we have
$$
 \Pa=\argmin_{\PP\,\in\, \Ca(\eta,\mu)}\mbox{\rm Ent}(\PP~|~\Pa_{2q})
$$\end{rmk}}
On the other hand, choosing $q>p=n$ in (\ref{pyth}) we obtain
\begin{eqnarray*}
\Pa\left(\log{\frac{d\Pa_{2q}}{d\Pa_{2n}}}\right)&=&\mbox{\rm Ent}(\Pa~|~\Pa_{2n})-\mbox{\rm Ent}(\Pa~|~\Pa_{2q})\\
&=&\sum_{n\leq p< q}\left(\mu\left(\log{\frac{d\mu}{d\pi_{2p}}}\right)+\eta\left(\log{\frac{d\eta}{d\pi_{2p+1}}}\right)\right).
\end{eqnarray*}
The above decompositions readily imply the following equivalence property.
\begin{theo}
Assume that the series in \eqref{hyp-cv} converges a.s. Then, we have
$$
\mbox{\rm Ent}(\Pa~|~\Pa_{2q})\stackrel{q \to \infty}{\longrightarrow} 0
$$
if, and only if, 
$$
\begin{array}{l}
\displaystyle\sum_{p\geq n}\mu\left(\log{\frac{d\mu}{d\pi_{2p}}}\right)+\sum_{p\geq n}\eta\left(\log{\frac{d\eta}{d\pi_{2p+1}}}\right)
\displaystyle=\mu\left(\sum_{p\geq n}\log{\frac{d\mu}{d\pi_{2p}}}\right)+\eta\left(\sum_{p\geq n}\log{\frac{d\eta}{d\pi_{2p+1}}}\right)
\end{array}
$$
for every $n\geq 0$. Moreover, for any $n\geq 0$ we have the entropy formulae
\begin{eqnarray}
  \mbox{\rm Ent}(\Pa~|~\Pa_{2n})
  &=&\sum_{p\geq  n} \left( \mbox{\rm Ent}(\eta~|~\pi_{2p+1})+ \mbox{\rm Ent}(\mu~|~\pi_{2p})\right).
  \label{entrop-formula}
\end{eqnarray}
\end{theo}
As a direct consequence of the dominated convergence theorem, Proposition \ref{cond-inversion} below provides a sufficient condition to interchange summation and integration. 
\begin{prop}\label{prop-marg-tot}
Assume that the series in \eqref{hyp-cv} converges a.s. and
\begin{equation}\label{cond-inversion}
\sum_{p\geq  0}\eta\left(\vert\log{\frac{d\eta}{d \pi_{2p+1}}}\vert\right)\vee
\sum_{p\geq 0}\mu(\vert
\log\frac{d\mu}{ d\pi_{2p}}\vert)<\infty.
\end{equation}
Then, the bridge distribution  $\Pa$ in (\ref{sinhorn-entropy-form-Sch-lim})  satisfies the entropy formulae (\ref{entrop-formula}).
\end{prop}

In terms of potential functions, condition \eqref{cond-inversion} takes the form of the inequalities
\begin{equation}\label{cond-inversion-pot}
\sum_{p\geq  0}\eta\left(\vert U_{2(p+1)}-U_{2p}\vert\right)<\infty
\quad \mbox{\rm and}\quad
\sum_{p\geq  0}\mu\left(\vert V_{2(p+1)}-V_{2p}\vert\right)<\infty,
\end{equation}
which are clearly verified when
$$
\sum_{p\geq  0}\eta\left(\vert U_{2p}-\UU\vert\right)<\infty
\quad \mbox{\rm and}\quad
\sum_{p\geq  0}\mu\left(\vert V_{2p+1}-\VV\vert\right)<\infty.
$$
An application of Proposition \ref{prop-marg-tot} to the the Gaussian model \eqref{gauss-model-6-1} is presented in Corollary~\ref{cor-marg-tot}. 
 
 \begin{rmk}\label{rmk-int-nutz}
Integrability conditions for the convergence  $\mbox{\rm Ent}(\Pa~|~\Pa_{2q})\stackrel{q \to \infty}{\longrightarrow} 0$ are presented in Section 3 in~\cite{ruschen} as well as in~\cite{promit-2022} and~\cite{nutz} (see for instance Theorem 6.15 in \cite{nutz}). We remark that some sufficient integrability conditions discussed in the literature rely on global minorisation conditions or exponential-type uniformly  integrability conditions which are generally not satisfied for the Gaussian model discussed in the present article. For instance, in terms of the cost function $c(x,y)=-\log{q(x,y)}$, the condition presented in Theorem 6.15 in~\cite{nutz} takes the form
\begin{equation}\label{cond-int-nutz-wiesel}
\exists \epsilon>1\quad\mbox{\rm such that}\quad \int\eta(dx)\mu(dy)~e^{\epsilon\,c(x,y)}<\infty.
\end{equation}
This condition is not met for the simple quadratic cost  (\ref{gibbs-cost}) in one dimension with $t=1$ and
the centered Gaussian $\eta=\mu=\nu_{0,1}$. We underline that the above condition is met when the regularization parameter $t$ is chosen sufficiently large. In a more recent article~\cite{nutz-wiesel}, the authors show the convergence of Sinkhorn iterates  $\Pa_n$ to the bridge distribution $\Pa$ as $n\rightarrow\infty$ when condition (\ref{cond-int-nutz-wiesel}) is met  for some $\epsilon>0$.
\end{rmk}

\subsection{Gaussian potential functions}\label{gauss-pot-sect}
 The main objective of this section is to obtain a closed form expression of the Schr\" odinger potential functions in (\ref{sinhorn-entropy-form-Sch}) for the Gaussian model described by (\ref{gauss-model}) and (\ref{gauss-model-6-1}). Recall that the Sinkhorn algorithm discussed in Section~\ref{sec-sinkhorn} starts at some reference parameter $\theta_0=\theta:=(\alpha,\beta,\tau)\in\Theta$ and consider the integral operators
\begin{eqnarray*}
Q_{\theta}(x,dy)&:=&q_{\theta}(x,y)~dy, \quad \text{and}\\
R_{\theta}(x,dy)&:=&r_{\theta}(x,y)~dy \quad \mbox{\rm with}\quad
q_{\theta}(x,y)=r_{\theta}(y,x):=g_{\tau}(y-(\alpha+\beta x)).
\end{eqnarray*} 
In what follows $ (m_{n},\sigma_n)$ and
$\theta_n=(\alpha_n,\beta_n,\tau_n)$ stands for the flow of Sinkhorn parameters (\ref{ref-mean-cov-intro}) starting at $\theta_0=\theta$.

 Our approach is based on the series expansions presented in Proposition~\ref{prop-series}. By Proposition~\ref{prop-schp} it suffices to analyze potential functions indexed by even indices.

\begin{lem}\label{lem-key-potentials}
For any $n\geq 0$ we have
\begin{equation}\label{rec-xii}
\sigma_{2n}^{-1}-\overline{\sigma}^{-1}=\tau^{-1}_{2n}-\tau^{-1}_{2(n+1)}
\quad\mbox{and}\quad
\sigma_{2n+1}^{-1}-\sigma^{-1}=\tau_{2n+1}^{-1}-\tau_{2n+3}^{-1}.
\end{equation}
We also have the determinant formulae
\begin{equation}
{\mbox{\rm det}(\sigma^{-1}\sigma_{2n+1})}={\mbox{\rm det}(\upsilon_{2n+1}~\upsilon_{2(n+1)}^{-1})}\quad \mbox{and}\quad
\mbox{\rm det}(\overline{\sigma}^{-1}\sigma_{2n})=\mbox{\rm det}(\upsilon_{2n}\upsilon^{-1}_{2n+1}),
\label{rec-det}
\end{equation}
as well as the variance equations
\begin{equation}
\sigma_{2n}^{-1}\beta_{2n}=\tau^{-1}\beta~\left(I-\beta_{2n+1}\beta_{2n}\right)
\quad \mbox{and}\quad
\sigma_{2n+1}^{-1}\beta_{2n+1}=
\beta^{\prime}~\tau^{-1}~\left(I-\beta_{2(n+1)}\beta_{2n+1}\right)\label{rec-sig}.
\end{equation}
\end{lem}
A proof is provided in Appendix \ref{sch-appendix} --see page~\pageref{lem-key-potentials-proof}.

Using (\ref{rec-det}),  we readily find that
\begin{eqnarray*}
V_{2n}(\overline{m}) &=&\frac{1}{2}~\sum_{0\leq p< n}\log{\mbox{\rm det}(\upsilon_{2p+1}\upsilon^{-1}_{2p})}
-\frac{1}{2}~\sum_{0\leq p< n}
\Vert\sigma_{2p}^{-1/2}\left(m_{2p}-\overline{m}\right)\Vert^2_F
\quad \text{and}
\\
U_{2n}(m)&=&U(m)+\frac{1}{2}~\sum_{0\leq p< n}\log{\mbox{\rm det}(\upsilon_{2(p+1)}\upsilon_{2p+1}^{-1})}-\frac{1}{2}\sum_{0\leq p<n}\Vert\sigma_{2p+1}^{-1/2}
\left(m_{2p+1}-m\right)\Vert_F^2.
\end{eqnarray*}
 Taking the sum of the above expressions we obtain the following decomposition.
\begin{prop}\label{prop-end}
For every $n\geq 0$ we have
$$
V_{2n}(\overline{m})+U_{2n}(m)
=
U(m)+\frac{1}{2}~\log{\mbox{\rm det}(\tau_{2n}\tau^{-1}_{0})}+\frac{1}{2}~\Vert\tau^{-1/2}_{2n}
\left(m_{2n}-\overline{m}\right)\Vert^2_F-\frac{1}{2}~\Vert\tau^{-1/2}_{0}
\left(m_{0}-\overline{m}\right)\Vert_F^2.
$$
\end{prop}
The proof of the above proposition is rather technical; it is in Appendix \ref{sch-appendix}, on page \pageref{prop-end-proof}.

Let us set
$$
\epsilon^V_{2n}(\overline{m}):=V_{2n}(\overline{m})-\VV_{\theta}(\overline{m})\quad \mbox{\rm and}\quad
\epsilon^U_{2n}(m):=U_{2n}(m)-\UU_{\theta}(m),
$$
with the limiting infinite series
\begin{equation}\label{ctm-pot}
\VV_{\theta}(\overline{m}) :=\lim_{n\rightarrow\infty}V_{2n}(\overline{m})\quad \mbox{\rm and}\quad
\UU_{\theta}(m):=\lim_{n\rightarrow\infty}U_{2n}(m).
\end{equation}
Recalling that $\tau_{2n}$ converges towards $\varsigma_{\theta}$ as $n\rightarrow\infty$  and $U(m)=\frac{1}{2}~\log{\mbox{\rm det}(\sigma)}$, Proposition~\ref{prop-end} also yields the formula (\ref{ref-uv-infty}).

Consider now the potential functions defined in Theorem~\ref{theo-s-pot} with the parameters
$(\UU_{\theta}(m),\VV_{\theta}(\overline{m}))$ defined in (\ref{ctm-pot}).
We are now in position to state the main result of this section.
\begin{theo}\label{theo-end}
For every $n\geq 1$ we have
$$
\displaystyle  V_{2n}(y)=\VV_{\theta}(y)
+\epsilon^V_{2n}(y)\quad\mbox{and}\quad
U_{2n}(x)=\UU_{\theta}(x)+\epsilon^U_{2n}(x),
$$
with the remainder functions
\begin{eqnarray*}
\epsilon^V_{2n}(y+\overline{m})&:=&\epsilon^V_{2n}(\overline{m})+y^{\prime}\tau^{-1}\beta~\beta^{\circ}_{2n-1,1}~\left(\overline{m}-m_0\right)+\frac{1}{2}~
y^{\prime}~\left(\tau^{-1}_{2n}-\varsigma_{\theta}^{-1}\right)~y,\\
\epsilon^U_{2n}(x+m)&:=&\epsilon^U_{2n}(m)+x^{\prime}~\tau_{1}^{-1}~\beta_{1}~
\beta^{\circ}_{2n,0}~\left(m-m_{1}\right)+\frac{1}{2}~x^{\prime}\left(\tau_{2n+1}^{-1}-\overline{\varsigma}_{\theta_1}^{-1}\right)
~x,
\end{eqnarray*}
and the directed products $\beta^{\circ}_{2n,0}$ and  $\beta^{\circ}_{2n-1,1}$ defined in \eqref{directed-prod}.
\end{theo}

See Appendix \ref{sch-appendix} (page~\pageref{theo-end-proof}) for a proof.

Next corollary is a direct consequence of the exponential estimates presented in Section~\ref{quant-sec}. A detailed proof is provided in Appendix~\ref{sch-appendix} --see page~\pageref{prop-ct-pot-proof}.

\begin{cor}\label{prop-ct-pot}
There exists some constants $c^U_{\theta},c^V_{\theta}$ and some parameter $n_{\theta}$ such that for any $n\geq n_{\theta}$ we have
\begin{eqnarray*}
\vert V_{2n}(y)-\VV_{\theta}(y)\vert&\leq & c^V_{\theta}~\left(~\rho_{\theta}^{n}+\overline{\rho}_{\theta_1}^{n/2}~\Vert y-\overline{m}\Vert+\rho_{\theta}^n~\Vert y-\overline{m}\Vert^2\right) \quad \text{and}\\
\vert U_{2n}(x)-\UU_{\theta}(x)\vert&\leq & c^V_{\theta}~\left(\overline{\rho}_{\theta_1}^{n}+\rho_{\theta}^{n/2}~\Vert x-m\Vert
+\overline{\rho}_{\theta_1}^n~\Vert x-m\Vert^2\right).
\end{eqnarray*}

\end{cor}

{Applying Proposition~\ref{prop-marg-tot} to
 the Gaussian model (\ref{gauss-model-6-1}) we have $(\eta,\mu):=(\nu_{m,\sigma},\nu_{\overline{m},\overline{\sigma}})$, $\pi_{n}=\nu_{m_{n},\sigma_{n}}$ as well as
$(\Pa,\Pa_{2n})=(P_{\SS(\theta)},P_{\theta_{2n}})$. In this context, condition (\ref{cond-inversion-pot}) is clearly satisfied.  Using the entropy estimates stated in Remark~\ref{ref-rmk-en-marg}  we readily find the following estimate.
\begin{cor}\label{cor-marg-tot}
There exists 
some constant $c_{\theta}>0$ and some $n_0$ such that for any $n\geq n_0$ we have
$$
  \mbox{\rm Ent}(P_{\SS(\theta)}~|~P_{\theta_{2n}})\leq c_{\theta}~(\rho_{\theta}\vee\overline{\rho}_{\theta_1})^{n}.
$$
\end{cor}
}

Applying Lebesgue's dominated convergence theorem, the integral equations stated in Proposition~\ref{prop-schp} converge as $n\rightarrow\infty$ to a
 system of integral equations
\begin{eqnarray*}
\UU_{\theta}=U+\log{Q_{\theta}(e^{-\VV_{\theta}})}\quad \mbox{\rm and}\quad
\VV_{\theta}=V+\log{R_{\theta}(e^{-\UU_{\theta}})},
\end{eqnarray*}
with the integral operators
$$
Q_{\theta}(x,dy):=q_{\theta}(x,y)~dy\quad \mbox{\rm and}\quad
R_{\theta}(x,dy):=r_{\theta}(x,y)~dy.
$$
The corollary below is a direct consequence of Theorem~\ref{theo-end} and Corollary~\ref{prop-ct-pot}. 
\begin{cor}\label{end-cor}
For any  $\theta\in \Theta$, the Schr\"odinger bridge (\ref{ref-uv-infty-intro-eq}) between the distributions $ \nu_{\overline{m},\overline{\sigma}}$ and  $\nu_{m,\sigma}$ with reference parameter $\theta$ satisfies the equation
\begin{eqnarray*}
P_{\SS(\theta)}= \PP_{\theta}\quad \mbox{and}\quad
K_{\SS(\theta)}(x,dy)= \frac{Q_{\theta}(x,dy)~e^{-\VV_{\theta}(y)}}{Q_{\theta}(e^{-\VV_{\theta}})(x)}.
\end{eqnarray*}
For any  $\theta\in \Theta$,  the dual Schr\"odinger bridge associated with (\ref{over-S-def}) between the distributions $ \nu_{\overline{m},\overline{\sigma}}$ and  $\nu_{m,\sigma}$ with reference parameter $\theta_1=\BB_{m,\sigma}(\theta)$  satisfies the equation
$$
\overline{P}_{\overline{\SS}(\theta_1)}=\PP_{\theta}^{\,\flat}
\quad \mbox{and}\quad
K_{\overline{\SS}(\theta_1)}(x,dy)=\frac{R_{\theta}(x,dy)~e^{-\UU_{\theta}(y)}}{R_{\theta}(e^{-\UU_{\theta}})(x)}.
$$
\end{cor}

The second assertion is also a direct consequence of the commutation property (\ref{comm-intro}).
An alternative and more direct proof of the above corollary based on the closed-forms of the Sch\" odinger potentials $(\UU_{\theta},\VV_{\theta})$   is provided in Appendix \ref{sch-appendix} (see page~\pageref{end-cor-proof}).

\section{Pseudocode and simulations of the Gaussian Sinkhorn algorithm}\label{sec:alg:gaussian-sinkhorn}

We provide below a pseudocode of the Gaussian Sinkhorn algorithm for practical implementation.
\begin{algorithm}[H]
\caption{The Gaussian Sinkhorn algorithm}
\label{alg:gaussian-sinkhorn}
\begin{algorithmic}[1]
\State Input: $(m,\sigma)$ and $(\overline{m},\overline{\sigma})$, reference parameters $\theta_0 = (\beta_0,\tau_0)$, the number of iterations $T$.
\State $\gamma_{\theta_0} = \overline{\sigma}^{1/2} \tau_0^{-1} \beta_0 \sigma^{1/2}$
\State Initialize $m_0$, $\sigma_0$.
\State Compute $v_0 = \overline{\sigma}^{-1/2} \tau_0 \overline{\sigma}^{-1/2}$.
\For{$n = 1, \ldots, T$}
\If{$n$ is even}
\State $v_n = (I_d + \gamma_{\theta_0} v_{n-1} \gamma_{\theta_0}')^{-1}$.
\State $\tau_n = \overline{\sigma}^{1/2} v_n \overline{\sigma}^{1/2}$.
\State $\beta_n = \tau_n \tau_0^{-1} \beta_0$.
\State $m_n = \overline{m} + \beta_n (m - m_{n-1})$.
\State $\sigma_n = \beta_n \sigma \beta_n' + \tau_n$.
\Else
\State $v_n = (I_d + \gamma_{\theta_0}' v_{n-1} \gamma_{\theta_0})^{-1}$.
\State $\tau_n = \sigma^{1/2} v_n \sigma^{1/2}$
\State $\beta_n = \tau_n \beta_0' \tau_0^{-1}$.
\State $m_n = m + \beta_n (\overline{m} - m_{n-1})$.
\State $\sigma_n = \beta_n \overline{\sigma} \beta_n' + \tau_n$. 
\EndIf
\EndFor
\end{algorithmic}
\end{algorithm}
For simplicity, we describe the algorithm with an \texttt{if} statement to separate the case when $n$ is even ($n=2k$) from the case when for $n$ is odd ($n=2k+1$). As discussed, this algorithm exactly, iteratively, solves the entropic optimal transport problem and provides estimates of the sufficient statistics of $\nu_{m, \sigma}$ and $\nu_{\overline{m}, \overline{\sigma}}$. Specifically, the sequence $(m_{2n}, \sigma_{2n})_{n\geq 0}$ provides the estimates of $(\overline{m}, \overline{\sigma})$ and converges as $n \to \infty$. Similarly, the sequence $(m_{2n + 1}, \sigma_{2n + 1})_{n\geq 0}$ provides estimates for $({m}, {\sigma})$ and similarly converges as $n\to\infty$. Below, we demonstrate the convergence behaviour in a simple 2D setting. This simulation also shows that this is a stable numerical algorithm that can be used to assess the performance of optimal transport methods.

\begin{figure}[htbp]
    \centering
    \setlength{\tabcolsep}{2pt}
    \renewcommand{\arraystretch}{0}
    \begin{tabular}{cccc}
        \begin{subfigure}[t]{0.23\textwidth}
            \includegraphics[width=\linewidth]{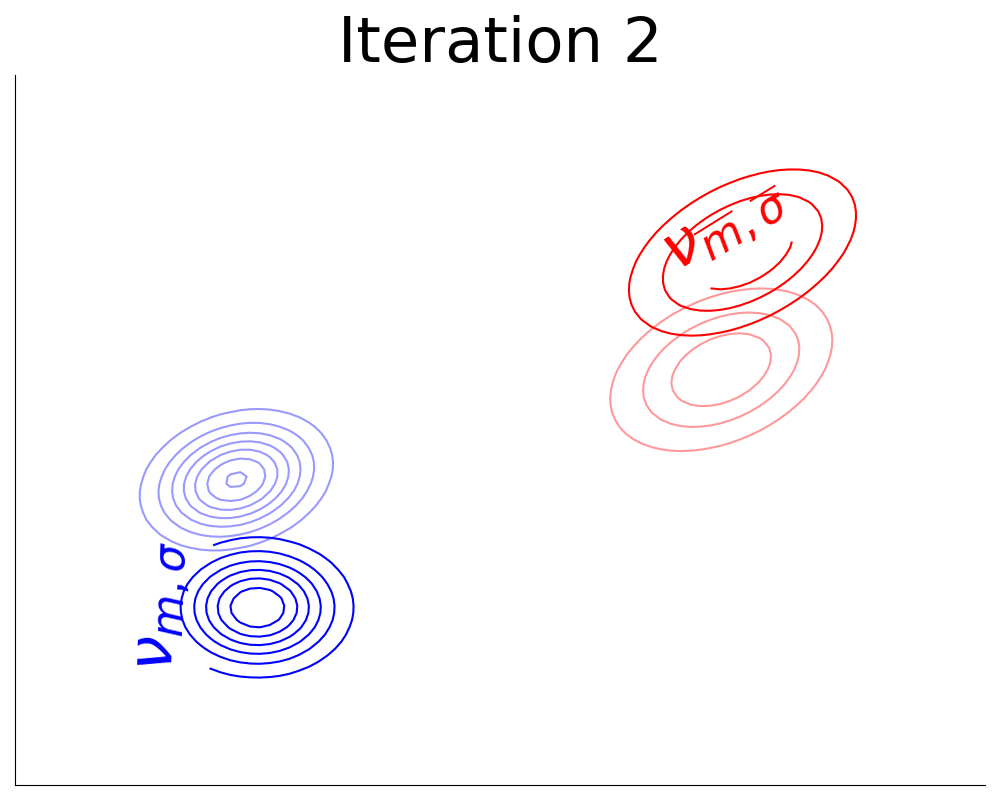}
        \end{subfigure} &
        \begin{subfigure}[t]{0.23\textwidth}
            \includegraphics[width=\linewidth]{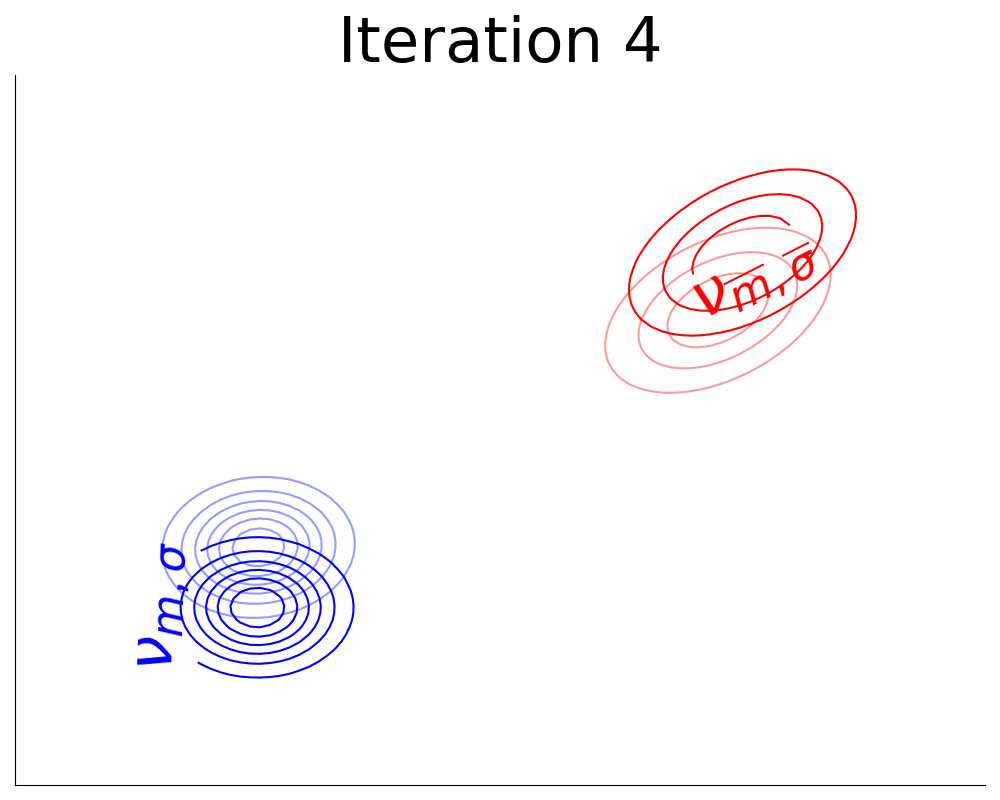}
        \end{subfigure} &
        \begin{subfigure}[t]{0.23\textwidth}
            \includegraphics[width=\linewidth]{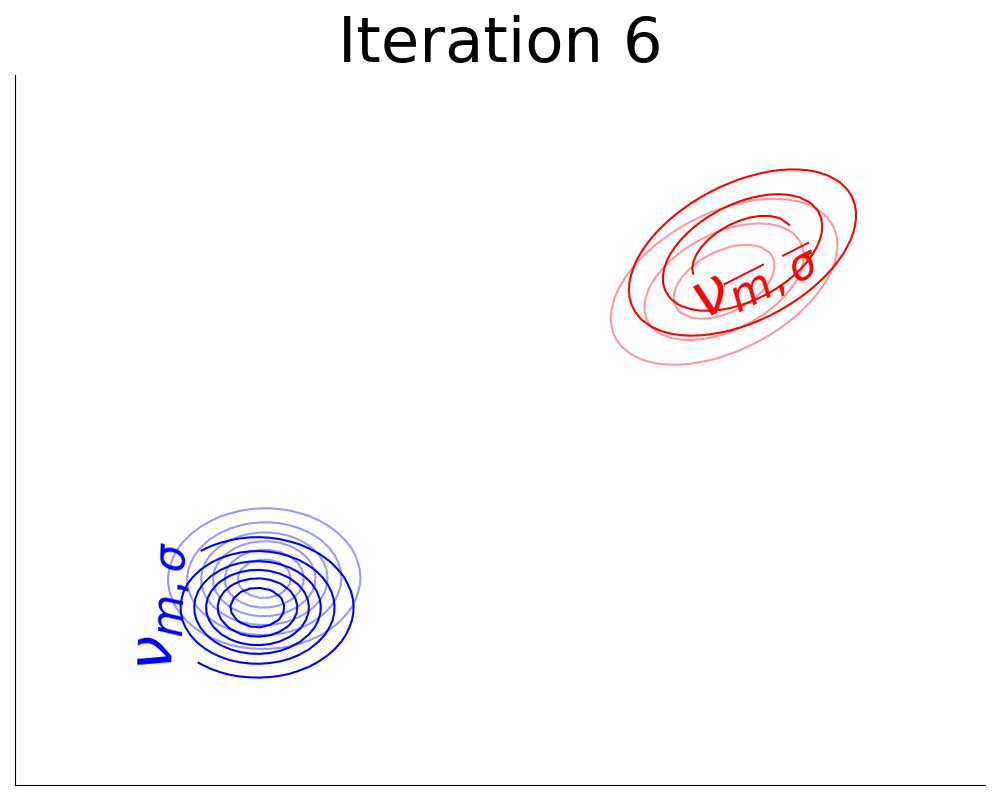}
        \end{subfigure} &
        \begin{subfigure}[t]{0.23\textwidth}
            \includegraphics[width=\linewidth]{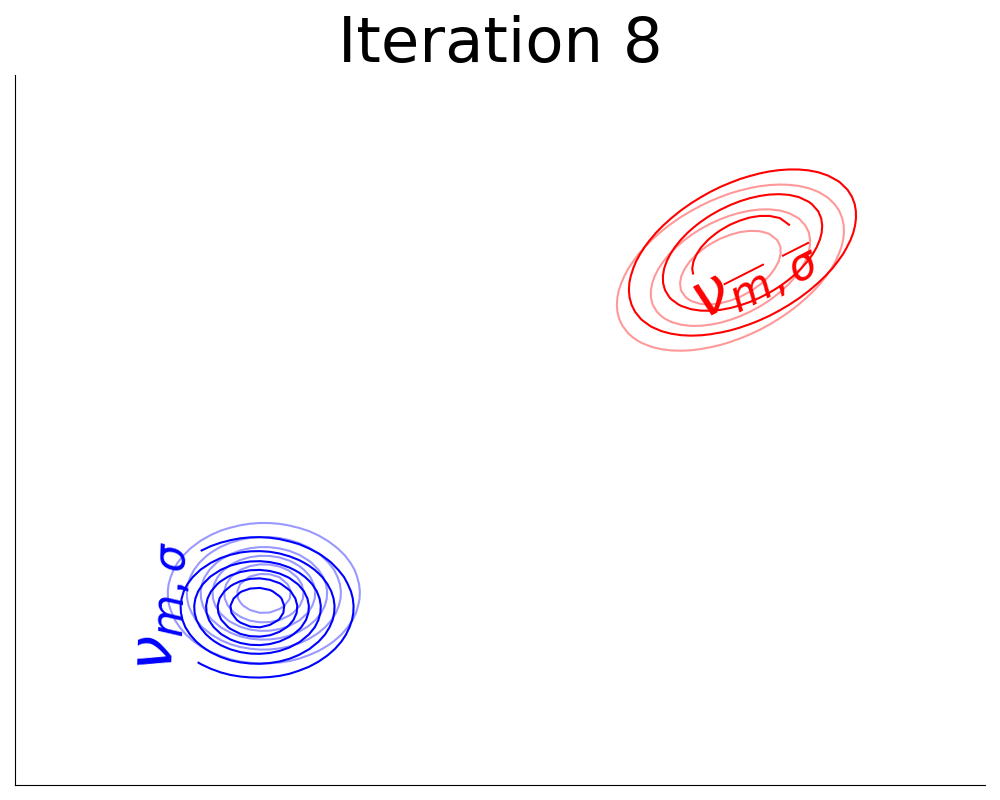}
        \end{subfigure} \\
        
        \begin{subfigure}[t]{0.23\textwidth}
            \includegraphics[width=\linewidth]{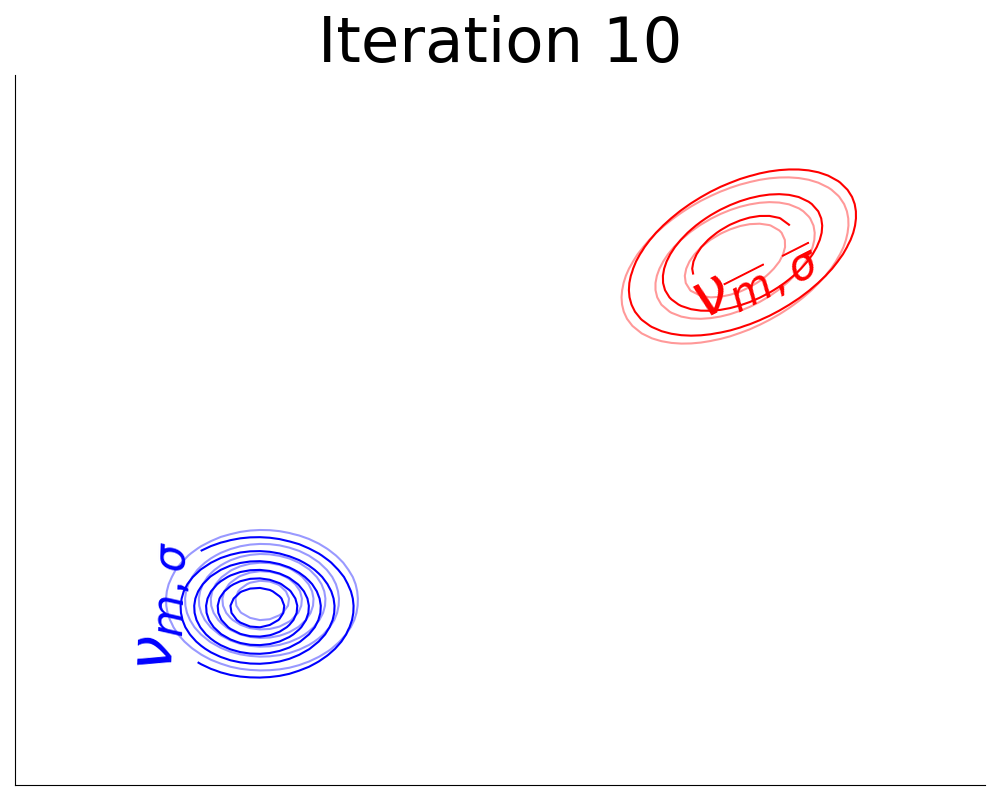}
        \end{subfigure} &
        \begin{subfigure}[t]{0.23\textwidth}
            \includegraphics[width=\linewidth]{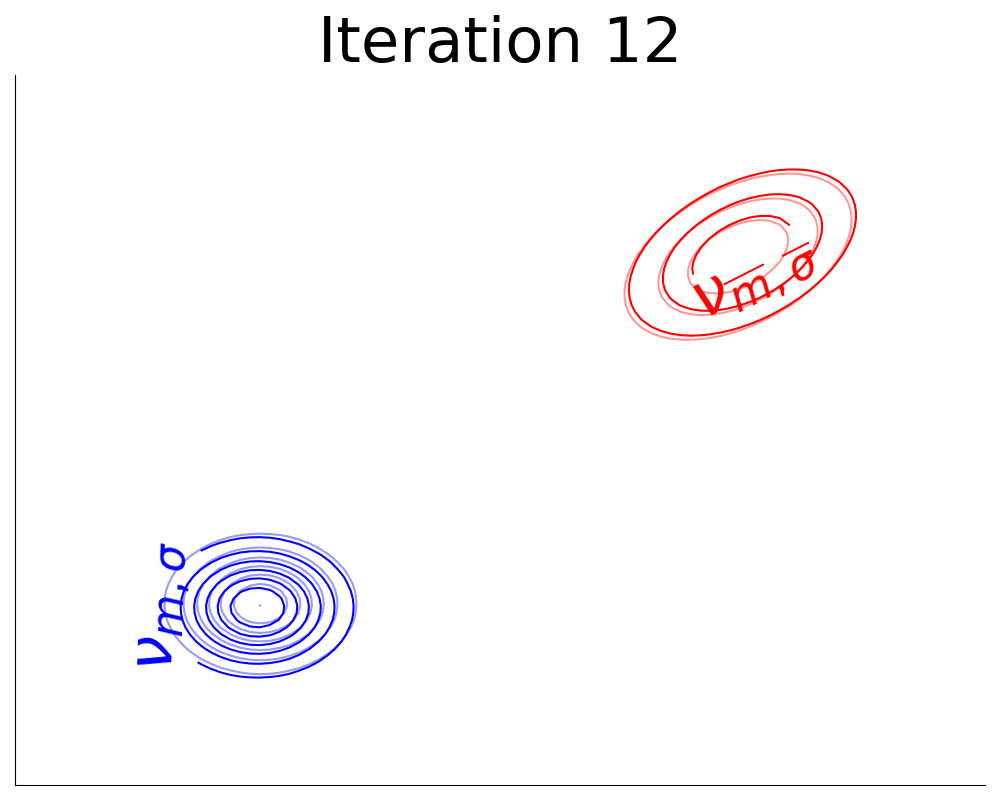}
        \end{subfigure} &
        \begin{subfigure}[t]{0.23\textwidth}
            \includegraphics[width=\linewidth]{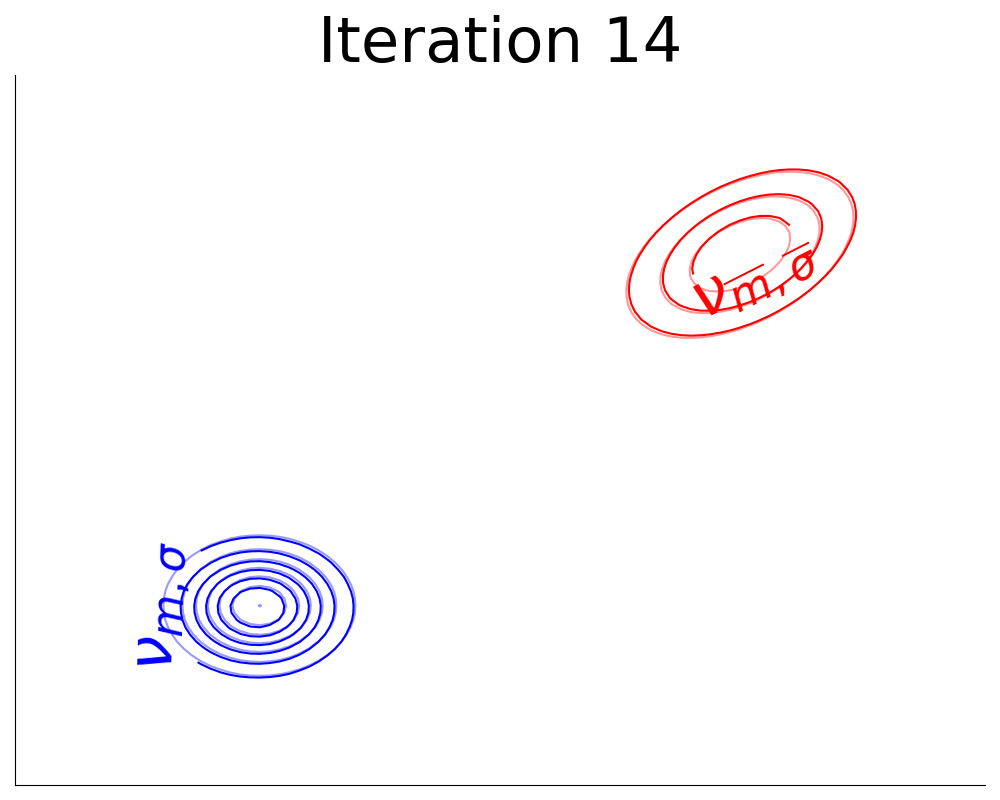}
        \end{subfigure} &
        \begin{subfigure}[t]{0.23\textwidth}
            \includegraphics[width=\linewidth]{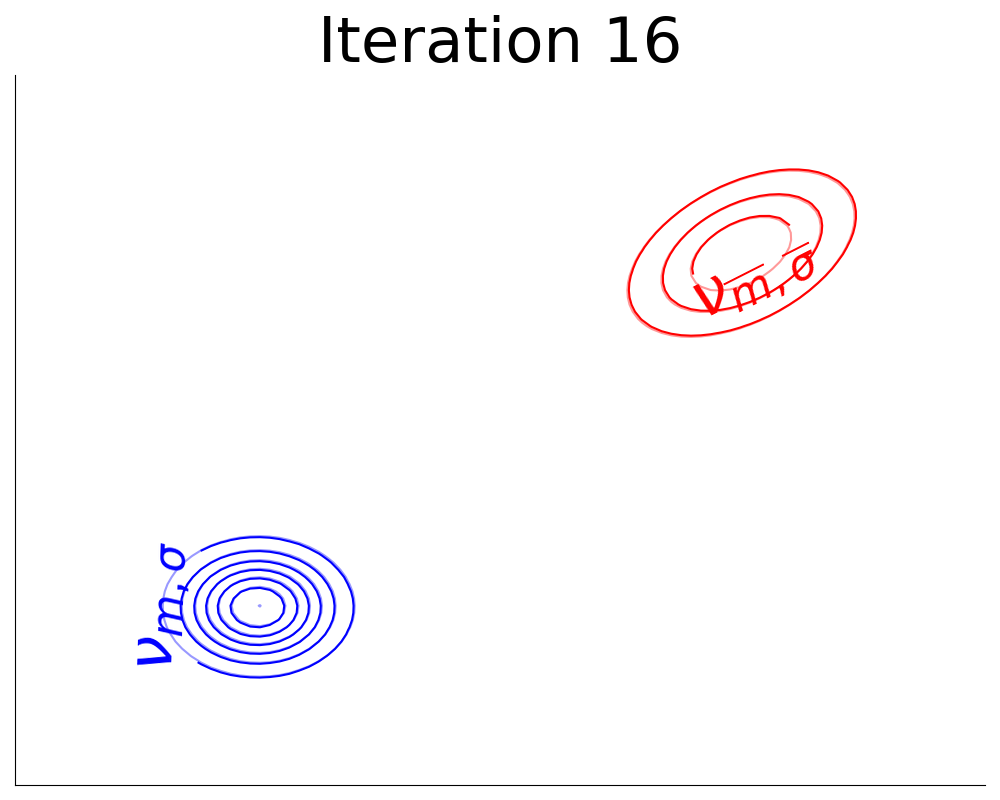}
        \end{subfigure} 
    \end{tabular}
    \caption{Evolution over time from $n=2$ to $n=24$ in steps of 2. The solid blue and red contours denote the distributions $\nu_{m, \sigma}$ (blue) and $\nu_{\overline{m},\overline{\sigma}}$ (red). The transparent contours shows Gaussian distributions that approximate the end points of the bridge iteratively. It can be seen that, from Iteration 2, Algorithm~\ref{alg:gaussian-sinkhorn} exhibits fast convergence to the distributions $\nu_{m,\sigma}$ and $\nu_{\overline{m},\overline{\sigma}}$, completely overlapping with the targets in around 10 iterations.}
    \label{fig:iteration_grid_3x4}
\end{figure}

The results of a numerical simulation in a 2D Gaussian setting can be seen in Fig.~\ref{fig:iteration_grid_3x4}. The iteration is initialized with
\begin{align*}
m_0 \sim \mathcal{N}(0, 10 I_2) \quad \quad \text{and} \quad \quad \sigma_0 = \begin{bmatrix}
10 & 9.99 \\
9.99 & 10
\end{bmatrix}.
\end{align*}

\begin{figure}[h]
    \centering
    \includegraphics[width=0.4\textwidth]{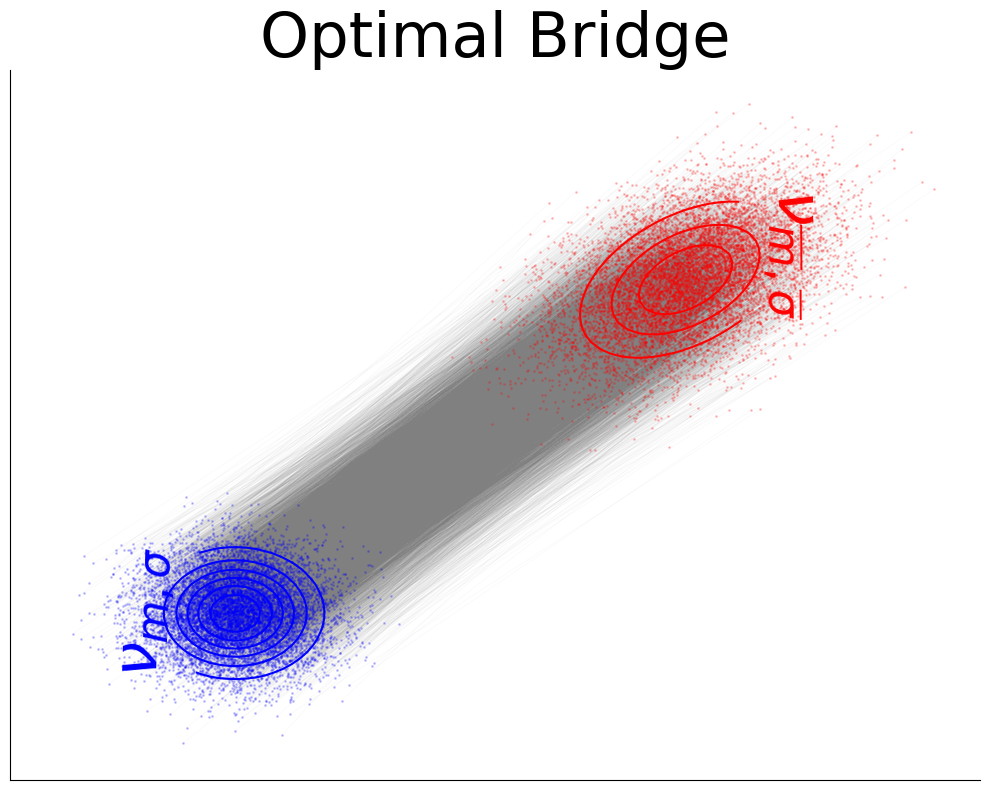}
    \caption{A numerical demonstration of the Schr\"odinger bridge from $\nu_{m, \sigma}$ to $\nu_{\overline{m},\overline{\sigma}}$ using samples from $\nu_{m,\sigma}$.}
    \label{fig:optimal_bridge}
\end{figure}

It can be seen from Fig.~\ref{fig:iteration_grid_3x4} that the sequence $(m_n, \sigma_n)_{n\geq 0}$ generated by the Algorithm~\ref{alg:gaussian-sinkhorn} exhibits a fast convergence as expected, in a numerically stable way. The associated optimal Schr\"odinger bridge given by the formulae in Theorem~\ref{Th1} can also be numerically demonstrated. To this end, we simulate $N = 10,000$ samples, i.e., draw $X_i \sim \nu_{m, \sigma}$ for $i = 1,\ldots, N$, and push these forward with the optimal formulae presented in Theorem~\ref{Th1}, namely,
\begin{align*}
Y_i = \iota_\theta + \kappa_\theta X_i + \varsigma_\theta^{1/2} \xi_i,
\end{align*}
where $\xi_i \sim \mathcal{N}(0, I_2)$. It is clear that, given the formulae in Theorem~\ref{Th1}, samples $\{Y_i\}_{i=1}^N$ are expected to be distributed as $\nu_{\overline{m}, \overline{\sigma}}$, which is illustrated by Fig.~\ref{fig:optimal_bridge}. A similar map can be constructed from $\nu_{\overline{m},\overline{\sigma}}$ to $\nu_{m, \sigma}$.

Next, we demonstrate the convergence rates derived in Section \ref{quant-sec}. We provide only a subset of possible numerical simulations as they are sufficient to demonstrate the behavior of the Sinkhorn iteration. In short, we consider Theorem~\ref{theo-qs} and Corollary~\ref{theo-cor-qs}. Specifically, we show that the slope of the approximation errors across the iterations has the slope predicted by the analysis (ignoring the constant $c_\theta$ as it is not explicit, although the numerical simulations also show that this constant is relatively small). We first consider Theorem~\ref{theo-qs}, recalling that
\begin{equation}\label{app:theo-5-1}
\Vert \tau_{2n}-\varsigma_{\theta}\Vert\vee \Vert \tau_{2n}^{1/2}-\varsigma_{\theta}^{1/2} \Vert\vee \Vert \beta_{2n}-\kappa_{\theta}\Vert \leq c_\theta \rho_{\theta}^n\Vert \tau_{0}-\varsigma_{\theta}\Vert,
\end{equation}
where
$
\rho_\theta = \left(1 + \lambda_{\min}(\varpi_\theta + r_\theta)\right)^{-2}.
$

\begin{figure}[t]
    \centering
    \includegraphics[width=0.32\textwidth]{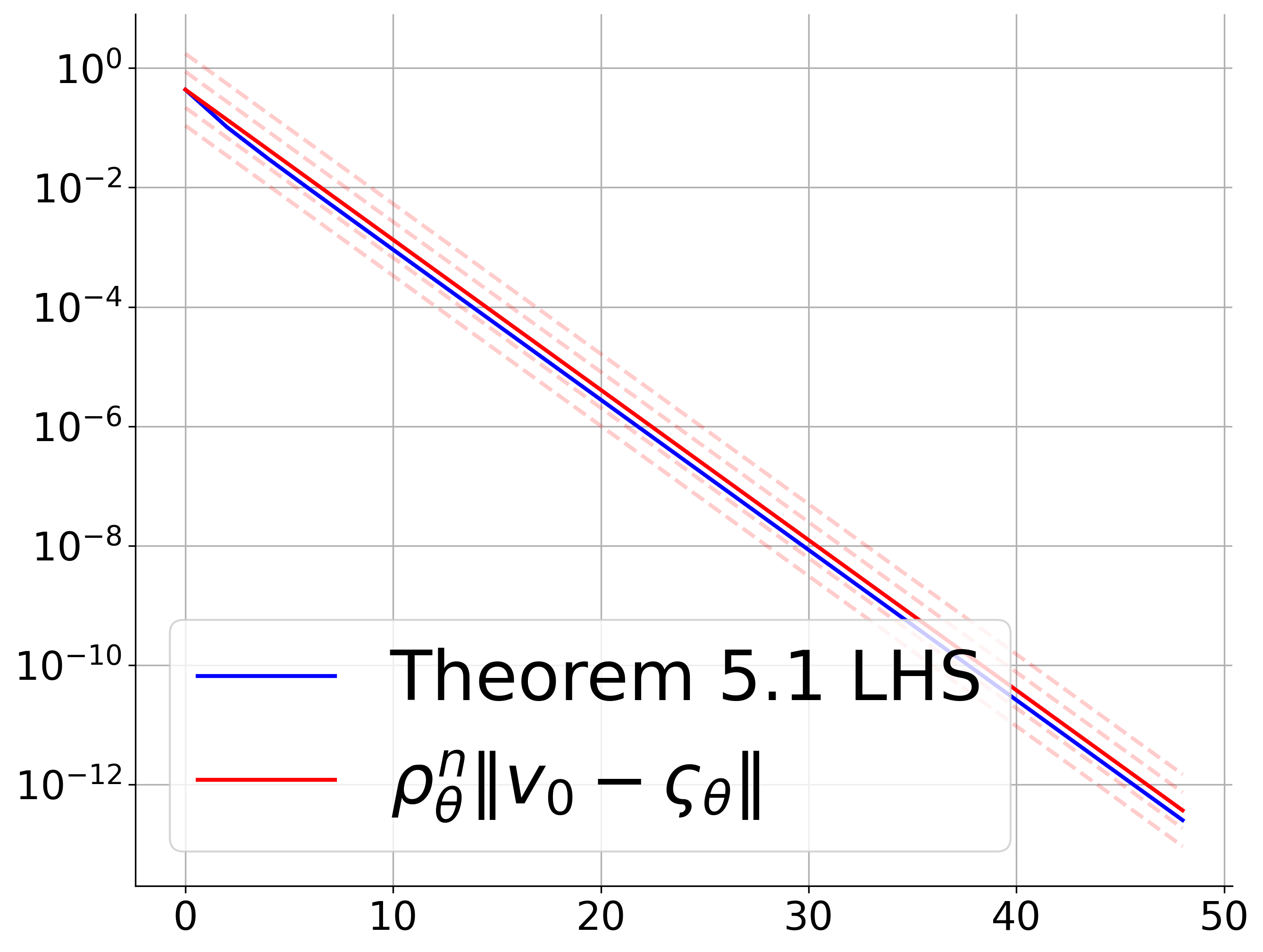}
    \hfill
    \includegraphics[width=0.32\textwidth]{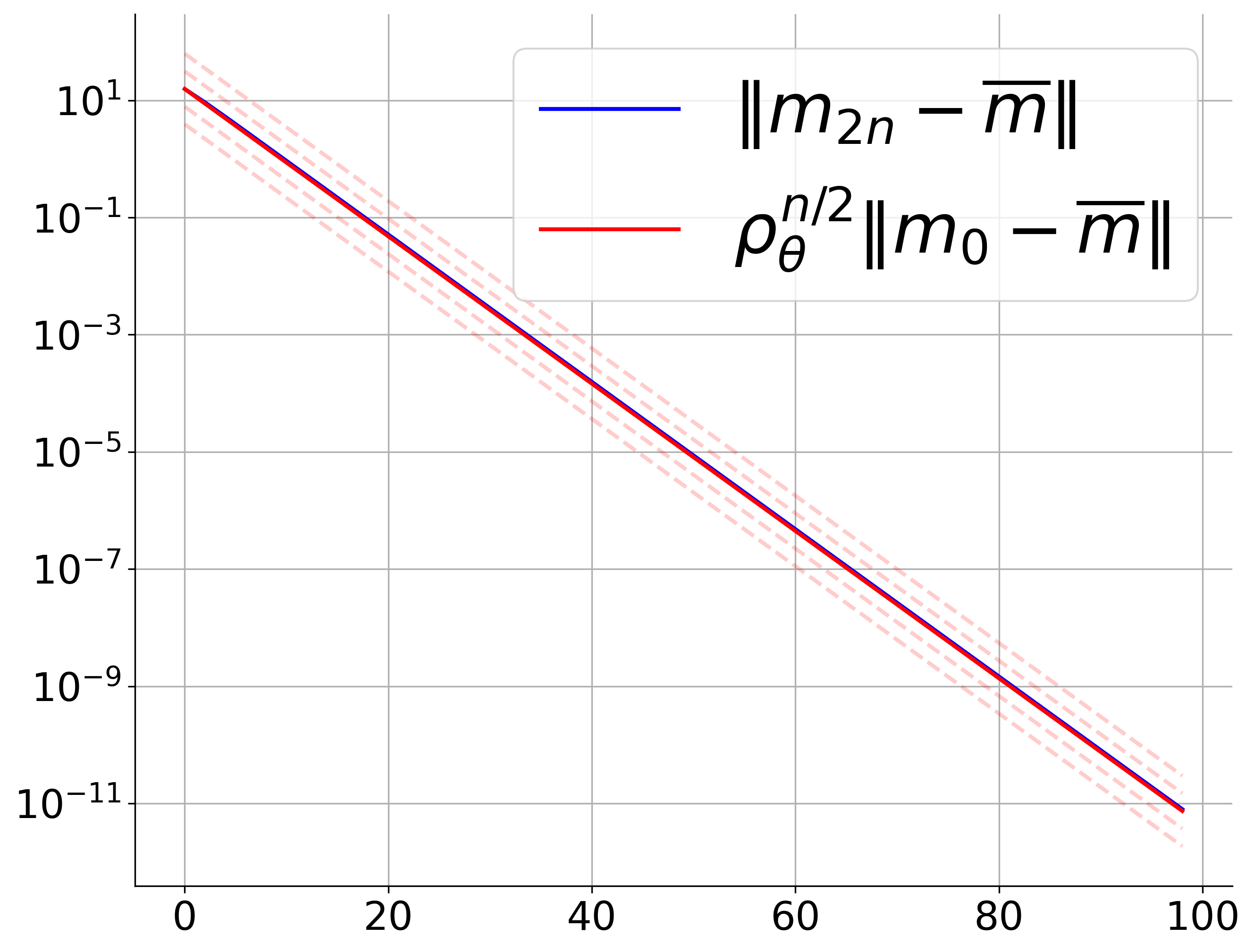}
    \hfill
    \includegraphics[width=0.32\textwidth]{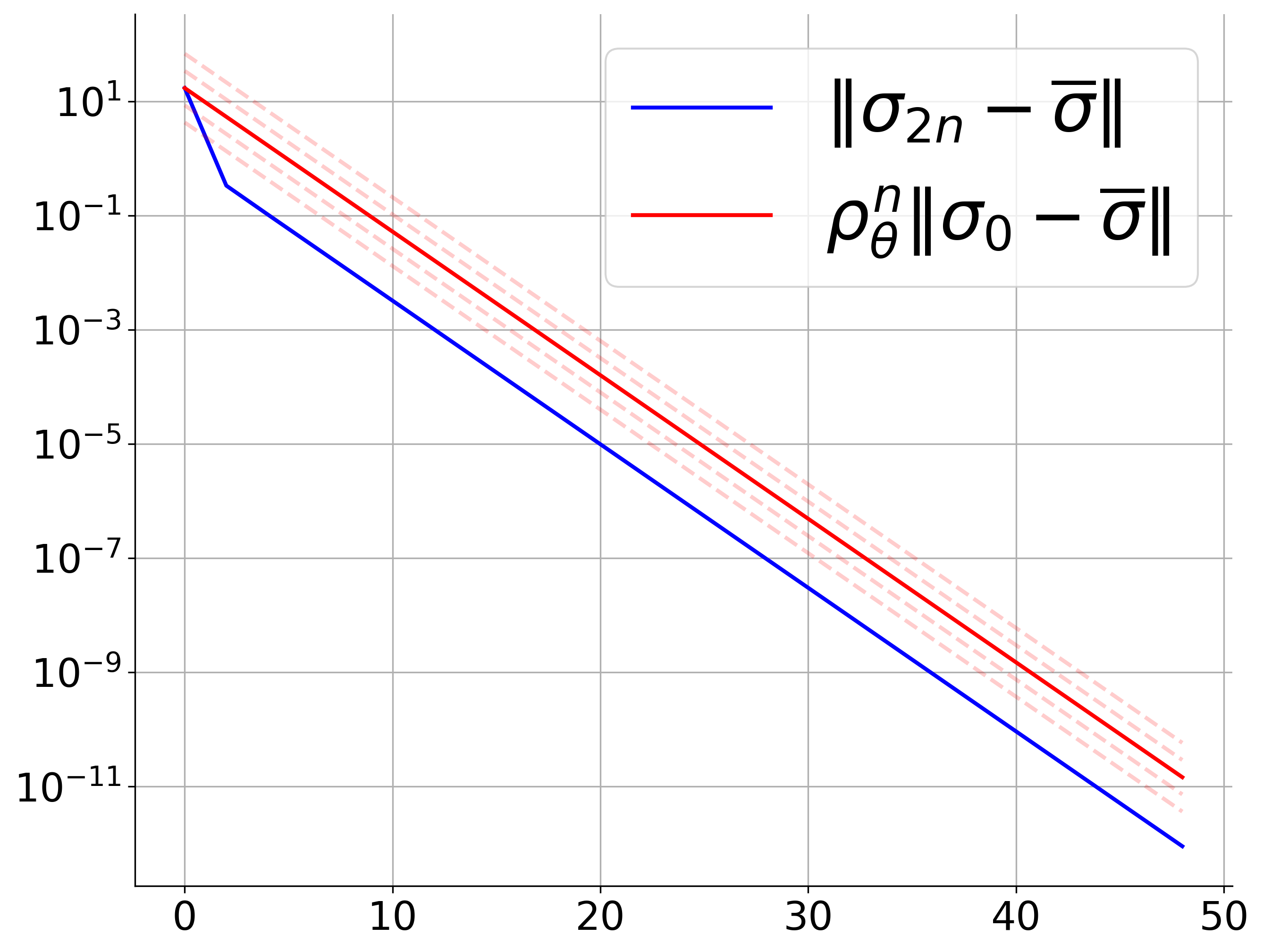}
    \caption{A demonstration of the convergence rates derived in the paper for the 2D example introduced above. On the left, one can see a numerical demonstration of Theorem~\ref{theo-qs}. In the middle and right, one can see a numerical demonstration of Corollary~\ref{theo-cor-qs}, indicating the rates we have derived are sharp, and constants $c_\theta$ are small since in the plotting it is ignored. Dashed lines are included just for a clearer visual demonstration of the rates w.r.t. the blue curves.}
    \label{fig:rates}
\end{figure}

We also recall Corollary~\ref{theo-cor-qs}, which yields
\begin{equation}\label{app:corr-theo-5-1}
\Vert m_{2n}-\overline{m}\Vert\leq  c_{\theta}~\rho_{\theta}^{n/2}~\Vert m_{0}-\overline{m}\Vert\quad
\mbox{and}\quad
\Vert \sigma_{2n}-\overline{\sigma}\Vert\leq  c_{\theta}~\rho_{\theta}^{n}~\Vert \sigma_{0}-\overline{\sigma}\Vert.
\end{equation}
Based on the same 2D example above, Fig.~\ref{fig:rates} illustrates these rates. In particular, the simulation shows that the theoretical rates exactly match the empirical behaviour of the method for various quantities (in particular, the quantities in the l.h.s. of the inequalities in \eqref{app:theo-5-1} and \eqref{app:corr-theo-5-1}).

\subsection*{Comparison of regularization effects}

\begin{figure}[t]
    \centering
    \includegraphics[width=0.43\textwidth]{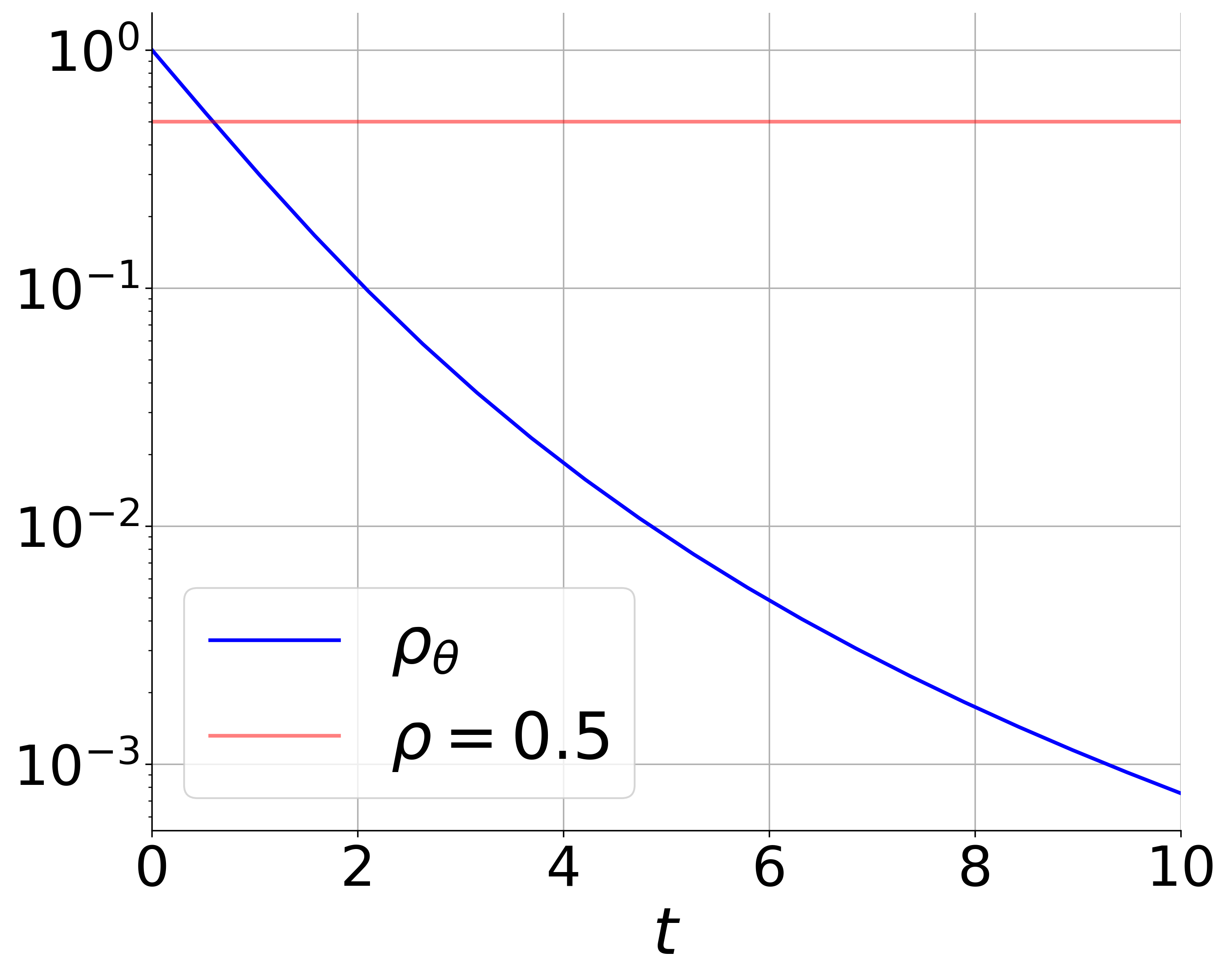}
    \hfill
    \includegraphics[width=0.47\textwidth]{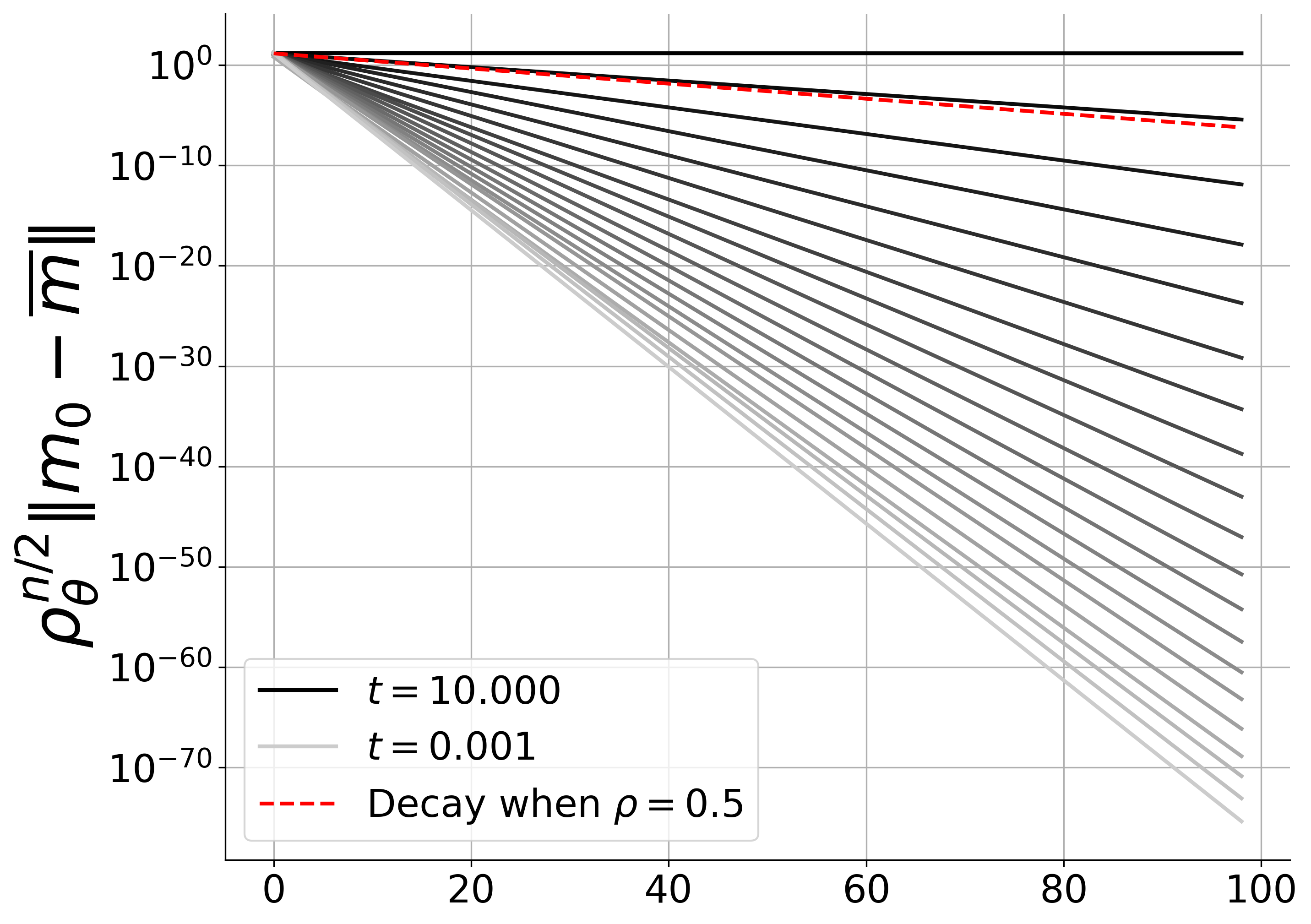}
    \caption{On the left, we demonstrate the value of $\rho_\theta$ we obtain w.r.t. the regularization parameter $t$. It can be seen that $\rho_\theta$ decays to $0$ exponentially fast, compared to the rate $1/2$ found in \cite{chiarini}. On the right, we demonstrate the convergence bound $\rho_\theta^{n/2}\|m_0 - \overline{m}\|$ with our $\rho_\theta$ estimates vs. $\rho = 0.5$.}
    \label{fig:rates_rho}
\end{figure}

Next, we provide a numerical demonstration of the regularization effect estimates, in light of Remark~\ref{comp-rmk}. Let $\tau_0 = t I$. As mentioned in Remark~\ref{comp-rmk}, the rates presented in \cite{chiarini} correspond to a rate $\lim_{t\to\infty }\rho_{\theta(t)} = 1/2$. We now demonstrate a comparison of the rates we obtain with this asymptotic rate. This is demonstrated in Fig.~\ref{fig:rates_rho}. One can see that, as $t$ grows, our coefficient $\rho_\theta$ which controls the speed of convergence decays to $0$ exponentially fast. We also plot the curve and bounds with $\rho = 1/2$ that is the asymptotic rate obtained in \cite{chiarini}. It can be seen that the rates we obtain are sharper than the ones obtained in \cite{chiarini}.

\subsection*{Contraction coefficient for degenerate covariance matrices}

{The contraction coefficient $\rho_\theta = \left( 1 + \lambda_{\rm min}(r_\theta+\varpi_\theta) \right)^{-2}$ in Theorem~\ref{theo-qs} depends directly on the smallest eigenvalue of the matrix $r_\theta+\varpi_\theta$, where $r_\theta$ is the fixed point of the Riccati difference equation characterized by $\varpi_\theta$. The eigenvalue $\lambda_{\rm min}(r_\theta+\varpi_\theta)$ can be written explicitly in terms of the parameter $\theta$ and the covariance matrices $\sigma$ and $\bar \sigma$, as shown by Eqs. \eqref{def-w-1}, \eqref{def-r-t} and \eqref{def-rho-t}.}

\begin{figure}[t]
    \centering
    \includegraphics[width=0.47\textwidth]{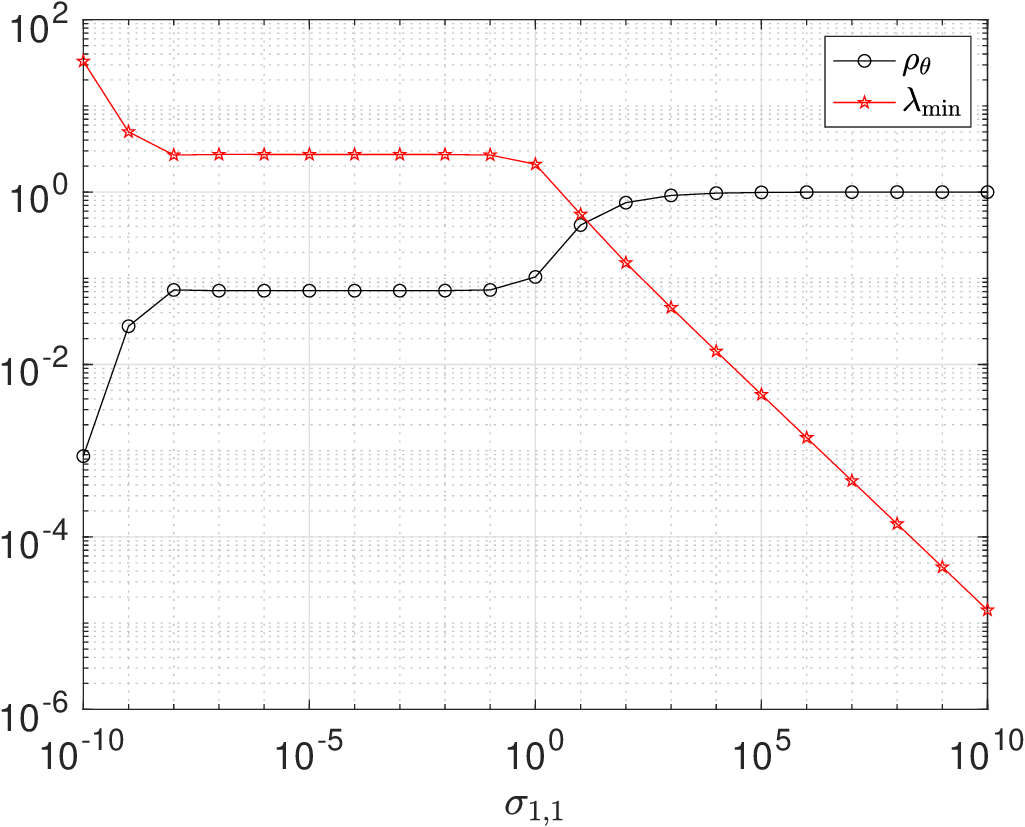}
    \hfill
    \includegraphics[width=0.48\textwidth]{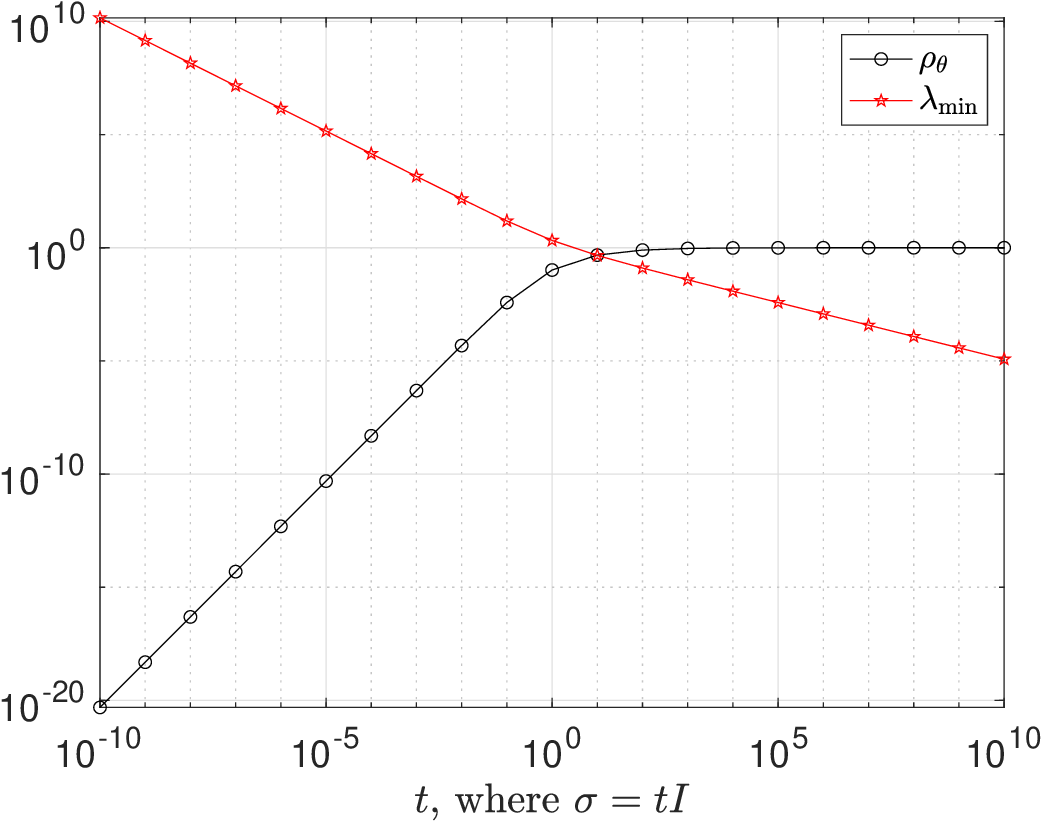}
    \caption{Contraction coefficient $\rho_\theta$ and eigenvalue $\lambda_{\rm min}(r_\theta+\varpi_\theta)$ for $
    \sigma=\left[ 
    	\begin{array}{cc} 
		t &0\\ 
		0 &1\\ 
		\end{array} 
	\right]$ (left) and $\sigma = tI$ (right). The reference parameter $\theta=(\alpha,\kappa,\tau)$ for this simulation is given by $\alpha=[0,0]'$, $\kappa = I_2$ and $\tau=I_2$.}
    \label{fig:rho}
\end{figure}

{Figure \ref{fig:rho} (left) displays the values of both $\rho_\theta$ and $\lambda_{\rm min}(r_\theta+\varpi_\theta)$ as the covariance matrix $\sigma = \left[
	\begin{array}{cc}
	t &0\\
	0	&1\\
	\end{array}
\right]$ changes with parameter $t = 10^{-10}, \ldots, 10^{10}$, while $\bar\sigma=\left[
	\begin{array}{cc}
	\frac{1}{2}	&-\frac{1}{5}\\
	-\frac{1}{5} &\frac{1}{2}\\
	\end{array}
\right]$ is kept fixed. Note that $\rho_\theta$ determines the speed of convergence of the bridge from $\nu_{m,\sigma}$ to $\nu_{\bar m,\bar \sigma}$. For $t>10$, the eigenvalue $\lambda_{\rm min}(r_\theta+\varpi_\theta)$ decreases quickly towards 0, leading to $\rho_\theta \to 1$ (and slow contraction). On the other hand, both the eigenvalue and the contraction coefficient remain relatively stable, $\rho_\theta\approx 0.07$, for a large range of values of $t$ ($10^{-8} \le t \le 10^{-1}$, approximately), but falls sharply for $t < 10^{-8}$.}

{Figure \ref{fig:rho} (right) displays, again, the values of both $\rho_\theta$ and $\lambda_{\rm min}(r_\theta+\varpi_\theta)$ as the covariance matrix $\sigma = tI$ changes with $t = 10^{-10}, \ldots, 10^{10}$. The covariance matrix of the target distribution is kept fixed, with the same value as in Fig. \ref{fig:rho} (left). In this case, there is no plateau either in $\lambda_{\rm min}(r_\theta+\varpi_\theta)$ or $\rho_\theta$. The contraction coefficient is $\rho_\theta \approx 1$ for $t \ge 10^2$ and decreases steadily towards 0 as $t \to 0$.}

\subsection*{Approximate transport of non-Gaussian distributions}

{In filtering theory, Gaussian filters are often used as suboptimal approximations of (intractable) nonlinear filters. Similarly, one can use the optimal Gaussian Schr\"odinger bridges to perform {\em approximate} transport between non-Gaussian distributions $\eta(dx)$ and $\mu(dy)$. To be specific, let $X \sim \eta$ and $Y\sim \mu$ be random variables with $\EE(X)=m$, $\text{Cov}(X)=\sigma$, $\EE(Y)=\bar m$ and $\text{Cov}(Y)=\bar \sigma$. We can solve the optimal Schr\"odinger bridge from the Gaussian distribution $\nu_{m,\sigma}$ to the Gaussian distribution $\nu_{\bar m,\bar \sigma}$, with reference parameter $\theta=(\alpha,\beta,\tau)$. In particular, as shown in Theorem \ref{Th1}, there are optimal parameters $\SB(\theta)=(\iota_\theta,\kappa_\theta,\varsigma_\theta)$ such that if $\hat X \sim \nu_{m,\sigma}$ then the transformation
\begin{equation}
\hat Y = \iota_\theta + \kappa_\theta \hat X + \varsigma_\theta^{\frac{1}{2}} G, 
\quad G \sim \nu_{0,I},
\label{eq_approx_1}
\end{equation}
yields $\hat Y \sim \nu_{\bar m,\bar \sigma}$ while solving problem \eqref{opt}. It is straightforward to check that the linear-Gaussian map \eqref{eq_approx_1} above preserves the target mean $\bar m$ and covariance $\bar \sigma$, i.e., if one chooses $X \sim \eta$, where $\eta \ne \nu_{m,\sigma}$ but $\EE(X)=m$ and $\text{Cov}(X)=\sigma$, then the random variable
$$
Y = \iota_\theta + \kappa_\theta X + \varsigma_\theta^{\frac{1}{2}} G, \quad G \sim \nu_{0,I},
$$ 
still has $\EE(Y)=\bar m$ and $\text{Cov}(Y)=\bar\sigma$, even if $Y \nsim \mu$. Therefore, the map \eqref{eq_approx_1} can be used to perform approximate transport of non-Gaussian distributions in the sense that it attains the target mean and covariance of $\mu$. 
} 

\begin{figure}[t]
    \centering
    \includegraphics[width=0.48\textwidth]{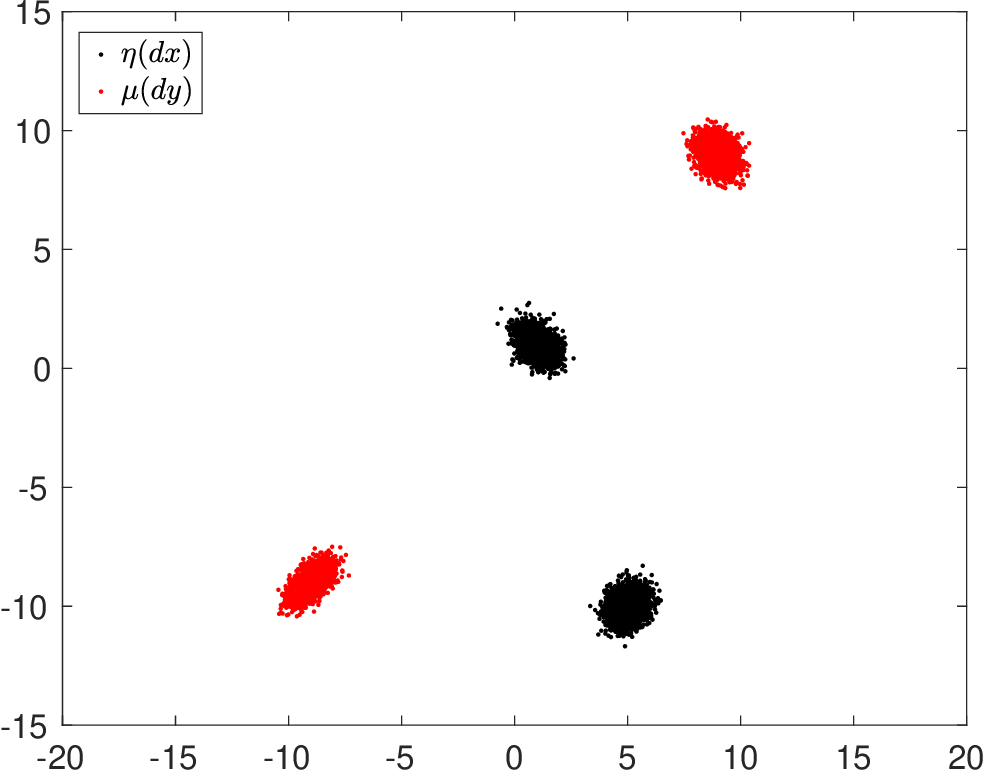}
    \hfill
    \includegraphics[width=0.48\textwidth]{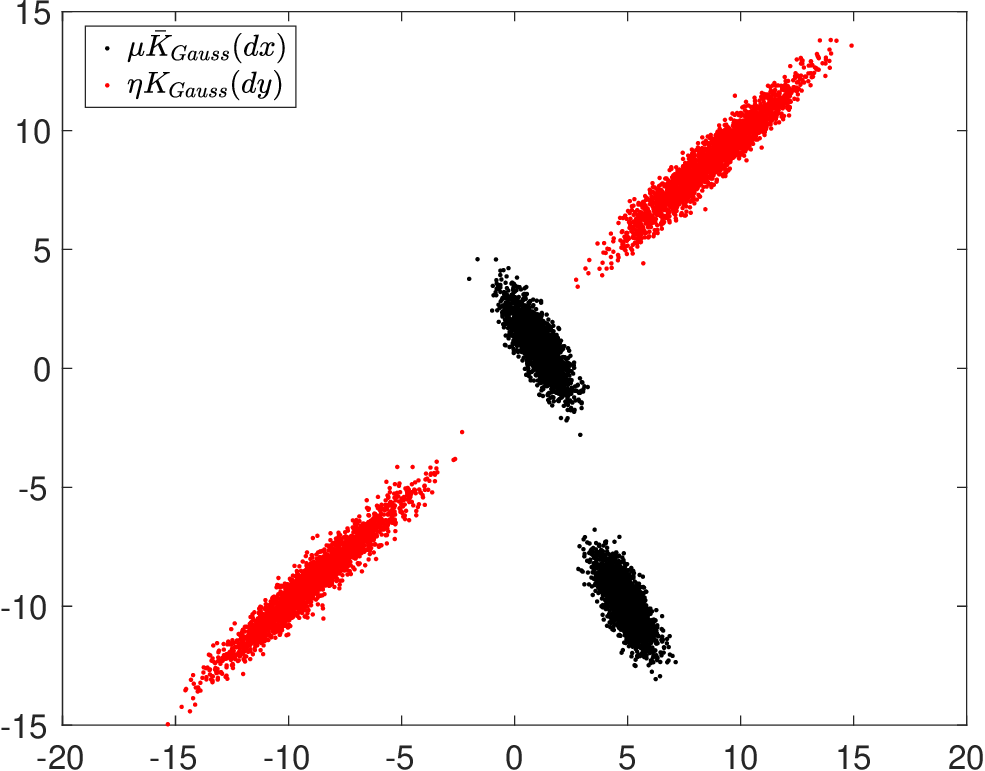}
    \caption{Left: Gaussian mixture distributions $\eta(dx)$ (red) and $\mu(dy)$ (blue). Right: approximate transport with the kernels $K_{Gauss}$ and $\bar K_{Gauss}$ that solve the Gaussian entropic optimal problem for normal distributions $\nu_{m,\sigma}(dx)$ (with the same mean and covariance as $\eta(dx)$) and $\nu_{\bar m,\bar\sigma}(dy)$ (with the same mean and covariance as $\mu(dy)$). The reference parameter is $\theta=(0,I,I)$.}
    \label{fig:mix}
\end{figure}

{
Figure \ref{fig:mix} (left) displays scatter plots of two Gaussian mixture distributions, namely 
$$
\eta(dx) = \frac{1}{2}\sum_{i=1}^2 \nu_{m_i,\sigma_i}(dx)
\quad \text{and} \quad
\mu(dy)=\frac{1}{2}\sum_{i=1}^2 \nu_{\bar m_i,\bar \sigma_i}(dy),
$$ 
where 
$$
m_1 = \left[
	\begin{array}{c}
	1\\
	1\\
	\end{array}
\right], ~~
m_2 = \left[
	\begin{array}{c}
	5\\
	-10\\
	\end{array}
\right], ~~
\sigma_1 = \left[
	\begin{array}{cc}
	\frac{1}{5} &-\frac{2}{25}\\
	-\frac{2}{25} &\frac{1}{5}\\
	\end{array}
\right] ~~\text{and}~~
\sigma_2 = \left[
	\begin{array}{cc}
	\frac{1}{5} &\frac{1}{25}\\
	\frac{1}{25} &\frac{1}{5}\\	
	\end{array}
\right],
$$
while
$$
\bar m_1 = \left[
	\begin{array}{c}
	-9\\
	-9\\
	\end{array}
\right], ~~
\bar m_2 = \left[
	\begin{array}{c}
	9\\
	9\\
	\end{array}
\right], ~~
\bar \sigma_1 = \left[
	\begin{array}{cc}
	\frac{1}{5} &-\frac{3}{25}\\
	-\frac{3}{25} &\frac{1}{5}\\
	\end{array}
\right] ~~\text{and}~~
\bar \sigma_2 = \left[
	\begin{array}{cc}
	\frac{1}{5} &-\frac{1}{25}\\
	-\frac{1}{25} &\frac{1}{5}\\	
	\end{array}
\right].
$$
Figure \ref{fig:mix} (right) shows how $\eta$ and $\mu$ are reshaped by the linear-Gaussian kernels $K_{Gauss}$ and $\bar K_{Gauss}$, respectively, that yield the entropic optimal transport $\nu_{\bar m,\bar\sigma} = \nu_{m,\sigma}K_{Gauss}$ and $\nu_{m,\sigma} = \nu_{\bar m,\bar \sigma}\bar K_{Gauss}$ with reference parameters $\theta = (0,I,I)$. We immediately observe that $\eta K_{Gauss} \ne \mu$ and $\mu\bar K_{Gauss} \ne \eta$. Only the mean and covariance are matched. In practice, $\eta K_{Gauss}$ is a mollified version of $\mu$ and $\mu\bar K_{Gauss}$ is a mollified version of $\eta$.}

\section{Discussion} \label{sec:Conclusions}

\subsection{Summary}

This paper provides a self-contained analysis of the Sinkhorn iterations and Schr\"odinger bridges for general Gaussian models. It includes a complete characterization of the Sinkhorn distribution flow $\Pa_n$ and the associated Schr\"odinger potentials $(U_n,V_n)$ in terms of Riccati matrix difference equations. To our best knowledge, this is the first finite-dimensional description of Sinkhorn iterations on non-finite spaces.

For the analysis, we have leveraged a novel closed-form solution of the fixed points of the Riccati equation\footnote{It is worth noting that, using Brascamp-Lieb and Cramer-Rao inequalities, the analysis of log-concave models developed in~\cite{dm-25,dm-25-2} follows the same Riccati analysis and it is based on the same closed form solution of the fixed points (see for instance Appendix A in~\cite{dm-25}).} (see (\ref{def-fix-ricc-1})) and further developed the Floquet-type representation of Riccati flows discussed in~\cite{dh-22} to obtain (sharp) exponential convergence rates for the Sinkhorn iterates. The quantitative estimates derived in Section~\ref{quant-sec} are sharper than any known exponential stability rates discussed in the literature on log-concave models \cite{chiarini,durmus,dm-25,dm-25-2} (see also Remark~\ref{comp-rmk} in the context of regularized models).  As an extension of these results, we have also analyzed, in Section~\ref{sec-schrod}, the stability properties of a class of Gibbs loop-type non-homogeneous Markov chains associated to the Sinkhorn iterations for general, possibly non-Gaussian, models. These properties have been further developed in~\cite{adm-25} to analyze the contraction properties of Sinkhorn semigroups.

Finally, we have investigated the class of regularized Gaussian models parameterized as $\theta(t)=(\alpha,\beta,tI)$ and obtained
\begin{itemize}
\item convergence rates of the bridge transport maps (and Schr\"odinger potentials) towards independent Gaussians as $t\to\infty$, and
\item convergence rates of the Gaussian bridge transport maps (and Schr\"odinger bridge measures) towards Monge maps as $t\to 0$.
\end{itemize}     
Remarkably, most of the literature on entropic transport problems only deals with the case $(\alpha,\beta,\tau)=(0,I,tI)$, i.e., with symmetric quadratic-type costs. This parameterization excludes the important cases when the reference measure in (\ref{def-entropy-pb-v2}) and (\ref{sinhorn-entropy-form}) is associated with linear Gaussian transitions arising in Ornstein-Uhlenbeck-type diffusion generative models, denoising diffusions, or flow-matching schemes (cf. for instance Section 3 in~\cite{dm-25}). The strength of our approach is that it is applicable to a large class of linear Gaussian models arising in machine learning and artificial intelligence algorithms --see Remark~\ref{ref-general-models} {as well as Section~\ref{sec-OU-bridges} dedicated to static and dynamic Ornstein-Uhlenbeck bridges}.

\subsection{Entropic optimal transport vs. Bayesian filtering}

While in statistical inference, signal processing and optimal control theory linear Gaussian models have been considered of fundamental importance, the entropic optimal transport community have paid comparably much less attention to them. Instead, most of the research has focused on the regularity properties of (general) Schr\"odinger bridges and the stability properties of the Sinkhorn recursions --which cannot be exactly solved unless the state is finite. 

It may be useful indeed to draw a parallel with Bayesian inference and filtering theory. The iterative proportional fitting procedure (Sinkhorn algorithm) for matrices~\cite{ruschen,sinkhorn,sinkhorn-2,sinkhorn-3} solves Bayes' formula using matrix operations, in essentially the same way as the well-known Wonham filter~\cite{wonham} (see also~\cite{kim} and references therein). In a similar manner, the Gaussian Sinkhorn algorithm analyzed in the present paper is based on the same linear regression formulae that solve the Bayes' rule for linear Gaussian state-space models and yield the celebrated Kalman filter. 

For more general settings, both Sinkhorn iterations and the nonlinear filtering equation involve sequential Bayes' updates that do not admit exact, finite-dimensional solutions. The complexity of sampling from Bayes' posterior distributions (a.k.a. dual or backward transitions) is also a key, well-known technical problem in Bayesian statistics and machine learning. For non-conjugate models, it requires to introduce an additional level of numerical approximation~\cite{damlen1999gibbs}. In Bayesian filtering, such approximations include particle filters \cite{Gordon93,Kitagawa96} (see also \cite{Liu98,Doucet00,Djuric03,DelMoral04}), ensemble Kalman filters \cite{Evensen96,Evensen03}, Gaussian sum filters \cite{Ito00}, 
Bayesian nested sampling~\cite{cai,skilling}, gradient-guided nested sampling~\cite{lemos}, etc. Similar numerical strategies can be developed for the efficient numerical (approximate) implementation of non-Gaussian Sinkhorn algorithms. 

\subsection{Complexity}

We have introduced a finite-dimensional description of Sinkhorn iterations in terms of Riccati matrix difference equations, including a closed-form solution of the limiting Schr\"odinger bridges for general Gaussian models. It is important to realize, however, that these objects are expressed in terms of matrix square roots and matrix inversions. The efficient computation of these quantities is a well-known bottleneck in solving high-dimensional linear-Gaussian filtering problems and linear-quadratic optimal control problems, see for instance~\cite{benner,massioni} and references therein. Specifically, the computational complexity of evaluating the square root, or the inverse, of a $d \times d$ matrix is $\mathcal{O}(d^3)$ in practice. In many real-world domains, the models of interest (e.g., graphical models, complex networks, etc.) often contain millions of nodes~\cite{maji}.
This renders the exact computation of matrix inversions and square roots infeasible in practical terms. In these high dimensional problems, the Sinkhorn algorithm as well as the explicit formulae for Schr\"odinger bridges for Gaussian models presented in this article cannot be solved on a computer without approximations. Several numerical approximation methods can be used, e.g., power-series expansion, Denman-Beavers square root iteration~\cite{denman} and stochastic/randomized algorithms~\cite{gower}. 

The availability of explicit solutions for the Gaussian Sinkhorn iterations and Schr\"odinger bridges for general (possibly very large) Gaussian models can be an incentive for the development of efficient approximation methods. In particular, it may be of interest to compare the accuracy and complexity of numerical approximations, built upon the formulae for exact solutions, with machine learning approximations based on neural networks and score-based optimization.

\subsection{Extended entropic projection methods}\label{sec-extended}

The connection between Schr\"odinger bridges and diffusion models has been highlighted before \cite{bunne2023schrodinger}. Indeed, Gaussian Schr\"odinger bridges and Sinkhorn algorithms can be formulated in a parametric variational form, similar to diffusion and other generative models that rely on the computation of scores \cite{doucet-bortoli}, to approximate Sinkhorn recursions (see (\ref{opt})) and Theorem~\ref{Th1}). To be specific, 
note that given the Sinkhorn bridge $\Pa_{2n}=(\eta\times\Ka_{2n})$, for any Markov transition $L$ we have the entropic formula
\begin{equation}\label{sinhorn-entropy-form-conc}
\mbox{\rm Ent}((\mu\times L)^{\flat}~|~\eta\times\Ka_{2n})=
\mbox{\rm Ent}(\mu~|~\eta\Ka_{2n})+\mbox{\rm Ent}(\mu\times L~|~\mu\times\Ka_{2n+1}).
\end{equation}
Recall from (\ref{s-2}) that 
 $\Ka_{2n+1}=\Ka_{2n}^{\sharp}$ coincides with the dual transition $\Ka_{2n}^{\sharp}$ associated with $\Ka_{2n}$ defined by
$$
(\eta\Ka_{2n})(dy)~\Ka_{2n}^{\sharp}(y,dx):=\eta(dx) \Ka_{2n}(x,dy).
$$
In the same vein, given the Sinkhorn bridge $\Pa_{2n+1}=(\mu\times\Ka_{2n+1})^{\flat}$, we have
\begin{equation}\label{sinhorn-entropy-form-conc-odd}
\mbox{\rm Ent}(\eta\times L~|~(\mu\times\Ka_{2n+1})^{\flat})=
\mbox{\rm Ent}(\eta~|~\mu\Ka_{2n+1})+\mbox{\rm Ent}(\eta\times L~|~\eta\times\Ka_{2(n+1)}).
\end{equation}
In this notation, Sinkhorn iterates are defined as follows:
\begin{itemize}
\item Given $\Pa_{2n}$, the optimal coupling $(\mu\times L)^{\flat}\in \Ca_Y(\mu)$ in (\ref{sinhorn-entropy-form-conc}) is obtained by
choosing $L=\Ka_{2n+1}$, so that $\Pa_{2n+1}=(\mu\times \Ka_{2n+1})^{\flat}$. 
\item Given $\Pa_{2n+1}$
the optimal coupling $(\eta\times L)\in \Ca_X(\eta)$ in
 (\ref{sinhorn-entropy-form-conc-odd})  is obtained by
choosing $L=\Ka_{2(n+1)}$, so that $\Pa_{2(n+1)}=(\eta\times \Ka_{2(n+1)})$. 
\end{itemize}
For Gaussian models we have seen in (\ref{ref-mean-cov-intro}) that 
$\Ka_{2n+1}=K_{\theta_{2n+1}}$, thus  the optimal coupling transition
$L=K_{\theta}$ in (\ref{sinhorn-entropy-form-conc}) is obtained by
choosing $\theta=\theta_{2n+1}$. Since
$\Ka_{2(n+1)}=K_{\theta_{2(n+1)}}$,  the optimal coupling
$L=K_{\theta}$ in (\ref{sinhorn-entropy-form-conc-odd}) is obtained by
choosing $\theta=\theta_{2(n+1)}$. 

In summary, the Gaussian Sinkhorn iterates are defined as
\begin{eqnarray}
\mbox{\rm Ent}((\mu\times K_{\theta_{2n+1}})^{\flat}~|~\eta\times\Ka_{2n})&=&\inf_{\theta\in\Theta} \mbox{\rm Ent}((\mu\times K_{\theta})^{\flat}~|~\eta\times\Ka_{2n})~~\text{and}\label{sinhorn-entropy-form-conc-para}
\\
\mbox{\rm Ent}(\eta\times K_{\theta_{2(n+1)}}~|~(\mu\times\Ka_{2n+1})^{\flat})&=&
\inf_{\theta\in\Theta}\mbox{\rm Ent}(\eta\times K_{\theta}~|~(\mu\times\Ka_{2n+1})^{\flat}).~\label{sinhorn-entropy-form-conc-22-para}
\end{eqnarray}

For non-Gaussian models, we can choose a judicious set of (non necessarily linear Gaussian)  Markov transitions $K_{\theta}$ indexed by some parameter $\theta$ on some parameter space $\Theta$. {For instance for multimodal marginals it may be judicious to choose mixtures of non necessarily linear  Gaussian transitions}. In this context, the optimal couplings defined sequentially by the above recursion are not unique (unless the set of transitions $K_{\theta}$ is convex).
For instance, given $\Ka_{2n}=K_{\theta_{2n}}$, as in (\ref{sinhorn-entropy-form-conc-para}) we choose the coupling transition  $\Ka_{2n+1}=K_{\theta_{2n+1}}$ by the formulae
\begin{eqnarray}
\mbox{\rm Ent}((\mu\times K_{\theta_{2n+1}})^{\flat}~|~\eta\times\Ka_{2n})&=&\inf_{\theta\in\Theta} \mbox{\rm Ent}((\mu\times K_{\theta})^{\flat}~|~\eta\times\Ka_{2n})\label{sinhorn-entropy-form-conc-para-g}\\
&=&\mbox{\rm Ent}(\mu~|~\eta K_{\theta_{2n}})+\inf_{\theta\in\Theta}\mbox{\rm Ent}(\mu\times K_{\theta}~|~\mu\times K_{\theta_{2n}}^{\sharp}).\nonumber
\end{eqnarray}
Note that
$
\inf_{\theta\in\Theta}\mbox{\rm Ent}(\mu\times K_{\theta}~|~\mu\times K_{\theta_{2n}}^{\sharp})=0
\Longleftrightarrow \exists \theta\in\Theta~~\mbox{\rm such that $K_{\theta}=K_{\theta_{2n}}^{\sharp}$}.
$

Whenever $K_{\theta_{2n+1}}=K_{\theta_{2n}}^{\sharp}$ we recover Sinkhorn recursion.
Otherwise the class of Markov transitions $K_{\theta}$ is too small and the projection (\ref{sinhorn-entropy-form-conc-para-g}) introduces an entropic bias $\mbox{\rm Ent}(\mu\times K_{\theta_{2n+1}}~|~\mu\times K_{\theta_{2n}}^{\sharp})>0$.
This entropic projection method is clearly related to parametric score-based methods often used to approximate the backward transition $K_{\theta_{2n}}^{\sharp}$ associated with the forward transition $K_{\theta_{2n}}$  in generative modeling. In this respect, we view the work on the Gaussian Sinkhorn iterations in this paper as a step towards the analysis of the mathematical foundations of a more general class of parametric models, including denoising diffusions \cite{doucet-bortoli}. Future work will include the effect of the bias in the entropic projection method of (\ref{sinhorn-entropy-form-conc-para-g}).

\section*{Acknowledgements}

This project originated during Pierre Del Moral’s visit to Imperial College London as a Nelder Visiting Fellow in March and April 2024. Pierre Del Moral is grateful to Valentin de Bortoli, George Deligiannidis and Arnaud Doucet for valuable discussions on the Sinkhorn algorithm. This work is also supported
by the Innovation and Talent Base for Digital Technology and Finance (B21038) and ‘the Fundamental Research Funds for the Central Universities’, Zhongnan University of Economics and Law
(2722023EJ002).

Joaquin Miguez acknowledges the financial support of the Office of Naval Research (award no. N00014-22-1-2647) and MICIU/AEI/10.13039/501100011033/FEDER UE (grants no. PID2021-125159NB-I00 TYCHE and PID2024-158181NB-I00 NISA).


\appendix
\section*{Appendix}

\section{Riccati difference equation}\label{app-ricc}

\subsection*{Some terminology}

We associate with some given $\varpi\in\Sa^+_d$ the matrix recursions
\begin{equation}\label{ricc-def-app}
u_{n+1}:=\mbox{\rm Ricc}^-_{\varpi}(u_{n})
\quad \mbox{\rm and}\quad
v_{n+1}=\mbox{\rm Ricc}_{\varpi}(v_n)
\end{equation}
with the increasing maps  $\mbox{\rm Ricc}^-_{\varpi}$ and $\mbox{\rm Ricc}_{\varpi}$ from $\Sa^0_d$ into itself defined by
$$
\mbox{\rm Ricc}^-_{\varpi}(u)=\varpi+(I+u)^{-1}u
\quad \mbox{\rm and}\quad
\mbox{\rm Ricc}_{\varpi}(v):=(I+(\varpi+v)^{-1})^{-1}.
$$
Observe that
\begin{equation}\label{ricc-monotone}
\begin{array}{rccrcl}
\mbox{\rm Ricc}^-_{\varpi}(0)=\varpi&\preceq& \mbox{\rm Ricc}^-_{\varpi}(u)&\preceq&\varpi+I
\quad \text{and}
\\
\mbox{\rm Ricc}_{\varpi}(0)=(I+\varpi^{-1})^{-1}&\preceq& \mbox{\rm Ricc}_{\varpi}(v)&\preceq& I.
\end{array}
\end{equation}
This shows that for any $n\geq 1$ and any $u_0,v_0\in\Sa^0_d$ we have
\begin{equation}\label{estimate-app-ricc}
\varpi\preceq u_n\preceq\varpi+I\quad \mbox{\rm and}\quad
(I+\varpi^{-1})^{-1}\preceq v_n\preceq I.
\end{equation}
These inequalities yield, for any $n\geq 1$, the upper bound
$$
v_{n+1}=(I+(\varpi+v_n)^{-1})^{-1}\preceq (I+(\varpi+I)^{-1})^{-1}=I-(I+(\varpi+I))^{-1}
$$
and, therefore, the estimates below.
\begin{lem}\label{lem-estimate-app-ricc-2}
For any $v\geq 0$ we have
\begin{equation}\label{estimate-app-ricc-im}
(I+\varpi^{-1})^{-1}\preceq v_{n+1}\preceq (I+(\varpi+I)^{-1})^{-1}
\end{equation}
or, equivalently,
\begin{equation}\label{estimate-app-ricc-2}
(I+(\varpi+I))^{-1}\preceq I-v_{n+1}\preceq I-(I+\varpi^{-1})^{-1}= (I+\varpi)^{-1}.
\end{equation}
\end{lem}
The r.h.s. assertion in (\ref{estimate-app-ricc-2}) comes from the fact that
$$
I-(I+\varpi^{-1})^{-1}= (I+\varpi^{-1})^{-1}((I+\varpi^{-1})-I)=
 (I+\varpi)^{-1}.
$$
In addition, $u_0\succ 0$ implies that $u_n\succ \varpi$ for every $n\geq 1$ and, in the same vein, $v_0\succ 0$ yields
$$
v_n\succ (I+\varpi^{-1})^{-1}\quad \mbox{\rm and}\quad I-v_n \prec (I+\varpi)^{-1} \quad \text{for every $n \ge 1$.}
$$
Recalling that the Riccati flow starting at the null matrix is increasing while le one starting at $I$ is decreasing we readily find the following estimates.
\begin{lem}
For any $n\geq 1$ and $v\geq 0$ we have
$$
0\preceq \mbox{\rm Ricc}_{\varpi}^{n-1}(0)\preceq \mbox{\rm Ricc}_{\varpi}^{n}(0)\preceq \mbox{\rm Ricc}_{\varpi}^{n}(v)\preceq \mbox{\rm Ricc}_{\varpi}^{n-1}(I)\preceq I
$$
as well as
$$
 \mbox{\rm Ricc}_{\varpi}^{n}(v)\preceq \mbox{\rm Ricc}_{\varpi}^{n-1}(I).
$$
\end{lem}

The recursions in (\ref{ricc-def-app}) are connected by the inductive formulae
$$
u_n\longrightarrow v_{n}=\left(I+u_n\right)^{-1}u_n\longrightarrow u_{n+1}=v_n+\varpi
\longrightarrow v_{n+1}=\left(I+u_{n+1}\right)^{-1}u_{n+1}.
$$
Starting from $v_0=0$, for every $n\geq 0$ we also have
$$
v_n\longrightarrow u_{n+1}:=v_n+\varpi
\longrightarrow v_{n+1}=\left(I+u_{n+1}^{-1}\right)^{-1}\succeq (I+\varpi^{-1})^{-1}.
$$
The assertions above are easily verified if we note that
$$
v_{n+1}=\left(I+u_{n+1}^{-1}\right)^{-1}=\left(I+\left(\varpi+(I+u_n^{-1})^{-1}\right)^{-1}\right)^{-1}=
\left(I+\left(\varpi+v_n\right)^{-1}\right)^{-1}
$$
and
$$
u_{n+1}=v_n+\varpi=\varpi+\left(I+u_n^{-1}\right)^{-1}.
$$
For any given $s\in\Sa^+_d$, if we let $\tau_{n}:=s^{1/2}~v_n~s^{1/2}$ then we obtain the recursion
$$
\tau_{n+1}=\mbox{\rm Ricc}_{s,\varpi_s}(\tau_n):=(s^{-1}+( \varpi_{s}+ \tau_n )^{-1})^{-1}, \quad \text{with} \quad \varpi_{s}:=s^{1/2}~ \varpi~s^{1/2}.
$$
{Also note that
\begin{equation}\label{v2u}
v_{n}=\left(I+u_{n}^{-1}\right)^{-1}\Longleftrightarrow I-v_{n}=(I+u_n)^{-1}.
\end{equation}}

\subsection*{Fixed point matrices}\label{Sec-fixed-p}
Riccati matrix difference equations of the form (\ref{ricc-def-app})  are rather well understood (cf.~\cite{dh-22} and references therein).
For instance,  $u_n$ and $v_n$ converge exponentially fast, as $n\rightarrow\infty$, to 
the unique positive definite fixed points 
\begin{equation}\label{ref-fix-app}
u_{\infty}=\mbox{\rm Ricc}^-_{\varpi}(u_{\infty})\quad \mbox{\rm and}\quad
r_{\infty}=\mbox{\rm Ricc}_{\varpi}(r_{\infty}).
\end{equation}
The fixed points are connected by the formula 
\begin{eqnarray*}
r_{\infty}:=(I+u_{\infty}^{-1})^{-1}
\end{eqnarray*}
which implies that
\begin{equation*}
r_{\infty}+\varpi=u_{\infty} 
\quad\text{and}\quad
\mbox{\rm Ricc}_{\varpi}(r_{\infty})=(I+(\varpi+r_{\infty})^{-1})^{-1}=
(I+u_{\infty}^{-1})^{-1}=
r_{\infty}.
\end{equation*}
Also notice that
\begin{eqnarray*}
r_{\infty}=\mbox{\rm Ricc}_{\varpi}(r_{\infty})&\Longrightarrow&
r_{\infty}^{-1}=I+(\varpi+r_{\infty})^{-1}\\
&\Longrightarrow&\varpi r_{\infty}^{-1}+I=
(\varpi+r_{\infty})r_{\infty}^{-1}=(\varpi+r_{\infty})+I\\
&\Longrightarrow&\varpi r_{\infty}^{-1}=\varpi+r_{\infty}
\Longrightarrow r_{\infty}^{-1}=I+\varpi^{-1}r_{\infty},
\end{eqnarray*}
hence we conclude that
\begin{equation}\label{fp-ricc}
I=r_{\infty}+r_{\infty}\varpi^{-1}r_{\infty}.
\end{equation}
More interestingly, the fixed point $r_{\infty}$ can be explicitly computed in terms of $\varpi$. Indeed, we have
\begin{eqnarray*}
r_{\infty}&=&(I+(\varpi+r_{\infty})^{-1})^{-1}\\
&=&\left(
(\varpi+r_{\infty})^{-1}
\left((\varpi+r_{\infty})+I\right)\right)^{-1}=
\left(
(\varpi+r_{\infty})+I\right)^{-1}~(\varpi+r_{\infty}),
\end{eqnarray*}
which implies the equivalence
$$
\left(
(\varpi+r_{\infty})+I\right)r_{\infty}=\varpi+r_{\infty}\Longleftrightarrow
r_{\infty}^2+\varpi r_{\infty}=\varpi.
$$
We may also note that
$$
\varpi=\varpi^{\prime}\quad \mbox{\rm and}\quad
r_{\infty}=r_{\infty}^{\prime}\Longrightarrow
\varpi ~r_{\infty}=r_{\infty}~\varpi 
$$
and, as a consequence,
$$
r_{\infty}^2+\varpi r_{\infty}=\left(r_{\infty}+\frac{\varpi}{2}\right)^2-\left(\frac{\varpi}{2}\right)^2=\varpi.
$$
We summarize the above discussion with the following proposition.
\begin{prop}\label{fix-p}
The unique positive definite fixed points of the matrix equations (\ref{ref-fix-app}) are given by the formulae
$$
r_{\infty}=-\frac{\varpi}{2}+\left(\varpi+\left(\frac{\varpi}{2}\right)^2\right)^{1/2}\quad
\mbox{\rm and}\quad
u_{\infty}=\frac{\varpi}{2}+\left(\varpi+\left(\frac{\varpi}{2}\right)^2\right)^{1/2}.
$$
\end{prop}
Let us also note that
$$
\varpi ~r_{\infty}=r_{\infty}~\varpi \Longleftrightarrow
\varpi^{-1}\left(\varpi+\left(\frac{\varpi}{2}\right)^2\right)^{1/2}\varpi=
\left(\varpi+\left(\frac{\varpi}{2}\right)^2\right)^{1/2},
$$
where the r.h.s. assertion is a direct consequence of the formula
$$
\left(\varpi^{-1}\left(\varpi+\left(\frac{\varpi}{2}\right)^2\right)^{1/2}\varpi\right)^2=\varpi^{-1}\left(\varpi+\left(\frac{\varpi}{2}\right)^2\right)\varpi=\varpi+\left(\frac{\varpi}{2}\right)^2.
$$
Finally observe that, for any given $s\in\Sa^+_d$, we have
\begin{equation}\label{fix-p-2}
r_{s,\infty}:=s^{1/2}~r_{\infty}~s^{1/2}=\mbox{\rm Ricc}_{s,\varpi_s}(r_{s,\infty})
\quad\text{which implies}\quad
r_{s,\infty}+r_{s,\infty}~\varpi_s^{-1}~r_{s,\infty}=s.
\end{equation}

\subsection*{Stability analysis}
By monotone arguments one can show that that 
$$
v_0\leq r_{\infty}\Longleftrightarrow v_n\leq r_{\infty} ~~\forall n\geq 0.
$$
Also for any $u_1,u_2\in \Sa^0_d$ we have
\begin{eqnarray*}
\mbox{\rm Ricc}^-_{\varpi}(u_1)-\mbox{\rm Ricc}^-_{\varpi}(u_2)&=&(I+u_1)^{-1}u_1-(I+u_2)^{-1}u_2\\
&=&(I+u_1)^{-1}(u_1(I+u_2)-(I+u_1)u_2)(I+u_2)^{-1},
\end{eqnarray*}
which yields the formulae
\begin{equation}\label{form-1}
\mbox{\rm Ricc}^-_{\varpi}(u_1)-\mbox{\rm Ricc}^-_{\varpi}(u_2)=\Ea(u_1)~(u_1-u_2)~\Ea(u_2)^{\prime} \quad \mbox{\rm with}\quad \Ea(u):=(I+u)^{-1}.
\end{equation}
Consider the directed matrix product $\Ea_n(u_0)$ defined by
\begin{equation}\label{En-def}
\begin{array}{ll}
 \Ea_{n+1}(u_0)&:=(I+u_{n})^{-1}\ldots (I+u_1)^{-1}(I+u_0)^{-1}, \quad \text{hence}\\
 \\
 \Ea_n(u_{\infty})&=(I+u_{\infty})^{-n}\\
\end{array}
\end{equation}
for the fixed point $u_\infty$. In \eqref{En-def}, $u_n$ solves
the matrix recursion  in the l.h.s. of (\ref{ricc-def-app}) starting from
some $u_0\in \Sa^0_d$. {In terms of the matrices $v_n$ defined in (\ref{ricc-def-app}) using (\ref{v2u}) we have the directed product formula
\begin{equation}\label{En-prod-uv}
 \Ea_{n+1}(u_0):=(I-v_{n})\ldots (I-v_{1})(I-v_{0}).
\end{equation}}
From the discussion above, it follows that 
$$
\text{if} \quad 
u_{\infty} \succ \lambda_{\text{\rm min}}(\varpi)~I
\quad \text{then} \quad
\Vert\Ea_n(u_{\infty})\Vert_2\leq (1+\lambda_{\text{\rm min}}(u_{\infty}))^{-n}\leq (1+\lambda_{\text{\rm min}}(\varpi))^{-n}.
$$
More refined estimates can be obtained using Proposition~\ref{fix-p}.
In our context, the Floquet-type formula presented in Theorem 1.3 in~\cite{dh-22} takes the form given below.
\begin{theo}[\cite{dh-22}]\label{theo-floquet}
For any $n\geq 0$, we have
$$
 \Ea_{n}(u)=(I+u_{\infty})^{-n}~\left(I+(u-u_{\infty})~\GG_n\right)^{-1}\quad
 \text{with}\quad
 \GG_n:=\sum_{0\leq k<n}(I+u_{\infty})^{-(2k+1)}.
$$
\end{theo}
Note that
$$
\lim_{n\to\infty} \GG_n = 
\GG:=(I+u_{\infty})^{-1}~\left(I-(I+u_{\infty})^{-2}\right)^{-1}\succ \GG_n
$$
which, rewritten in a slightly different form, yields
$$
\GG^{-1}_n\succ \GG^{-1}=u_{\infty}+\widehat{u}_{\infty}^{-1},
\quad\text{where}\quad
\widehat{u}_{\infty}^{-1}:=I-(I+u_{\infty})^{-1}=(I+u_{\infty}^{-1})^{-1}\preceq I.
$$
Using  (\ref{ref-fix-app}), we check the fix point equations
$$
\widehat{u}_{\infty}=I+u_{\infty}^{-1}\Longrightarrow
\varpi+\widehat{u}_{\infty}^{-1}=u_{\infty}\Longrightarrow
\widehat{u}_{\infty}=I+\left(\varpi+\widehat{u}_{\infty}^{-1}\right)^{-1},
$$
and using the decomposition
$$
I+(u-u_{\infty})~\GG_n=\left((\GG_n^{-1}-\GG^{-1})+(\GG^{-1}-u_{\infty})+u\right)~\GG_n
$$
we verify that
$$
I+(u-u_{\infty})~\GG_n=\left((\GG_n^{-1}-\GG^{-1})+\widehat{u}_{\infty}^{-1}+u\right)~\GG_n.
$$
This yields the uniform estimates
$$
\Vert \left(I+(u-u_{\infty})~\GG_n\right)^{-1}\Vert_2\leq \Vert \widehat{u}_{\infty}\Vert_2~\Vert \GG^{-1}\Vert_2\leq 
\left(1+\Vert u^{-1}_{\infty}\Vert_2\right)~\left(1+\Vert u_{\infty}\Vert_2\right).
$$
We summarize the above discussion with the following proposition.
\begin{prop}\label{prop-Ea-u}
For any $n\geq 0$ and any $u\in\Sa^0_+$, we have the inequality
$$
\Vert  \Ea_{n}(u)\Vert_2\leq \psi(u_{\infty})~(1+\lambda_{\text{\rm min}}(u_{\infty}))^{-n},
$$
with the parameter
\begin{equation}\label{def-psi}
\psi(u_{\infty}):=\left(1+\Vert u^{-1}_{\infty}\Vert_2\right)~\left(1+\Vert u_{\infty}\Vert_2\right).
\end{equation}
\end{prop}

Denote by $u_n$ and $\overline{u}_n$ 
the solutions of the matrix recursion defined in the l.h.s. of (\ref{ricc-def-app}) starting from
some $u_0, \overline{u}_0\in \Sa^0_d$, respectively. 
Using (\ref{form-1}), for any $u_0,\overline{u}_0\in \Sa^0_d$ we have 
\begin{equation}\label{phi-form}
u_n-\overline{u}_n=\Ea_n(u_0)\,(u_0-\overline{u}_0)\,\Ea_n(\overline{u}_0)^{\prime},
\end{equation}
which yields the estimate
$$
\Vert u_n-\overline{u}_n\Vert_2\leq \psi(u_{\infty})^2~(1+\lambda_{\text{\rm min}}(u_{\infty}))^{-2n}~
\Vert u_0-\overline{u}_0\Vert_2.
$$
Similarly, denote by $v_n$ and $\overline{v}_n$ 
the solutions of the matrix recursion defined in the r.h.s. of (\ref{ricc-def-app}) starting from
some $v_0$ and $\overline{v}_0\in \Sa^0_d$, respectively. We note that
\begin{eqnarray*}
v_n-\overline{v}_n&=&\left(I+u_n\right)^{-1}u_n-\overline{u}_n\left(I+\overline{u}_n\right)^{-1}\\
&=&
\left(I+u_n\right)^{-1}\overline{u}_n^{-1}(u_n-\overline{u}_n)~u_n^{-1}\left(I+\overline{u}_n\right)^{-1},
\end{eqnarray*}
where $u_n$ and $\overline{u}_n$ are solutions of the matrix recursion defined in the l.h.s. of (\ref{ricc-def-app}) starting from some $u_1=v_0+\varpi$ and $\overline{u}_1=\overline{v}_0+\varpi\in \Sa^0_d$, respectively, at rank $n=1$. The above decomposition combined with the estimates (\ref{estimate-app-ricc})  yields the following result.

\begin{prop}\label{prop-cv-app}
For every $n\geq 0$, we have the exponential estimates
\begin{equation}\label{cv-ricc}
 \Vert\mbox{\rm Ricc}_{\varpi}^{n+1}(v_0)-\mbox{\rm Ricc}_{\varpi}^{n+1}(\overline{v}_0)\Vert_2\leq
 \varphi_{\varpi}(u_{\infty})^2~(1+\lambda_{\text{\rm min}}(u_{\infty}))^{-2n}~
\Vert v_0-\overline{v}_0\Vert_2
 \end{equation}
 with the semigroup $\mbox{\rm Ricc}_{\varpi}^{n+1}:=\mbox{\rm Ricc}_{\varpi}\circ \mbox{\rm Ricc}_{\varpi}^{n}$ and the parameter
 $$
 \varphi_{\varpi}(u_{\infty}):= (1+\lambda_{\text{\rm min}}(\varpi))^{-1}~
~ \lambda_{\text{max}}(\varpi)\psi(u_{\infty}).
 $$
 In the above display, $u_{\infty}$ stands for the fixed point matrix  defined in Proposition~\ref{fix-p} and $\psi(u_{\infty})$ is the parameter defined in (\ref{def-psi}).
 \end{prop}
 
Using (\ref{estimate-app-ricc}) we see that
$$
\psi(u_{\infty})\leq \left(1+(1+\lambda_{\text{\rm min}}(\varpi))^{-1}\right)~\left(2+\lambda_{\text{max}}(\varpi)\right),
$$
which yields the rather crude estimate
 $$
 \varphi_{\varpi}(u_{\infty})\leq 
~~\left(1+(1+\lambda_{\text{\rm min}}(\varpi))^{-1}\right)^2~\left(1+\lambda_{\text{max}}(\varpi)\right)^2.
 $$

\subsection*{Proof of Theorem~\ref{th-2}}\label{th-2-proof}

Formula   (\ref{hom}) yields the matrix Riccati difference equations
$$
\tau_{2n+1}^{-1}=\sigma^{-1}+ \left(\tau_{2n}^{-1}\beta_{2n}\right)^{\prime}~\tau_{2n} ~\left(\tau_{2n}^{-1}\beta_{2n}\right)=\sigma^{-1}+ \cchi_{\theta}^{\prime}~\tau_{2n} ~\cchi_{\theta}
$$
with the parameter $\cchi_{\theta}$ defined in (\ref{def-w-1}) so that
$$
 \gamma_{\theta}=\overline{\sigma}^{1/2}~\cchi_{\theta}~\sigma^{1/2}.
$$
In the same vein, using (\ref{hom}) we obtain
$$
\tau_{2(n+1)}^{-1}=\overline{\sigma}^{-1}+ \cchi_{\theta}
~\tau_{2n+1}~
 \cchi_{\theta}^{\prime}.
$$
This ends the proof of (\ref{k-ricc-0}) up to a rescaling. 

Next, observe that
\begin{equation}\label{proof-ricc}
u=(I+\gamma^{\prime}v\gamma)^{-1}
\Longrightarrow
(I+\gamma u\gamma^{\prime})^{-1}=\left(I+\left(\varpi+v\right)^{-1}\right)^{-1},~~\text{with}\quad\varpi:=\left(\gamma\gamma^{\prime}\right)^{-1}.
\end{equation}
To verify this claim, we first use the  matrix inversion lemma to prove that 
$$
u=I-\gamma^{\prime}~(v^{-1}+\gamma\gamma^{\prime})^{-1}\gamma.
$$
This implies that
$$
\gamma u \gamma^{\prime}=\gamma\gamma^{\prime}-\gamma\gamma^{\prime}~(v^{-1}+\gamma\gamma^{\prime})^{-1}\gamma
\gamma^{\prime}=\left(\left(\gamma\gamma^{\prime}\right)^{-1}+v\right)^{-1}=
\left(\varpi+v\right)^{-1}
$$
and completes the proof of (\ref{proof-ricc}). Formulae (\ref{k-ricc}) are now a direct consequence of the above decompositions. \cqfd

\subsection*{Dual Riccati fixed point matrices}\label{stat-Ricc-sec}

By (\ref{k-ricc-0}), the  fixed point  matrices $(r_{\theta},\overline{r}_{\theta_1})$ defined in (\ref{def-r-t}) and (\ref{def-over-r}) are connected with the formulae (\ref{connect-fp}).
We can check directly this assertion if we let
$$
\widehat{r}_{\theta}^{\,-1}:=
I+  \gamma_{\theta}^{\prime}~r_{\theta}~ \gamma_{\theta}
$$
and then apply the matrix inversion lemma to arrive at
\begin{eqnarray*}
\gamma_{\theta}\widehat{r}_{\theta}\gamma_{\theta}^{\prime}&=&\gamma_{\theta}\gamma_{\theta}^{\prime}-\gamma_{\theta} \gamma_{\theta}^{\prime} (r_{\theta}^{-1}+\gamma_{\theta}\gamma_{\theta}^{\prime})^{-1}\gamma_{\theta}\gamma_{\theta}^{\prime}\\
\\
&=&\left((\gamma_{\theta}\gamma_{\theta}^{\prime})^{-1}+r_{\theta}\right)^{-1}=\left(\varpi_{\theta}+r_{\theta}\right)^{-1}=r_{\theta}^{-1}-I.
\end{eqnarray*}
This implies the equivalences
$$
(I+\gamma_{\theta}~\widehat{r}_{\theta}~\gamma_{\theta}^{\prime})^{-1}=r_{\theta}\Longleftrightarrow r_{\theta}^{-1}=\gamma_{\theta}~\widehat{r}_{\theta}~\gamma_{\theta}^{\prime}+I
\Longleftrightarrow
\widehat{r}_{\theta}+(\gamma_{\theta}^{\prime}\gamma_{\theta})^{-1}=(
\gamma_{\theta}^{\prime}r_{\theta}\gamma_{\theta})^{-1}
$$
from where we see that
$$
I+\left(\widehat{r}_{\theta}+(\gamma_{\theta}^{\prime}\gamma_{\theta})^{-1}\right)^{-1}=I+\left(\widehat{r}_{\theta}+\overline{\varpi}_{\theta_1}\right)^{-1}=I+
\gamma_{\theta}^{\prime}r_{\theta}\gamma_{\theta}=\widehat{r}_{\theta}^{\,-1}
$$
By uniqueness of the positive fixed point we conclude that
$$
\widehat{r}_{\theta}=\overline{r}_{\theta_1}\Longrightarrow \overline{r}_{\theta_1}^{-1}=
I+  \gamma_{\theta}^{\prime}~r_{\theta}~ \gamma_{\theta}.
$$
This completes the proof of (\ref{connect-fp}). 

The matrix inversion lemma also yields the formulae
\begin{eqnarray*}
r_{\theta}&=&I- \gamma_{\theta}~\left(\overline{r}_{\theta_1}^{-1}+ \overline{\varpi}_{\theta_1}^{-1}\right)^{-1}~ \gamma_{\theta}^{\prime} \quad\text{and}\\
\overline{r}_{\theta_1}&=&I- \gamma_{\theta}^{\prime}\left(r_{\theta}^{-1}+ \varpi_{\theta}^{-1}\right)^{-1} \gamma_{\theta},
\end{eqnarray*}
which imply that
\begin{eqnarray*}
 \gamma_{\theta}^{\prime}~r_{\theta}&=& \gamma_{\theta}^{\prime}- \overline{\varpi}_{\theta_1}^{-1}~\left(
\overline{r}_{\theta_1}^{-1}+ \overline{\varpi}_{\theta_1}^{-1}\right)^{-1}~ \gamma_{\theta}^{\prime}\\
&=& \gamma_{\theta}^{\prime}-~\left(
\overline{r}_{\theta_1}+ \overline{\varpi}_{\theta_1}\right)^{-1}~\overline{r}_{\theta_1}~ \gamma_{\theta}^{\prime}\\
&=& \gamma_{\theta}^{\prime}-~\left(\overline{r}_{\theta_1}^{-1}-I\right)~\overline{r}_{\theta_1}~ \gamma_{\theta}^{\prime}=\overline{r}_{\theta_1}~ \gamma_{\theta}^{\prime}.
\end{eqnarray*}
This yields the commutation property
\begin{equation}\label{commut}
 \gamma_{\theta}^{\prime}~r_{\theta}=\overline{r}_{\theta_1}~ \gamma_{\theta}^{\prime}\Longleftrightarrow
r_{\theta}~ \gamma_{\theta}= \gamma_{\theta}~\overline{r}_{\theta_1}.
\end{equation}

\section{Relative entropy}\label{appendix-entro}

\subsection*{Sinkhorn conditioning formulae}\label{appendix-entro-p}

With some abuse of notation, consider the conditional decompositions
\begin{eqnarray*}
\Pa_{1,2}(d(x_1,x_2))&:=&\Pa_2(dx_2)~\Pa_{1|2}(x_2,dx_1)=\Pa_1(dx_1)~\Pa_{2|1}(x_1,dx_2)\\
\overline{\Pa}_{1,2}(d(x_1,x_2))&:=&
\overline{\Pa}_2(dx_2)~\overline{\Pa}_{1|2}(x_2,dx_1)=\overline{\Pa}_1(dx_1)~\overline{P}_{2|1}(x_1,dx_2).
\end{eqnarray*}
Observe that
$$
\mbox{\rm Ent}\left(\Pa_{1,2}~|~\overline{\Pa}_{1,2}\right)=
\mbox{\rm Ent}\left(\Pa_2~|~\overline{\Pa}_2\right)+\int~
\Pa_2(dx_2)~\mbox{\rm Ent}\left(\Pa_{1|2}(x_2,\point)~|~\overline{\Pa}_{1|2}(x_2,\point)\right).
$$
Thus, given $\overline{\Pa}_{1,2}$ and a prescribed marginal $\Pa_2$, we obtain
$$
(\mathcal{P}_2\times \overline{\mathcal{P}}_{1|2})^{\flat}=
\argmin_{\Pa_{1,2}}\mbox{\rm Ent}\left(\Pa_{1,2}~|~\overline{\Pa}_{1,2}\right).
$$
In the same way,
$$
\mbox{\rm Ent}\left(\Pa_{1,2}~|~\overline{\Pa}_{1,2}\right)=
\mbox{\rm Ent}\left(\Pa_1~|~\overline{\Pa}_1\right)+\int~
\Pa_1(dx_1)~\mbox{\rm Ent}\left(\Pa_{2|1}(x_1,\point)~|~\overline{\Pa}_{2|1}(x_1,\point)\right)
$$
and, given $\Pa_{1,2}$ and a prescribed marginal $\overline{\Pa}_1$, we have
$$
\overline{\Pa}_1\times \Pa_{2|1} =
\argmin_{\overline{\Pa}_{1,2}}\mbox{\rm Ent}\left(\Pa_{1,2}~|~\overline{\Pa}_{1,2}\right).
$$

\subsection*{Some inequalities}

Taking together (\ref{lem-tech-2}) and $A= I-\sigma_2^{-1}\sigma_1$ we see that
\begin{equation*}
\Vert \sigma_1-\sigma_2\Vert_F
\leq \frac{1}{2 \Vert \sigma_2^{-1}\Vert_F} 
\quad \text{implies} \quad
\Vert I-\sigma_2^{-1}\sigma_1\Vert_F\leq \Vert \sigma_2^{-1}\Vert_F~\Vert \sigma_1-\sigma_2\Vert_F\leq \frac{1}{2}
\end{equation*}
which, in turn, yields
\begin{equation*}
\left\vert\log{\mbox{\rm det}\left(\sigma_2^{-1}\sigma_1\right)}\right\vert 
\leq \frac{3}{2}~\Vert \sigma_2^{-1}\Vert_F~\Vert \sigma_1-\sigma_2\Vert_F.
\end{equation*}
Thus, we have the following lemma.
\begin{lem}\label{lem-detlog}
If
$$
\Vert \sigma_1-\sigma_2\Vert_F
\leq \frac{1}{2 \Vert \sigma_2^{-1}\Vert_F}
$$ 
we have the estimate
$$
\left\vert\log{\mbox{\rm det}\left(\sigma_1^{-1}\sigma_2\right)}\right\vert 
\leq \frac{3}{2}~\Vert \sigma_2^{-1}\Vert_F~\Vert \sigma_1-\sigma_2\Vert_F.
$$
\end{lem}

On the other hand,  we have
  \begin{eqnarray*}
\left\vert\tr\left(\sigma_2^{-1}\sigma_1-I\right)\right\vert&\leq & \left\Vert \sigma_2^{-1}\right\Vert_F~\left\Vert \sigma_2-\sigma_1\right\Vert_F,
\end{eqnarray*}
which yields the implication
$$
 \displaystyle\Vert \sigma_1-\sigma_2\Vert_F
\leq \frac{1}{2 \Vert \sigma_2^{-1}\Vert_F}\Longrightarrow
 D(\sigma_1~|~\sigma_2)\leq \frac{5}{2}
  \left\Vert \sigma_2^{-1}\right\Vert_F~\left\Vert \sigma_2-\sigma_1\right\Vert_F.
$$
We summarize the above discussion with the following proposition
\begin{prop}
Assume that
$$
\Vert \sigma_1-\sigma_2\Vert_F
\leq \frac{1}{2 \Vert \sigma_2^{-1}\Vert_F}.
$$
Then, we have
$$
\mbox{\rm Ent}\left(\nu_{m_1,\sigma_1}~|~\nu_{m_2,\sigma_2}\right)\leq \frac{5}{4}
\displaystyle  \left\Vert \sigma_2^{-1}\right\Vert_F~\left(\left\Vert \sigma_2-\sigma_1\right\Vert_F+\Vert m_1-m_2\Vert^2\right).
$$
 \end{prop}

\subsection*{Proof of (\ref{BL-intro-def})}\label{BL-intro-def-proof}

Take some parameters 
$$
\theta=(\alpha,\beta,\tau)\in \Theta_{d,\overline{d}}\quad\mbox{\rm and}\quad
\theta_1=(\iota,\kappa,\varsigma)\in \Theta_{d,\overline{d}}.
$$
Applying (\ref{KL-def}) with
$$
\begin{array}{llllll}
m_1 &= &(\iota+\kappa m)+\kappa(x-m), &\sigma_1 &= &\varsigma,
\\
m_2 &= &(\alpha+\beta m)+\beta(x-m), &\sigma_2 &= &\tau,
\end{array}$$
we find that
\begin{eqnarray*}
2\mbox{\rm Ent}\left(\delta_xK_{\theta_1}~|~\delta_xK_{\theta}\right) &=&
\tr\left(\tau^{-1} \varsigma - I\right)-\log{\mbox{det}\left( \tau^{-1} \varsigma\right)} \nonumber\\
&&+\Vert~\tau^{-1/2}\left[ ((\iota+\kappa m)-(\alpha+\beta m))+(\kappa-\beta)~(x-m)\right]\Vert^2_F,
\end{eqnarray*}
which implies that
\begin{eqnarray*}
2\text{Ent}\left(P_{\theta_1}~|~P_{\theta}\right) &=& \displaystyle 2~\int \nu_{m,\sigma}(dx)~ \text{Ent}\left(\delta_xK_{\theta_1}~|~\delta_xK_{\theta}\right) \nonumber \\
&=&\tr\left(\tau^{-1} \varsigma -I\right)-\log{\mbox{det}\left( \tau^{-1} \varsigma\right)} \nonumber\\
&&+\Vert\tau^{-1/2} ((\iota+\kappa m)-(\alpha+\beta m))\Vert_F^2+\Vert~\tau^{-1/2}(\kappa-\beta)~\sigma^{1/2}\Vert_F^2.\nonumber
\end{eqnarray*}
This ends the proof of (\ref{BL-intro-def}).\cqfd

 \subsection*{Proof of (\ref{ent-intro-even})}\label{ent-intro-even-proof}
 
 Let us first observe that
 $$
 \Pa_{2n}=P_{\theta_{2n}}\quad \text{with}\quad \theta_{2n}=
 \left( \alpha_{2n},\beta_{2n},\tau_{2n}\right)\quad \text{and}\quad
 \alpha_{2n}+\beta_{2n}m=m_{2n}.
 $$
 On the other hand, we have
 $$
 \SS(\theta)=\left( \iota_{\theta},\kappa_{\theta},\varsigma_{\theta}\right)
 \quad \text{with}\quad
 \iota_{\theta}+\kappa_{\theta}~m
=\overline{m} $$
We verify (\ref{ent-intro-even}) by replacing in (\ref{BL-intro-def}) the parameters $(\theta_1,\theta)$ by $(\theta_{2n},\Sa(\theta))$ and recalling (see for instance (\ref{ref-var-2n})) that 
$$
\alpha_{2n}+\beta_{2n}m=m_{2n}.
$$
\cqfd

\subsection*{Proof of Theorem~\ref{theo-entr-wass}}\label{theo-entr-wass-proof}

Denote by $(e^{-U},e^{-V})$ the densities of $(\eta,\mu)=(\nu_{m,\sigma},\nu_{\overline{m},\overline{\sigma}})$. In this notation, we have
$$
\mu(V)=\nu_{\overline{m},\overline{\sigma}}\left(V\right)=\frac{d}{2}+\frac{1}{2}\log{(\mbox{\rm det}(2\pi\overline{\sigma}))}.
$$
On the other hand, using (\ref{BL-intro-def}) for any $\theta=(\alpha,\beta,\tau)$ we obtain
$$
2\mbox{\rm Ent}\left(P_{\SS(\theta)}~|~P_{\theta}\right)\\
=D(\varsigma_{\theta}~|~\tau)+\Vert\tau^{-1/2} (\overline{m}-(\alpha+\beta m))\Vert_F^2+\Vert\tau^{-1/2}(\kappa_{\theta}-\beta)~\sigma^{1/2}\Vert^2_F 
$$
with the Burg divergence $D$ defined in (\ref{burg-def}) and the Schr\"odinger bridge map $\SS$ defined in Theorem~\ref{Th1}. Choosing $\theta=\theta(t):=(\alpha,\beta,tI)$ we have
$$
\nu_{m,\sigma}K_{\theta}=\nu_{m_0,\sigma_0}\quad \mbox{\rm with}\quad
m_0:=\alpha+\beta m\quad \mbox{\rm and}\quad\sigma_0=\sigma_{\beta}+tI
$$
with the rescaled covariance matrix $\sigma_{\beta}$ defined in (\ref{rmk-theta-t-1}).
Also recall from (\ref{ref-fix-point-intro}) that
$$
 \varsigma_{\theta(t)}+\frac{\varsigma_{\theta(t)}}{t}~\sigma_{\beta}~\frac{\varsigma_{\theta(t)}}{t}=\overline{\sigma},
$$
which implies that
\begin{eqnarray*} 
t\Vert t^{-1/2} (\overline{m}-(\alpha+\beta m))\Vert_F^2+t\Vert t^{-1/2}(\kappa_{\theta}-\beta)~\sigma^{1/2}\Vert^2_F 
&=& \Vert\overline{m}-m_0\Vert_F^2 \\
&& + \tr\left(\left(\frac{\varsigma_{\theta(t)}}{t}-I\right) \sigma_{\beta} \left(\frac{\varsigma_{\theta(t)}}{t}-I\right)\right)\\
&=& \Vert\overline{m}-m_0\Vert_F^2+\tr\left(\overline{\sigma}\right)+\tr(\sigma_{\beta})\\
&&-2\tr\left(\frac{\varsigma_{\theta(t)}}{t}\sigma_{\beta}\right)-\tr\left(\varsigma_{\theta(t)}\right).
\end{eqnarray*}
If we now recall that
\begin{eqnarray*}
t~ D(\varsigma_{\theta(t)}~|~tI)&=&\tr\left(\varsigma_{\theta(t)}-tI\right)-t~\log{\mbox{det}\left(\frac{\varsigma_{\theta(t)}}{t}\right)}\\
&=&\tr\left(\varsigma_{\theta(t)}-tI\right)-t~\log{\mbox{det}\left(\frac{r_{\theta(t)}}{t}
\right)}-t~\log{\mbox{det}\left(\overline{\sigma}
\right)}
\end{eqnarray*}
then we readily find that
$$
 \begin{array}{l}
  \displaystyle  2 t\left(\mbox{\rm Ent}\left(P_{\SS(\theta(t))}~|~P_{\theta(t)}\right)+\mu(V)\right)
\\
\\
=  \displaystyle  \Vert\overline{m}-m_0\Vert_F^2+\tr\left(\overline{\sigma}\right)
+\tr(\sigma_{\beta})
-2
\tr\left(\frac{\varsigma_{\theta(t)}}{t}~\sigma_{\beta}\right)
+t\left(d\log{(2\pi)}-\log{\mbox{det}\left(\frac{r_{\theta(t)}}{t}
\right)}\right).
\end{array}
$$
If we also observe that
$$
\tr\left((\overline{\sigma}~\sharp~\sigma_{\beta}^{-1})~\sigma_{\beta}\right)=\tr\left(\left(\sigma_{\beta}^{1/2}~\overline{\sigma}~\sigma_{\beta}^{1/2}\right)^{1/2}\right)
$$
then we arrive at the decomposition (\ref{decomp-ww}).

On the other hand, 
we have
$$
\log{\mbox{det}\left(\frac{r_{\theta(t)}}{t}
\right)}
=\log{\mbox{det}\left(I-\left(I-\frac{r_{\theta(t)}}{t}\omega^{-1/2}\right)
\right)}+\log{\mbox{det}\left(\omega^{1/2}
\right)}
$$
and by (\ref{monge-maps}) we obtain the estimates
$$
\Vert {\varsigma_{\theta(t)}}/{t}-(\sigma_{\beta}^{-1}~\sharp~ \overline{\sigma})\Vert\vee    \Vert t^{-1}r_{\theta(t)}~\omega^{-1/2}-I\Vert\leq c~t.
$$
Therefore, by (\ref{lem-tech-2}) there exists some constant $c<\infty$ and some $t_0>0$
sufficiently small such that for any $0<t\leq t_0$ we have
$$
\left\vert\log{\mbox{\rm det}\left(\frac{r_{\theta(t)}}{t}\right)}\right\vert 
\leq  c~t.
$$
\cqfd

\section{Gaussian Sinkhorn algorithm}\label{gauss-sinhorn-details}

\subsection*{Conjugate formulae}
Assume that $\Pa_{2n}=P_{\theta_{2n}}$ (equivalently, $\Ka_{2n}=K_{\theta_{2n}}$). In this case, we have $\pi_{2n}=\eta K_{\theta_{2n}}=\nu_{m_{2n},\sigma_{2n}}$ with the parameters
\begin{eqnarray}
(m_{2n},\sigma_{2n})&=&(\alpha_{2n}+\beta_{2n}m,\beta_{2n}~\sigma~\beta_{2n}^{\prime}+\tau_{2n})\nonumber\\
&=&(a_m(\theta_{2n}),b_{\sigma}(\theta_{2n}))\nonumber\\
&=&h_{m,\sigma}(\theta_{2n}).
\label{ref-var-2n}
\end{eqnarray}
The conjugate formula (\ref{ref-conjug}) yields
$$
(\nu_{h_{m,\sigma}(\theta_{2n})}\times K_{\BB_{m,\sigma}(\theta_{2n})})^{\flat}=\nu_{m,\sigma}\times K_{\theta_{2n}}
$$
which implies that
$$
\Ka_{2n+1}=K_{\theta_{2n+1}},\quad \text{with}\quad
\theta_{2n+1}=(\alpha_{2n+1},\beta_{2n+1},\tau_{2n+1}):=\BB_{m,\sigma}(\theta_{2n})
$$
or, equivalently,
 \begin{equation}\label{beta-s-2-i}
\left\{ \begin{array}{rcl}
 \alpha_{2n+1}&=&m-\beta_{2n+1}m_{2n}=m-\beta_{2n+1}a_m(\theta_{2n})\\
 &&\\
\beta_{2n+1}&=&\sigma~\beta_{2n}^{\prime}\sigma_{2n}^{-1}=\sigma~\beta_{2n}^{\prime}~b_{\sigma}(\theta_{2n})^{-1}\quad \mbox{\rm and}\quad
\tau_{2n+1}^{-1}=\sigma^{-1}+ \beta_{2n}^{\prime}~\tau_{2n}^{-1}~\beta_{2n}
\end{array}\right..
\end{equation}

In terms of the random map
(\ref{ref-conjug-Z}) we have $$K_{\theta_{2n+1}}(y,dx)=\PP(Z_{\theta_{2n+1}}(y)\in dx),$$ with
 \begin{equation}\label{Xa-1}
Z_{\theta_{2n+1}}(y)=m+\beta_{2n+1}(y-m_{2n})+\tau_{2n+1}^{1/2}~G.
\end{equation}
This implies that
$$
 \Pa_{2n+1}=(\nu_{\overline{m},\overline{\sigma}}\times K_{\theta_{2n+1}})^{\flat}=\overline{P}^{\,\flat}_{\theta_{2n+1}}\quad \mbox{\rm and}\quad
 \pi_{2n+1}:=\nu_{\overline{m},\overline{\sigma}} K_{\theta_{2n+1}}=\nu_{m_{2n+1},\sigma_{2n+1}}
$$
with the parameters
\begin{eqnarray}
(m_{2n+1},\sigma_{2n+1})&=&(\alpha_{2n+1}+\beta_{2n+1}\overline{m},\beta_{2n+1}~\overline{\sigma}~\beta_{2n+1}^{\prime}+\tau_{2n+1})\nonumber\\
&=&\left(a_{\overline{m}}(\theta_{2n+1}),b_{\overline{\sigma}}(\theta_{2n+1})\right)\nonumber\\
&=&h_{\overline{m},\overline{\sigma}}(\theta_{2n+1}).
\label{ref-var-2n1}
\end{eqnarray}

The conjugate formula (\ref{ref-conjug}) yields
$$
\nu_{h_{\overline{m},\overline{\sigma}}(\theta_{2n+1})
}\times K_{\BB_{\overline{m},\overline{\sigma}}(\theta_{2n+1})}=(\nu_{\overline{m},\overline{\sigma}}\times K_{\theta_{2n+1}})^{\flat}\\
$$
ans, as a consequence,
$$
\Ka_{2(n+1)}=K_{\theta_{2(n+1)}},\quad \text{with}\quad
\theta_{2(n+1)}:=
\BB_{\overline{m},\overline{\sigma}}(\theta_{2n+1}).
$$
Equivalently, we have
 \begin{equation}
\left\{ \begin{array}{rcl}
\alpha_{2(n+1)}&=&\overline{m}-\beta_{2(n+1)}~m_{2n+1}=\overline{m}-\beta_{2(n+1)}~a_{\overline{m}}(\theta_{2n+1})\\
&&\\
\beta_{2(n+1)}&=&\overline{\sigma}~\beta_{2n+1}^{\prime}\sigma_{2n+1}^{-1}=
\overline{\sigma}~\beta_{2n+1}^{\prime}~b_{\overline{\sigma}}(\theta_{2n+1})^{-1}\\
&&\\ \tau_{2(n+1)}^{-1}&:=&\overline{\sigma}^{-1}+ \beta_{2n+1}^{\prime}~\tau_{2n+1}^{-1}~
\beta_{2n+1}\end{array}\right.,\label{beta-s-2-ii}
\end{equation} 
and, in terms of the random map (\ref{ref-conjug-Z}), 
$$
K_{\theta_{2(n+1)}}(x,dy)=\PP(Z_{\theta_{2(n+1)}}(x)\in dy)
$$ with
 \begin{equation}\label{Ya-1}
Z_{\theta_{2(n+1)}}(x)=\overline{m}+\beta_{2(n+1)}~(x-m_{2n+1})+\tau_{2(n+1)}^{1/2}~G.
\end{equation}

\subsection*{Proof of Lemma~\ref{lem-pivot}}\label{lem-pivot-proof}

Applying the matrix inversion lemma to (\ref{ref-var-2n}) we find that
\begin{equation}\label{hom-mi}
\sigma_{2n}^{-1}=\left(\beta_{2n}\sigma\beta_{2n}^{\prime}+\tau_{2n}\right)^{-1}=\tau_{2n}^{-1}-\tau^{-1}_{2n}\beta_{2n}~\tau_{2n+1}~\beta_{2n}^{\prime}\tau_{2n}^{-1}
\end{equation}
and, on the other hand, by (\ref{beta-s-2-i}) we have
$$
\tau_{2n+1}^{-1}-\sigma^{-1}=\beta_{2n}^{\prime}\tau_{2n}^{-1}\beta_{2n},
$$
which together imply the equalities
\begin{eqnarray*}
\beta_{2n+1}=\sigma~\beta_{2n}^{\prime}\sigma_{2n}^{-1}
&=&\sigma\beta_{2n}^{\prime}\tau_{2n}^{-1}-\sigma~\left(\beta_{2n}^{\prime}\tau^{-1}_{2n}\beta_{2n}\right)~\tau_{2n+1}~\beta_{2n}^{\prime}\tau_{2n}^{-1}\\
&=&\sigma\beta_{2n}^{\prime}\tau_{2n}^{-1}-\sigma~\left(\tau_{2n+1}^{-1}-\sigma^{-1}\right)~\tau_{2n+1}~\beta_{2n}^{\prime}\tau_{2n}^{-1}=\tau_{2n+1}~\beta_{2n}^{\prime}\tau_{2n}^{-1}.
\end{eqnarray*}
This yields the commutation formula
\begin{equation}\label{k-1}
\tau_{2n+1}^{-1}~\beta_{2n+1}=\beta_{2n}^{\prime}~\tau_{2n}^{-1}.
\end{equation}

In the same vein, using (\ref{ref-var-2n1}) we see that
\begin{equation}\label{hom-mii}
\sigma_{2n+1}^{-1}=\left(\beta_{2n+1}\overline{\sigma}\beta_{2n+1}^{\prime}+\tau_{2n+1}\right)^{-1}=
\tau_{2n+1}^{-1}-\tau_{2n+1}^{-1}\beta_{2n+1} \tau_{2(n+1)}      \beta_{2n+1}^{\prime}\tau_{2n+1}^{-1}
\end{equation}
and, by (\ref{beta-s-2-ii}), 
$$
\tau_{2(n+1)}^{-1}-\overline{\sigma}^{-1}= \beta_{2n+1}^{\prime}\tau_{2n+1}^{-1}
\beta_{2n+1}.
$$
The equations above lead to
\begin{eqnarray*}
\beta_{2(n+1)}=\overline{\sigma}~\beta_{2n+1}^{\prime}\sigma_{2n+1}^{-1}&=&
\overline{\sigma}~\beta_{2n+1}^{\prime}\tau_{2n+1}^{-1}-\overline{\sigma}~\left(\beta_{2n+1}^{\prime}\tau_{2n+1}^{-1}\beta_{2n+1} \right)\tau_{2(n+1)}      \beta_{2n+1}^{\prime}\tau_{2n+1}^{-1}\\
&=&\overline{\sigma}~\beta_{2n+1}^{\prime}\tau_{2n+1}^{-1}-\overline{\sigma}~\left(\tau_{2(n+1)}^{-1}-\overline{\sigma}^{-1} \right)\tau_{2(n+1)}      \beta_{2n+1}^{\prime}\tau_{2n+1}^{-1}\\
&=&\tau_{2(n+1)}      \beta_{2n+1}^{\prime}\tau_{2n+1}^{-1},
\end{eqnarray*}
which yields the commutation formula
\begin{equation}\label{kk-2}
\tau_{2(n+1)}^{-1}~\beta_{2(n+1)}= \beta_{2n+1}^{\prime}\tau_{2n+1}^{-1}=\tau_{2n}^{-1}~\beta_{2n}.
\end{equation}
We complete the proof of (\ref{hom}) by choosing $n=0$ in the r.h.s. of (\ref{kk-2}).
\cqfd

 \subsection*{Sinkhorn Gibbs-loop process}\label{gibbs-sec}

The Gibbs transitions discussed in (\ref{fixed-points-gibbs}) can also be rewritten as
$$
 \Ka^{\circ}_{2n+1}(x_1,dx_2)=\PP\left(\Za^{\circ}_{2n+1}(x_1)\in dx_2\right)
$$
with the random maps
\begin{equation}\label{map-gibbs-odd}
\Za^{\circ}_{2n+1}(x)=
m+\beta_{2n+1}^{\circ}(x-m)+\left(\tau_{2n+1}^{\circ}\right)^{1/2}~G
\end{equation}
defined in terms of the parameter $\beta_{2n+1}^{\circ}$ in (\ref{ref-m-o-intro}) and 
$$
\tau_{2n+1}^{\circ}:=\tau_{2n+1}+\beta_{2n+1}~\tau_{2n}~
\beta_{2n+1}^{\prime}.
$$
The above assertion is a direct consequence of the linear-Gaussian structure of the random maps discussed in (\ref{Xa-1}) and  (\ref{Ya-1}). 

In the same vein, for any $n\geq 1$ we have
$$
\Ka^{\circ}_{2n}(y_1,dy_2):=\PP\left(\Za^{\circ}_{2n}(y_1)\in dy_2\right)
$$
with the random maps
\begin{eqnarray}
\Za^{\circ}_{2n}(y)&=&\overline{m}+\beta^{\circ}_{2n}(y-\overline{m})+\left(\tau^{\circ}_{2n}\right)^{1/2}~G\label{Z-o-even}
\end{eqnarray}
defined in terms of the matrix $\beta^{\circ}_{2n}$ in (\ref{ref-m-intro}) and 
$$
\tau^{\circ}_{2n}:=\tau_{2n}+
\beta_{2n}~\tau_{2n-1}~\beta_{2n}^{\prime}.
$$

 \subsection*{Proof of (\ref{directed-prod-sig})}\label{directed-prod-sig-proof}
Iterating the random maps (\ref{Z-o-even}) {we readily find that
$$
\pi_{2n}=\mbox{\rm Law}(\Xa^{\circ}_{2n})=\pi_{2(n-1)}\Ka^{\circ}_{2n}
$$ with the random variables}
\begin{eqnarray*}
\Xa^{\circ}_{2n}-\overline{m}
&=&\beta^{\circ}_{2n,0}~(\Xa^{\circ}_{0}-\overline{m})+\left(\tau^{\circ}_{2n,0}\right)^{1/2}~G.
\end{eqnarray*}
{In the above display, $\beta^{\circ}_{2n,0}$ stands for the directed matrix product defined in (\ref{directed-prod}). In addition, using}  the fixed point equations
(\ref{fixed-points-gibbs}) we also see that
$$
\left(\beta^{\circ}_{2n,0}\right)\overline{\sigma}
\left(\beta^{\circ}_{2n,0}\right)^{\prime}+\tau^{\circ}_{2n,0}=\overline{\sigma}.
$$
The l.h.s. assertion in (\ref{directed-prod-sig}) is a direct consequence of the above formula.
In the same vein, using (\ref{gibbs-tv}), for any $n\geq 1$ we arrive at
$$
\pi_{2n+1}=\mbox{\rm Law}(\Xa^{\circ}_{2n+1})=\pi_{2n-1}\Ka^{\circ}_{2n+1}
$$
with the initial condition
$$
\pi_{1}=\mbox{\rm Law}(\Xa^{\circ}_{1})=\nu_{m_1,\sigma_1}.
$$
Iterating the map (\ref{map-gibbs-odd}) we obtain
$$
\Xa^{\circ}_{2n+1}-m
=\beta^{\circ}_{2n+1,1}~(\Xa^{\circ}_{1}-m)+\left(\tau^{\circ}_{2n+1,1}\right)^{1/2}~G
$$
with the directed matrix product $\beta^{\circ}_{2n+1,1}$ defined in (\ref{directed-prod}).
Using  the fixed point equations
(\ref{fixed-points-gibbs}) we also see that
$$
\beta^{\circ}_{2n+1,1}:=\beta^{\circ}_{2n+1}\beta^{\circ}_{2n-1}\ldots \beta^{\circ}_{3}\quad
\mbox{\rm and}\quad
\left(\beta^{\circ}_{2n+1,1}\right)\sigma
\left(\beta^{\circ}_{2n+1,1}\right)^{\prime}+\tau^{\circ}_{2n+1,1}=\sigma.
$$
The r.h.s. assertion in (\ref{directed-prod-sig}) is a direct consequence of the above formula. \cqfd

 \subsection*{Proof of Lemma~\ref{prop-beta-o}}\label{prop-beta-o-proof}

Using (\ref{ref-var-2n1}) and (\ref{hom}) we see that
\begin{eqnarray*}
\beta_{2n-1}^{\prime}\sigma_{2n-1}^{-1}\beta_{2n-1} &=&
\beta_{2n-1}^{\prime}\left(\beta_{2n-1}~\overline{\sigma}~\beta_{2n-1}^{\prime}+\tau_{2n-1}
\right)^{-1}\beta_{2n-1}\\
&=&\tau^{-1}\beta~\tau_{2n-1}~\left(\tau_{2n-1}~\beta^{\prime}\tau^{-1}~\overline{\sigma}~\tau^{-1}\beta~\tau_{2n-1}+\tau_{2n-1}
\right)^{-1}\tau_{2n-1}~\beta^{\prime}\tau^{-1}
\end{eqnarray*}
and, using the l.h.s. description of $\beta_{2n}$ given in (\ref{beta-s-2-ii}), this implies that
$$
\overline{\sigma}^{-1/2}~\beta^{\circ}_{2n}~\overline{\sigma}^{1/2}=
\overline{\sigma}^{1/2}\beta_{2n-1}^{\prime}\sigma_{2n-1}^{-1}\beta_{2n-1}\overline{\sigma}^{1/2}
= \gamma_{\theta}~\left( \gamma_{\theta}^{\prime} \gamma_{\theta}+\upsilon_{2n-1}^{-1}
\right)^{-1}~ \gamma_{\theta}^{\prime}
$$
with the matrices $ \gamma_{\theta}$ and $\upsilon_{2n-1}$ defined in (\ref{def-w-1}) and (\ref{resc-M}). 
{On the other hand, combining  (\ref{k-ricc-0}) with the matrix inversion lemma we also have
\begin{eqnarray*}
\upsilon_{2n}&=&\left(I+ \gamma_{\theta}~\upsilon_{2n-1}~  \gamma_{\theta}^{\prime}\right)^{-1}=I-
\gamma_{\theta}\left(\upsilon_{2n-1}^{-1}+\gamma_{\theta}^{\prime}\gamma_{\theta}\right)^{-1}\gamma_{\theta}^{\prime}.
\end{eqnarray*}
Recalling that $\overline{\varpi}_{\theta_1}^{-1}=\gamma_{\theta}^{\prime}\gamma_{\theta}$ and $\overline{\gamma}_{\theta_1}=\gamma_{\theta}^{\prime}$, this ends the proof of (\ref{beta-2-tau-1}).
}

In the same vein, using (\ref{ref-var-2n}) and (\ref{hom}) we have
\begin{eqnarray*}
\beta_{2n}^{\prime}~\sigma_{2n}^{-1}~\beta_{2n}
&=&
\beta_{2n}^{\prime}\left(\beta_{2n}~\sigma~\beta_{2n}^{\prime}+\tau_{2n}\right)^{-1}\beta_{2n}\\
\\
&=&\beta^{\prime}~\tau^{-1}~\tau_{2n}~\left(\tau_{2n}~\tau^{-1}\beta~\sigma~\beta^{\prime}~\tau^{-1}~\tau_{2n}~+\tau_{2n}\right)^{-1}\tau_{2n}~\tau^{-1}\beta
\end{eqnarray*}
and, using the l.h.s. description of $\beta_{2n+1}$ given in (\ref{beta-s-2-i}), this yields
$$
\sigma^{-1/2}\beta_{2n+1}^{\circ}\sigma^{1/2}=
\sigma^{1/2}~\beta_{2n}^{\prime}\sigma_{2n}^{-1}\beta_{2n}~\sigma^{1/2}
= \gamma_{\theta}^{\prime}~\left( \gamma_{\theta} \gamma_{\theta}^{\prime}+\upsilon_{2n}^{-1}\right)^{-1}~ \gamma_{\theta}
$$
with the matrices $ \gamma_{\theta}$ and $\upsilon_{2n}$ defined in (\ref{def-w-1}) and (\ref{resc-M}).  
{Combining  (\ref{k-ricc-0}) with the matrix inversion lemma we also have
\begin{eqnarray*}
\upsilon_{2n+1}&=&\left(
I+  \gamma_{\theta}^{\prime}~\upsilon_{2n}~ \gamma_{\theta}\right)^{-1}=I-
\gamma_{\theta}^{\prime}\left(\upsilon_{2n}^{-1}+\gamma_{\theta}\gamma_{\theta}^{\prime}\right)^{-1}\gamma_{\theta}.
\end{eqnarray*}} This ends the proof of (\ref{beta-2-tau-2}). \cqfd

 \subsection*{{Proof of (\ref{def-rho-1})}}\label{def-rho-12-proof}

Using (\ref{beta-2-tau-1}) and (\ref{beta-2-tau-2})  we check that
\begin{eqnarray*}
\overline{\sigma}^{-1/2}~\beta^{\circ}_{2n,0}~\overline{\sigma}^{1/2}&=&\left(I-\upsilon_{2n}\right)\left(I-\upsilon_{2(n-1)}\right)\ldots\left(I-\upsilon_{2}\right) \quad \text{and}\\
\sigma^{-1/2}~\beta_{2n+1,1}^{\circ}~\sigma^{1/2}&=&\left(I-\upsilon_{2n+1}\right)\left(I-\upsilon_{2n-1}\right)\ldots\left(I-\upsilon_{3}\right).
\end{eqnarray*}
The estimate (\ref{def-rho-1}) is now a direct consequence of the product formula (\ref{En-prod-uv}), Proposition~\ref{fix-p} and Proposition~\ref{prop-Ea-u}. \cqfd

\section{Schr\"odinger potentials}\label{sch-appendix}

\subsection*{Proof of Theorem~\ref{theo-s-pot}}\label{theo-s-pot-proof}

Using the decomposition
$$
y-(\alpha+\beta x)=(y-\overline{m})-\left((m_0-\overline{m})+\beta (x-m)\right)
$$
we readily check that
\begin{eqnarray*}
\Vert \tau^{-1/2}(y-(\alpha+\beta x))\Vert^2_F
&=&\Vert \tau^{-1/2}(y-\overline{m})\Vert^2_F+\Vert \tau^{-1/2}\left((m_0-\overline{m})+\beta (x-m)\right)\Vert^2_F\\
&&-2(y-\overline{m})^{\prime}\tau^{-1}\left((m_0-\overline{m})+\beta (x-m)\right).
\end{eqnarray*}
Recalling that $\varsigma_{\theta}^{-1} \kappa_{\theta}=\tau^{-1}\beta$ (with $\kappa_{\theta}$ defined in (\ref{def-Sa})), this implies that
$$
\begin{array}{l}
\displaystyle-\frac{1}{2}\Vert \tau^{-1/2}(y-(\alpha+\beta x))\Vert^2_F-\left(\VV_{\theta}(y)-\VV_{\theta}(\overline{m})\right)\\
\\
=
\displaystyle-\frac{1}{2}\Vert \tau^{-1/2}(y-(\alpha+\beta x))\Vert^2_F\\
\\
\hskip.4cm\displaystyle-\left((y-\overline{m})^{\prime}~\tau^{-1}~\left(m_{0}-\overline{m}\right)+\frac{1}{2}~\Vert \varsigma_{\theta}^{-1/2}(y-\overline{m})\Vert_F^2
-\frac{1}{2}~\Vert \tau^{-1/2}(y-\overline{m})\Vert_F^2
\right)\\
\\
\displaystyle=-\frac{1}{2}\Vert \tau^{-1/2}\left((m_0-\overline{m})+\beta (x-m)\right)\Vert^2_F\\
\\
\displaystyle\hskip0.4cm-\frac{1}{2}~\Vert \varsigma_{\theta}^{-1/2}(y-\overline{m})\Vert_F^2+(y-\overline{m})^{\prime}\varsigma_{\theta}^{-1}\left(\kappa_{\theta} (x-m)\right)
\end{array}
$$
and, rewriting in a slightly different form, we have proved that
\begin{eqnarray*}
-\frac{1}{2}\Vert \tau^{-1/2}(y-(\alpha+\beta x))\Vert^2_F-\left(\VV_{\theta}(y)-\VV_{\theta}(\overline{m})\right) &=&
-\frac{1}{2}\Vert \tau^{-1/2}\left((m_0-\overline{m})+\beta (x-m)\right)\Vert^2_F \\
&& +\frac{1}{2}\Vert \varsigma_{\theta}^{-1/2}\left(\kappa_{\theta} (x-m)\right)\Vert^2_F\\
&&-\frac{1}{2}~\Vert \varsigma_{\theta}^{-1/2}~\left((y-\overline{m})-\kappa_{\theta} (x-m)\right)\Vert^2_F.
\end{eqnarray*}
By (\ref{inter-m-sigma-1}) and (\ref{inter-beta-01}) we have
$$
\tau^{-1}_1(m_1-m)=
\tau^{-1}_1\beta_1(\overline{m}-m_0)=\beta^{\prime}\tau^{-1}(\overline{m}-m_0)
$$
and, on the other hand, using (\ref{inter-tau-1-i}) (\ref{def-Sa}) and (\ref{connect-fp-v2}) we readily check that
$$
\begin{array}{l}
\displaystyle
\overline{\varsigma}_{\theta_1}^{-1}=\sigma^{-1}+\kappa_{\theta}^{\prime}\varsigma_{\theta}^{-1}
\kappa_{\theta}\quad\mbox{\rm and}\quad
\tau_1^{-1}=\sigma^{-1}+\beta^{\prime}\tau^{-1}\beta\\
\end{array}
$$
which implies
$$
\kappa_{\theta}^{\prime}~\varsigma_{\theta}^{-1}~
\kappa_{\theta}-\beta^{\prime}~\tau^{-1}~\beta=\overline{\varsigma}_{\theta_1}^{-1}-\tau_1^{-1}.
$$
This yields
$$
\begin{array}{l}
\displaystyle-\frac{1}{2}\Vert \tau^{-1/2}\left((m_0-\overline{m})+\beta (x-m)\right)\Vert^2_F+\frac{1}{2}\Vert \varsigma_{\theta}^{-1/2}\left(\kappa_{\theta} (x-m)\right)\Vert^2_F\\
\\
=\displaystyle-\frac{1}{2}\Vert \tau^{-1/2}\left(m_0-\overline{m}\right)\Vert^2_F+
(x-m)^{\prime}\tau^{-1}_1(m_1-m)+\frac{1}{2} (x-m)^{\prime}\left(\overline{\varsigma}_{\theta_1}^{-1}-\tau_1^{-1}
\right) (x-m)\\
\\
=\displaystyle-\frac{1}{2}\Vert \tau^{-1/2}\left(m_0-\overline{m}\right)\Vert^2_F+\UU_{\theta}(x)-\left(\UU_{\theta}(m)+(U(x)-U(m)\right)
\end{array}
$$
from where we find that
$$
\begin{array}{l}
\displaystyle-\frac{1}{2}\Vert \tau^{-1/2}(y-(\alpha+\beta x))\Vert^2_F-\VV_{\theta}(y)-\UU_{\theta}(x)\\
\\
\displaystyle=-\frac{1}{2}~\Vert\tau^{-1/2}\left(\overline{m}-m_0\right)\Vert^2_F-(\VV_{\theta}(\overline{m})+(\UU_{\theta}(m)-U(m)))\\
\\
\hskip3cm\displaystyle-U(x)-\frac{1}{2}~\Vert\varsigma_{\theta}^{-1/2}~\left((y-\overline{m})-\kappa_{\theta} (x-m)\right)\Vert^2_F.
\end{array}
$$
We end the proof of (\ref{ref-uv-infty-intro-eq}) using the fact that
$$
\frac{\sqrt{\mbox{\rm det}(\tau)}}{\sqrt{\mbox{\rm det}(\varsigma_{\theta})}}~\exp{\left(\VV_{\theta}(\overline{m})+(\UU_{\theta}(m)-U(m))+\frac{1}{2}~\Vert \tau^{-1/2}\left(m_0-\overline{m}\right)\Vert^2_F\right)}=1.
$$
\cqfd

\subsection*{Proof of Proposition~\ref{prop-schp}}\label{prop-schp-proof}
Assume that at some rank $n\geq 0$ we have
$$
\Pa_{2n}(d(x,y))=e^{-U_{2n}(x)}~q(x,y)~e^{-V_{2n}(y)}~dxdy
$$
for some  potential functions $(U_{2n},V_{2n})$ such that
$$
U_{2n}=U+\log{Q(e^{-V_{2n}})}
\quad\text{and}\quad
\Ka_{2n}(x,dy)=Q_{V_{2n}}(x,dy):=\frac{Q(x,dy)e^{-V_{2n}(y)}}{Q(e^{-V_{2n}})(x)}.
$$
This condition is met at rank $n=0$ with $(U_0,V_0)=(U,0)$.
By (\ref{s-2}) we have
$$
\Ka_{2n+1}(y,dx)=R_{U_{2n+1}}(y,dx):=\frac{R(y,dx)~e^{-U_{2n+1}(x)}}{R(e^{-U_{2n+1}})(y)}
\quad\mbox{\rm with}\quad U_{2n+1}=U_{2n},
$$
which yields
$$
\Pa_{2n+1}(d(x,y))=e^{-U_{2n+1}(x)}~q(x,y)~e^{-V_{2n+1}(y)}~dxdy
$$
with the potential function
$$
V_{2n+1}=V+\log{R(e^{-U_{2n+1}})}.
$$
In this case, using (\ref{s-2}) we see that
$$
\Ka_{2(n+1)}=Q_{V_{2(n+1)}}
\quad\mbox{\rm with}\quad V_{2(n+1)}=V_{2n+1}
$$
and, as a consequence,
$$
\Pa_{2(n+1)}(d(x,y))=e^{-U_{2(n+1)}(x)}~q(x,y)~e^{-V_{2(n+1)}(y)}~dxdy
$$
with
$$
U_{2(n+1)}:=U+\log{Q(e^{-V_{2(n+1)}})}.
$$
This ends the proof of the proposition.\cqfd

\subsection*{Proof of Lemma~\ref{lem-key-potentials}}\label{lem-key-potentials-proof}

Following (\ref{resc-M}), consider the rescaled covariance matrices
\begin{eqnarray}
\xi_{2n+1}&:=&
\sigma^{-1/2}\sigma_{2n+1}\sigma^{-1/2}\quad\mbox{\rm and}\quad
\xi_{2n}:=\overline{\sigma}^{-1/2}\sigma_{2n}~\overline{\sigma}^{-1/2}.
\label{resc-M-sigma}
\end{eqnarray}
Combining (\ref{ref-var-2n}) with (\ref{hom}) we can write
\begin{eqnarray*}
\xi_{2n}&=&\overline{\sigma}^{-1/2}
\left(\tau_{2n}~\tau^{-1}\beta\right)
~\sigma~
\left(\beta^{\prime}~\tau^{-1}~\tau_{2n}\right)
\overline{\sigma}^{-1/2}+\upsilon_{2n}\\
&=&\overline{\sigma}^{-1/2}
\tau_{2n}~\overline{\sigma}^{-1/2}~\left(\overline{\sigma}^{1/2}~\tau^{-1}~\beta ~\sigma^{1/2}\right)
~
\left(\sigma^{1/2}~\beta^{\prime}~\tau^{-1}~\overline{\sigma}^{1/2}\right)~\overline{\sigma}^{-1/2}\tau_{2n}~
\overline{\sigma}^{-1/2}+\upsilon_{2n},
\end{eqnarray*}
which yields the formula
\begin{equation}\label{ref-xi-even}
\xi_{2n}=\upsilon_{2n}+\upsilon_{2n}~ \varpi_{\theta}^{-1}~\upsilon_{2n}
\end{equation}
with the matrix $\varpi_{\theta}$ defined in (\ref{def-w-1}). Using the matrix sum inversion formula
$$
(\upsilon+\upsilon~\varpi^{-1}~\upsilon)^{-1}=\upsilon^{-1}-(\varpi+\upsilon)^{-1}
$$
we readily see that
$$
\xi_{2n}^{-1}-I=\upsilon_{2n}^{-1}-\left(I+\left( \varpi_{\theta}+\upsilon_{2n}\right)^{-1}\right)
=\upsilon_{2n}^{-1}-\left(\mbox{\rm Ricc}_{\varpi_{\theta}}(\upsilon_{2n})\right)^{-1}.$$
This ends the proof of the l.h.s. assertion in (\ref{rec-xii}).

In the same vein, combining (\ref{ref-var-2n1}) with (\ref{hom}) leads to
\begin{equation}\label{ref-xi-odd}
\xi_{2n+1}=\upsilon_{2n+1}+\upsilon_{2n+1}~ \overline{\varpi}_{\theta_1}^{-1}~\upsilon_{2n+1}\quad\mbox{\rm and}\quad
\xi_{2n+1}^{-1}-I=\upsilon_{2n+1}^{-1}-\upsilon_{2n+3}^{-1}
\end{equation}
which, rewritten in terms of $\sigma_n$, yields the r.h.s. of (\ref{rec-xii}).

Next, observe that
$$
\xi_{2n}=\upsilon_{2n}+\upsilon_{2n}~ \varpi_{\theta}^{-1}~\upsilon_{2n}.
$$
Combining (\ref{def-w-1}) and (\ref{k-ricc-0}) with the l.h.s. formula in (\ref{ref-xi-odd})  we also have the factorisation
$$
 \gamma_{\theta}~\xi_{2n+1}~ \gamma_{\theta}^{\prime}= \gamma_{\theta}~\upsilon_{2n+1} ~\gamma_{\theta}^{\prime}\left(I+ \gamma_{\theta}~\upsilon_{2n+1} \gamma_{\theta}^{\prime}\right)=\left( \gamma_{\theta}~\upsilon_{2n+1}~ \gamma_{\theta}^{\prime}\right)~
\upsilon_{2(n+1)}^{-1}$$
The l.h.s. assertion in (\ref{rec-det}) now follows elementary arguments. The proof of the r.h.s. assertion in (\ref{rec-det}) follows exactly from the same argument.

Finaly, using (\ref{hom}) we obtain
$$
\left(\tau^{-1}_{2n}\beta_{2n}\right)~\tau_{2n+1}~\left(\beta_{2n}^{\prime}\tau_{2n}^{-1}\right)=
\tau^{-1}\beta~\left(\tau_{2n+1}~\beta^{\prime}~\tau^{-1}\right)=\tau^{-1}\beta~\beta_{2n+1}
$$
and combining the equation above with (\ref{hom-mi}) we arrive at
\begin{equation}\label{ref-sigma-0}
\sigma_{2n}^{-1}=\left(\beta_{2n}\sigma\beta_{2n}^{\prime}+\tau_{2n}\right)^{-1}=
\tau_{2n}^{-1}-\tau^{-1}\beta~\beta_{2n+1}.
\end{equation}
Then, using (\ref{hom}), we conclude that
$$
\sigma_{2n}^{-1}\beta_{2n}=\tau^{-1}\beta~\left(I-\beta_{2n+1}\beta_{2n}\right).
$$
In the same vein, from (\ref{hom}) we have
$$
\left(\tau_{2n+1}^{-1}\beta_{2n+1}\right) \tau_{2(n+1)}     \left( \beta_{2n+1}^{\prime}\tau_{2n+1}^{-1}\right)
=\beta^{\prime}\tau^{-1} \left(\tau_{2(n+1)} \tau^{-1}\beta\right)=\beta^{\prime}\tau^{-1}\beta_{2(n+1)}
$$
and combining the above with (\ref{hom-mii}) we find that
\begin{equation}\label{ref-sigma-01}
\sigma_{2n+1}^{-1}=
\tau_{2n+1}^{-1}-\beta^{\prime}\tau^{-1}\beta_{2(n+1)}.
\end{equation}
Using (\ref{hom}), we conclude that
$$
\sigma_{2n+1}^{-1}\beta_{2n+1}=
\beta^{\prime}\tau^{-1}~\left(I-\beta_{2(n+1)}\beta_{2n+1}\right).
$$
This ends the proof of (\ref{rec-sig}).
\cqfd

\subsection*{Proof of Proposition~\ref{prop-end}}\label{prop-end-proof}

Using (\ref{beta-s-2-ii})  and (\ref{rec-xii}) we check that
$$
\sigma_{2n}^{-1}=\tau^{-1}_{2n}-(\tau^{-1}_{2(n+1)}-\overline{\sigma}^{-1})=
\tau^{-1}_{2n}-\beta_{2n+1}^{\prime}\tau_{2n+1}^{-1}
\beta_{2n+1}
$$
and combining the above with (\ref{def-m-s}) we arrive at
$$
\left(m_{2p}-\overline{m}\right)^{\prime}\sigma_{2p}^{-1}\left(m_{2p}-\overline{m}\right)
=\left(m_{2p}-\overline{m}\right)^{\prime}\tau^{-1}_{2p}\left(m_{2p}-\overline{m}\right)-\left(m_{2p+1}-m\right)^{\prime}\tau_{2p+1}^{-1}
\left(m_{2p+1}-m\right).
$$
In the same vein, using (\ref{beta-s-2-i}) and (\ref{rec-xii}) we obtain
$$
\sigma_{2n+1}^{-1}=\tau_{2n+1}^{-1}-\left(\tau_{2n+3}^{-1}-\sigma^{-1}\right)=
\tau_{2n+1}^{-1}-\beta_{2(n+1)}^{\prime}\tau^{-1}_{2(n+1)}\beta_{2(n+1)}
$$
and, therefore,
\begin{eqnarray*}
\left(m_{2p+1}-m\right)^{\prime}\sigma_{2p+1}^{-1}
\left(m_{2p+1}-m\right)
&=&\left(m_{2p+1}-m\right)^{\prime}\tau_{2p+1}^{-1}\left(m_{2p+1}-m\right) \nonumber\\
&& -\left(m_{2(p+1)}-\overline{m}\right)^{\prime}\tau^{-1}_{2(p+1)}
\left(m_{2(p+1)}-\overline{m}\right).
\end{eqnarray*}
This implies that
\begin{eqnarray*}
2\left(V_{2n}(\overline{m})+U_{2n}(m)-U(m)\right) &=&
\log{\mbox{\rm det}(\upsilon_{2n}\upsilon^{-1}_{0})}
+\left(m_{2n}-\overline{m}\right)^{\prime}\tau^{-1}_{2n}
\left(m_{2n}-\overline{m}\right)\nonumber\\
&&-\left(m_{0}-\overline{m}\right)^{\prime}\tau^{-1}_{0}
\left(m_{0}-\overline{m}\right)
\end{eqnarray*}
and concludes the proof. \cqfd

\subsection*{Proof of Theorem~\ref{theo-end}}\label{theo-end-proof}

Combining  the decomposition
$$
\begin{array}{l}
\displaystyle
\left(y-m_{2p}\right)^{\prime}\sigma_{2p}^{-1}\left(y-m_{2p}\right)-\left(\overline{m}-m_{2p}\right)^{\prime}\sigma_{2p}^{-1}\left(\overline{m}-m_{2p}\right)\\
\\
\displaystyle=
2~\left(y-\overline{m}\right)^{\prime}\sigma_{2p}^{-1}\left(\overline{m}-m_{2p}\right)+
\left(y-\overline{m}\right)^{\prime}\sigma_{2p}^{-1}\left(y-\overline{m}\right)
\end{array}$$
with (\ref{rec-xii}) we obtain
$$
V_{2(n+1)}(y)-V_{2(n+1)}(\overline{m}) =\sum_{0\leq p\leq  n}
\left(y-\overline{m}\right)^{\prime}\sigma_{2p}^{-1}\left(m_{2p}-\overline{m}\right)+\frac{1}{2}
\left(y-\overline{m}\right)^{\prime}\left(\tau^{-1}_{2(n+1)}-\tau^{-1}_{0}\right)\left(y-\overline{m}\right).
$$
On the other hand, using (\ref{ref-m-intro}) and (\ref{ref-sigma-0}) for find that
$$
\sigma_{2p}^{-1}\left(m_{2p}-\overline{m}\right)=\sigma_{2p}^{-1}~\beta_{2p,0}~\left(m_{0}-\overline{m}\right)\quad\mbox{\rm and}\quad
\sigma^{-1}_0=\tau_{0}^{-1}-\tau^{-1}\beta~\beta_{1}
$$
for any $p \ge 1$, with the directed matrix product
$$
\beta_{q,0}:=\beta_{q}\beta_{q-1}\ldots \beta_{1}.
$$
Now, using (\ref{rec-sig}) we readily check that
$$
\sigma_{2p}^{-1}\beta_{2p,0}=\tau^{-1}\beta~\left(\beta_{2p-1,0}-\beta_{2p+1,0}\right)
$$
and this yields
\begin{eqnarray*}
V_{2(n+1)}(y)-V_{2(n+1)}(\overline{m})&=&
\left(y-\overline{m}\right)^{\prime}~\left(\tau_{0}^{-1}-\tau^{-1}\beta~\beta_{1}\right)~\left(m_{0}-\overline{m}\right)\\
&& +\frac{1}{2}
\left(y-\overline{m}\right)^{\prime}\left(\tau^{-1}_{2(n+1)}-\tau^{-1}_{0}\right)\left(y-\overline{m}\right)\\
&&+\left(y-\overline{m}\right)^{\prime}\tau^{-1}\beta~\left(\beta_{2n+1,0}-\beta_{1,0}\right)~\left(\overline{m}-m_0\right).
\end{eqnarray*}
We conclude that
\begin{eqnarray*}
V_{2(n+1)}(y)-V_{2(n+1)}(\overline{m})&=&
\left(y-\overline{m}\right)^{\prime}~\tau_{0}^{-1}~\left(m_{0}-\overline{m}\right) \\
&& +\frac{1}{2}\left(y-\overline{m}\right)^{\prime}\left(\tau^{-1}_{2(n+1)}-\tau^{-1}_{0}\right)\left(y-\overline{m}\right)\\
&& +\left(y-\overline{m}\right)^{\prime}\tau^{-1}\beta~\beta_{2n+1,0}~\left(\overline{m}-m_0\right).
\end{eqnarray*}
which completes the proof of the first assertion.

In the same vein, combining the decomposition
$$
\begin{array}{l}
\left(x-m_{2p+1}\right)^{\prime}\sigma_{2p+1}^{-1}
\left(x-m_{2p+1}\right)-\left(m-m_{2p+1}\right)^{\prime}\sigma_{2p+1}^{-1}
\left(m-m_{2p+1}\right)\\
\\
=2\left(x-m\right)^{\prime}\sigma_{2p+1}^{-1}
\left(m-m_{2p+1}\right)+\left(x-m\right)^{\prime}\sigma_{2p+1}^{-1}
\left(x-m\right)
\end{array}$$
with (\ref{rec-xii}) we arrive at
\begin{eqnarray*}
U_{2n}(x)-U_{2n}(m) &=& (U(x)-U(m)) \sum_{0\leq p< n}\left(x-m\right)^{\prime}\sigma_{2p+1}^{-1}
\left(m_{2p+1}-m\right)\\
&& +\frac{1}{2}~\left(x-m\right)^{\prime}\left(\tau_{2n+1}^{-1}-\tau_{1}^{-1}\right)
\left(x-m\right).
\end{eqnarray*}
On the other hand, using (\ref{ref-m-o-intro}) and (\ref{ref-sigma-01}), for any $p\geq 0$ we check that
$$
\sigma_{2p+1}^{-1}
\left(m_{2p+1}-m\right)=\sigma_{2p+1}^{-1}~\beta_{2p+1,1}~\left(m_{1}-m\right)\quad\mbox{\rm and}\quad 
\sigma_{1}^{-1}=\tau_{1}^{-1}-\beta^{\prime}\tau^{-1}\beta_{2}
$$
with the directed matrix product
$$
\beta_{q,1}:=\beta_{q}\beta_{q-1}\ldots \beta_{2}.
$$
Equivalently, in terms of the directed products (\ref{directed-prod}) we have
$$
\beta_{2n,1}=\beta^{\circ}_{2n,0}\quad \mbox{\rm and}\quad
\beta_{2n-1,0}=\beta^{\circ}_{2n-1,1}.
$$
Using (\ref{rec-sig}) we can write
$$
\sigma_{2p+1}^{-1}\beta_{2p+1,1}
=
\beta^{\prime}~\tau^{-1}~\left(
\beta_{2p,1}-\beta_{2(p+1),1}\right)
$$
which implies that
$$
\begin{array}{l}
U_{2n}(x)-U_{2n}(m)\\
\\
\displaystyle=(U(x)-U(m))+\left(x-m\right)^{\prime}\left(
\beta^{\prime}~\tau^{-1}\beta_2-\tau_1^{-1}\right)~\left(m-m_{1}\right)\\
\\
+\left(x-m\right)^{\prime}\left(\beta^{\prime}~\tau^{-1}~\left(
\beta_{2n,1}-\beta_{2,1}\right)\right)~\left(m-m_{1}\right)+\frac{1}{2}~\left(x-m\right)^{\prime}\left(\tau_{2n+1}^{-1}-\tau_{1}^{-1}\right)
\left(x-m\right).
\end{array}
$$
Finally, we conclude that
\begin{eqnarray*}
U_{2n}(x)-U_{2n}(m)&=&(U(x)-U(m))+\left(x-m\right)^{\prime}~\tau_1^{-1}~\left(m_{1}-m\right)\\
&&+\frac{1}{2}~\left(x-m\right)^{\prime}\left(\tau_{2n+1}^{-1}-\tau_{1}^{-1}\right)
\left(x-m\right) \\
&& +\left(x-m\right)^{\prime}~\beta^{\prime}~\tau^{-1}~
\beta_{2n,1}~\left(m-m_{1}\right).
\end{eqnarray*}
\cqfd

\subsection*{Proof of Corollary~\ref{prop-ct-pot}}\label{prop-ct-pot-proof}

The proof if based on the following technical lemma.

\begin{lem}\label{lem-ct-pot}
There exists some constant $c_{\theta}$ and some parameter $n_{\theta}$ such that for every $n\geq n_{\theta}$ we have
$$
\vert \epsilon^V_{2n}(\overline{m})\vert\leq c_{\theta}~\rho_{\theta}^{n}~\quad
\mbox{and}\quad
\vert \epsilon^U_{2n}(m)\vert\leq c_{\theta}~\overline{\rho}_{\theta_1}^{n}.
$$

\end{lem} 

\proof
Using Corollary~\ref{cor-beta-est} and Corollary~\ref{theo-cor-qs} 
we can find some  finite constant $c_{1,\theta}$ such that
$$
\Vert\sigma_{2n}^{-1/2}\left(m_{2n}-\overline{m}\right)\Vert\leq 
c_{1,\theta}~\rho_{\theta}^{n/2}
\quad \text{for every $n\geq 0$.}
$$
Next, we choose $n_0$ such that
$$
\Vert \sigma_{2n_0}-\overline{\sigma}\Vert\leq
c_{\theta}~\rho_{\theta}^{n_0}~\Vert \sigma_{0}-\overline{\sigma}\Vert\leq \frac{1}{2 \Vert \overline{\sigma}^{-1}\Vert_F}
$$
with the constant  $c_{\theta}$ as in Corollary~\ref{theo-cor-qs}. 
Then, by (\ref{rec-det}) and Lemma~\ref{lem-detlog} there exists some $c_{2,\theta}$ such that for any $n\geq n_0$ we have
$$
\left\vert\log{\mbox{\rm det}\left(\upsilon_{2n+1}\upsilon_{2n}^{-1}\right)}\right\vert 
\leq c_{2,\theta}~\rho_{\theta}^{n}~
\quad \text{and, therefore,}\quad
\vert \epsilon^V_{2n}(\overline{m})\vert\leq \frac{1}{2}~\frac{c_{1,\theta}+c_{2,\theta}}{1-\rho_{\theta}}~\rho_{\theta}^{n}.$$
This completes the proof of the first assertion. The proof of the second estimate follows exactly the same argument.\cqfd

Now we come to the proof of  Corollary~\ref{prop-ct-pot}.

\proof The estimates stated in Corollary~\ref{cor-beta-est} as well as in (\ref{def-rho-1})  and Theorem~\ref{theo-qs} imply that

$$
\Vert \beta^{\circ}_{2n,0}\Vert\leq c_{0,\theta}~\rho_{\theta}^{n/2}
\quad
\mbox{\rm and}\quad\Vert \beta^{\circ}_{2n-1,1}\Vert\leq c_{1,\theta}~\overline{\rho}_{\theta_1}^{n/2}
$$ 
as well as
$$
\Vert\tau^{-1}_{2n}-\varsigma_{\theta}^{-1}\Vert\leq c_{0,\theta}~\rho_{\theta}^n
\quad
\mbox{\rm and}\quad
\Vert\tau^{-1}_{2n+1}-\overline{\varsigma}_{\theta_1}^{-1}\Vert\leq c_{1,\theta}~\overline{\rho}_{\theta_1}^n
$$ 
for some constants $c_{0,\theta}$ and $c_{1,\theta}$.
Using Theorem~\ref{theo-end} we also check that
$$
\vert\epsilon^V_{2n}(y)\vert\leq 
c_{\theta}~\left(\rho_{\theta}^{n}+\overline{\rho}_{\theta_1}^{n/2}~\Vert y-\overline{m}\Vert+\rho_{\theta}^n~\Vert y-\overline{m}\Vert^2\right)
$$
and
$$
\vert\epsilon^U_{2n}(x)\vert\leq c_{\theta}~\left(\overline{\rho}_{\theta_1}^{n}+\rho_{\theta}^{n/2}~\Vert x-m\Vert
+\overline{\rho}_{\theta_1}^n~\Vert x-m\Vert^2
\right).
$$
\cqfd

\subsection*{Proof of Corollary~\ref{end-cor}}\label{end-cor-proof}

The optimal bridge $\SS(\theta)$ yields
\begin{eqnarray*}
P_{\SS(\theta)}(d(x,y))&=&
~\nu_{m,\sigma}(dx)~K_{\SS(\theta)}(x,dy)= e^{-\UU_{\theta}(x)}~ q_{\theta}(x,y)~e^{-\VV_{\theta}(y)}~dxdy
\end{eqnarray*}
and
$$
Q_{\theta}(x,dy):=q_{\theta}(x,y)dy\Longrightarrow
K_{\SS(\theta)}(x,dy)= \frac{Q_{\theta}(x,dy)~e^{-\VV_{\theta}(y)}}{Q_{\theta}(e^{-\VV_{\theta}})(x)}.
$$
On the other hand, by (\ref{comm-intro}) we have $\BB_{m,\sigma}\left(\SS(\theta_0)\right)=\overline{\SS}(\theta_1)$ with $\theta_1=\BB_{m,\sigma}(\theta_0)$. Recalling that $\nu_{m,\sigma}K_{\SS(\theta)}=\nu_{\overline{m},\overline{\sigma}}$,
the conjugate formula (\ref{ref-conjug}) implies that
\begin{eqnarray*}
P_{\SS(\theta_0)}&=&\overline{P}^{\flat}_{\overline{\SS}(\theta_1)}.
\end{eqnarray*}
Equivalently, we have
$$
\overline{P}_{\overline{\SS}(\theta_1)}(d(x,y))=\nu_{\overline{m},\overline{\sigma}}(dx)~K_{\overline{\SS}(\theta_1)}(x,dy)
=e^{-\VV_{\theta_0}(x)}~ r_{\theta_0}(x,y)~e^{-\UU_{\theta_0}(y)}~~dxdy
$$
and
$$
K_{\overline{\SS}(\theta_1)}(x,dy)=\frac{R_{\theta_0}(x,dy)~e^{-\UU_{\theta_0}(y)}}{R_{\theta_0}(e^{-\UU_{\theta_0}})(x)}.
$$
This completes the proof of the corollary.\cqfd

\section{Some technical proofs}\label{sec-tech-proofs}

\subsection*{Proof of (\ref{sym-sharp})}\label{sym-sharp-proof}

The symmetric property comes from the fact that
\begin{eqnarray}
u^{1/2}~ \left(u^{-1/2}~v~u^{-1/2}\right)^{1/2}~u^{1/2}&=&
u^{1/2}~ \left(u^{1/2}~v^{-1}~u^{1/2}\right)^{-1/2}~u^{1/2}\nonumber\\
&=&v^{1/2}~ \left(v^{-1/2}~u~v^{-1/2}\right)^{1/2}~v^{1/2}\label{ref-half}
\end{eqnarray}
and the last assertion comes from the fact that
\begin{eqnarray*}
\left(v^{-1/2}u^{1/2}~ \left(u^{1/2}~v^{-1}~u^{1/2}\right)^{-1/2}~u^{1/2}v^{-1/2}\right)^2
&=&v^{-1/2}u^{1/2}~u^{1/2}v^{-1/2}=v^{-1/2}~u~v^{-1/2}.
\end{eqnarray*}
\cqfd

\subsection*{Proof of Theorem~\ref{theo-2-S-o-S}}\label{theo-2-S-o-S-proof}
By Theorem~\ref{Th1} we have
$$
\overline{\SS}(\theta_1):=(\overline{\iota}_{\theta_1},\overline{\kappa}_{\theta_1},\overline{\varsigma}_{\theta_1})
$$
with the parameters $(\overline{\iota}_{\theta_1},\overline{\kappa}_{\theta_1},\overline{\varsigma}_{\theta_1})
$ defined in (\ref{def-over-Sa}).
By (\ref{inter-beta-01}) we have
\begin{eqnarray*}
\theta_1:=\BB_{m,\sigma}(\theta)\quad \mbox{\rm and}\quad
\theta=(\alpha,\beta,\tau)&\Longrightarrow&
\cchi_{\theta_1}=\tau_1^{-1}\beta_1=\beta^{\prime}\tau^{-1}=\cchi_{\theta}^{\prime}\\
&\Longrightarrow& \overline{\gamma}_{\theta_1}=\gamma_{\theta}^{\prime}\quad \mbox{\rm and}\quad \overline{\kappa}_{\theta_1}:=\overline{\varsigma}_{\theta_1}~\beta^{\prime}\tau^{-1}.
\end{eqnarray*}
We recall from (\ref{def-cchi}) (see also Appendix~\ref{stat-Ricc-sec} on page~\pageref{stat-Ricc-sec}) that
$$
\overline{\varsigma}_{\theta_1}^{-1}=\sigma^{-1}+ \cchi_{\theta}^{\prime}~
\varsigma_{\theta}~\cchi_{\theta}=\sigma^{-1}+\kappa_{\theta}^{\prime}~\varsigma_{\theta}^{-1}\kappa_{\theta}
$$
with the matrices $(\kappa_{\theta},\varsigma_{\theta})$ defined in (\ref{def-Sa}).
In the reverse direction,  consider the initial parameter associated with the bridge parameters (\ref{def-Sa}), that is
$$
\theta_0=(\alpha_0,\beta_0,\tau_0)=(\iota_{\theta},\kappa_{\theta},\varsigma_{\theta})\Longrightarrow m_0=\overline{m}\quad\mbox{\rm and}\quad
\sigma_0=\overline{\sigma}\quad \mbox{\rm by  (\ref{ref-fix-point-intro})}.
$$
In this situation, applying the Bayes' map (\ref{ref-conjug-Z-par}) we have
$$
\BB_{m,\sigma}(\theta_0)=\theta_1=(\alpha_1,\beta_1,\tau_1)
$$
with the parameters $(\alpha_1,\beta_1,\tau_1)$ defined below. Using (\ref{inter-beta-01}) and recalling that $\sigma_0=\overline{\sigma}$ we also have
\begin{eqnarray*}
\tau_1^{-1}\beta_1&=&\beta_0^{\prime}~\tau_0^{-1}=\kappa_{\theta}^{\prime}~\varsigma_{\theta}^{-1}
\\
\beta_1&=&\sigma~\kappa_{\theta}^{\prime}~\overline{\sigma}^{-1}=
\sigma~\cchi_{\theta}^{\prime}~\varsigma_{\theta}~~\overline{\sigma}^{-1}=
\overline{\varsigma}_{\theta_1}~\cchi_{\theta}^{\prime}=\overline{\varsigma}_{\theta_1}~\beta^{\prime}\tau^{-1}=\overline{\kappa}_{\theta_1}.
\end{eqnarray*}
The last assertion comes from the fact that
\begin{eqnarray*}
\overline{\varsigma}_{\theta_1}^{-1}~\beta_1&=&\left(\sigma^{-1}+ \cchi_{\theta}^{\prime}~
\varsigma_{\theta}~\cchi_{\theta}\right)\sigma~\cchi_{\theta}^{\prime}~\varsigma_{\theta}~~\overline{\sigma}^{-1}\\
&=&\cchi_{\theta}^{\prime}~\left(\varsigma_{\theta}+ 
\varsigma_{\theta}~\left(\cchi_{\theta}~\sigma~\cchi_{\theta}^{\prime}\right)~\varsigma_{\theta}\right)~\overline{\sigma}^{-1}=\cchi_{\theta}^{\prime}\quad \mbox{\rm by (\ref{ref-fix-point-intro})}.
\end{eqnarray*}
Using (\ref{inter-tau-1-i}), finally note that
$$
 \tau_1^{-1}=\sigma^{-1}+\kappa_{\theta}^{\prime}~\varsigma_{\theta}^{-1}~\kappa_{\theta}=
 \sigma^{-1}+\cchi_{\theta}^{\prime}~
 \varsigma_{\theta}~\cchi_{\theta}=\overline{\varsigma}_{\theta_1}^{-1}
 \quad \mbox{\rm and}\quad
 \alpha_1=\overline{\iota}_{\theta_1}:=m-\overline{\kappa}_{\theta_1}~\overline{m}.
$$
We conclude that
$$
\BB_{m,\sigma}(\SS(\theta))=\BB_{m,\sigma}(\iota_{\theta},\kappa_{\theta},\varsigma_{\theta})=\left(\overline{\iota}_{\theta_1},\overline{\kappa}_{\theta_1},\overline{\varsigma}_{\theta_1}\right)=\overline{\SS}(\BB_{m,\sigma}(\theta)).
$$
This ends the proof of the theorem.\cqfd

\subsection*{Proof of Corollary~\ref{cor-monge-maps-ind}}\label{cor-monge-maps-ind-proof}

Using (\ref{rmk-theta-t-1}) we have
$$
  r_{\theta(t)}-I=\left(  \varpi_{\theta(t)}+\left(\frac{  \varpi_{\theta(t)}}{2}\right)^2\right)^{1/2}-\left(I+\frac{  \varpi_{\theta(t)}}{2}\right)
$$
and, on the other hand,
$$
  \varpi_{\theta(t)}+\left(\frac{  \varpi_{\theta(t)}}{2}\right)^2-\left(I+\frac{  \varpi_{\theta(t)}}{2}\right)^2=-I
$$
as well as
$$
  \varpi_{\theta(t)}+\left(\frac{  \varpi_{\theta(t)}}{2}\right)^2\succeq \frac{  \varpi_{\theta(t)}}{2}\succeq \frac{t^2}{2}~\varpi_{I}.
$$
Now, using the Ando-Hemmen inequality we find that
$$
\begin{array}{l}
\displaystyle \Vert r_{\theta(t)}-I\Vert_2\leq \frac{1}{t}~\frac{\sqrt{2}}{\lambda^{1/2}_{min}(\omega)}
\Longrightarrow
\Vert\varsigma_{\theta(t)}-\overline{\sigma}\Vert_2=\Vert \overline{\sigma}^{1/2}~(  r_{\theta(t)}-I)~\overline{\sigma}^{1/2}\Vert_2\leq \frac{1}{t}~\frac{\sqrt{2}~\Vert \overline{\sigma}\Vert_2}{\lambda^{1/2}_{min}(\omega)}.
\end{array}
$$
Finally, note that
$$
\overline{m}-
\iota_{\theta(t)}=\kappa_{\theta(t)}~m\quad\mbox{\rm and}\quad
 \kappa_{\theta(t)}=t^{-1}\varsigma_{\theta(t)}~\beta
$$
which implies the estimate
$$
\Vert
\iota_{\theta(t)}-\overline{m}\Vert\vee\Vert r_{\theta(t)}-I\Vert\vee \Vert\varsigma_{\theta(t)}-\overline{\sigma}\Vert\vee \Vert \kappa_{\theta(t)}\Vert\leq c/t.
$$
This ends the proof of (\ref{monge-maps-ind}).

Using (\ref{fix-intro}) we check that
\begin{eqnarray*}
\overline{\iota}_{\theta_1(t)}&=&m-\overline{\kappa}_{\theta_1(t)}~\overline{m},\\
\overline{\kappa}_{\theta_1(t)}&=&\sigma~\kappa_{\theta(t)}^{\prime}~\overline{\sigma}^{-1}\quad \mbox{\rm and}\quad
\overline{\varsigma}_{\theta_1(t)}^{-1}=\sigma^{-1}+\frac{1}{t^2}~\beta^{\prime}~\varsigma_{\theta(t)}~\beta.
\end{eqnarray*}
Recalling that the Sinkhorn iteration is initialized at $\theta_0(t)=\theta(t)=(\alpha,\beta,tI)$ we see that
$$
\tau_{1}(t)^{-1}=\sigma^{-1}+\frac{1}{t}~ \beta^{\prime}\beta
\Longrightarrow \Vert \overline{\kappa}_{\theta_1(t)}\Vert\vee \Vert \overline{\varsigma}_{\theta_1(t)}^{-1}-\tau_{1}(t)^{-1} \Vert\leq c/t.
$$
In addition, using the decomposition
$$
\overline{\varsigma}_{\theta_1(t)}-\tau_{1}(t)=\overline{\varsigma}_{\theta_1(t)}\left(\tau_{1}(t)^{-1}-\overline{\varsigma}_{\theta_1(t)}^{-1}\right)\tau_{1}(t)
$$
and recalling that
$$
\tau_{1}(t)\preceq \sigma\quad\mbox{\rm and}\quad
\overline{\varsigma}_{\theta_1(t)}\preceq \sigma
$$
we arrive at
$$
\Vert \overline{\varsigma}_{\theta_1(t)}-\tau_{1}(t) \Vert\leq c/t.
$$
Using (\ref{def-m-s}) and (\ref{hom}) we also obtain
$$
m_{1}(t)-m=\frac{1}{t}~
\tau_{1}(t)~\beta^{\prime}~ (\overline{m}-m_{0})\Longrightarrow\Vert m_{1}(t)-m
\Vert\leq c/t
$$
as well as
$$
m-\overline{\iota}_{\theta_1(t)}=\overline{\kappa}_{\theta_1(t)}~\overline{m}
\Longrightarrow\Vert \overline{\iota}_{\theta_1(t)}-m
\Vert\leq c/t.
$$
We complete the proof of (\ref{monge-maps-ind-2}) arguing as in the proof of (\ref{monge-maps-ind}).
\cqfd

\subsection*{Proof of Proposition~\ref{est-ct-m-intro}}\label{est-ct-m-intro-proof}
Rewriting (\ref{connect-fp-v2}) in a sightly different form, we obtain
$$
\overline{\sigma}^{1/2}\displaystyle\varsigma_{\theta(t)}^{-1}~\overline{\sigma}^{1/2}-I=\frac{1}{t^2}~\overline{\sigma}^{1/2}\beta~\sigma^{1/2}~\overline{r}_{\theta_1(t)}~\sigma^{1/2}~\beta^{\prime}~\overline{\sigma}^{1/2}
$$
and using (\ref{monge-maps-ind-2}) we see that
$$
\Vert I-\overline{\sigma}^{1/2}\displaystyle\varsigma_{\theta(t)}^{-1}\overline{\sigma}^{1/2}\Vert\leq  {c_0}/{t^2}.
$$
By (\ref{lem-tech-2}) there is also some $t_0$ such that for any $t\geq t_0$ we have 
$$
\left\vert\log{\mbox{\rm det}\left(\displaystyle\varsigma_{\theta(t)}^{-1}\overline{\sigma}\right)}\right\vert 
\leq {c_0}/{t^2}.
$$
This ends the proof of the first assertion. 

On the other hand, from (\ref{inter-tau-1-i}) and (\ref{connect-fp-v2})  we also have
$$
\begin{array}{l}
\displaystyle\tau_{1}(t)^{-1}=\sigma^{-1}+\frac{1}{t}~ \beta^{\prime}\beta
\quad \mbox{\rm and}\quad
\overline{\varsigma}_{\theta_1(t)}^{-1}=\sigma^{-1}+ \frac{1}{t}~ \beta^{\prime}~
\overline{\sigma}^{1/2}~\frac{r_{\theta(t)}}{t}~\overline{\sigma}^{1/2}~\beta.
\end{array}$$
In this context, formula (\ref{alter-UU}) takes the form
$$
\UU_{\theta(t)}(x+m)-\UU_{\theta(t)}(m)-2^{-1}x^{\prime}~\sigma^{-1}~x =
\frac{1}{t}~(\beta x)^{\prime}~(\overline{m}-m_0)
+\frac{1}{2t}~(\beta x)^{\prime}
\left(\overline{\sigma}^{1/2}~\frac{r_{\theta(t)}}{t}~\overline{\sigma}^{1/2}-I\right)(\beta x).
$$
The second assertion is now a direct consequence of (\ref{monge-maps-ind}).
Finally, using (\ref{connect-fp}) we have
$$
\begin{array}{l}
\displaystyle\varsigma_{\theta(t)}^{-1}=\overline{\sigma}^{-1}+\frac{1}{t}~\beta~\sigma^{1/2}~\frac{\overline{r}_{\theta_1(t)}}{t}~\sigma^{1/2}~\beta^{\prime}.
\end{array}
$$
Therefore, by Theorem~\ref{theo-s-pot} we obtain
\begin{eqnarray*}
\VV_{\theta(t)}(y+\overline{m})-\VV_{\theta(t)}(\overline{m})-2^{-1}y^{\prime}~\overline{\sigma}^{-1}~y
&=& \frac{1}{t}~
y^{\prime}~\left(m_{0}-\overline{m}\right) \\
&& +\frac{1}{2t}~
y^{\prime}~\left(\left(\beta~\sigma^{1/2}~\frac{\overline{r}_{\theta_1(t)}}{t}~\sigma^{1/2}~\beta^{\prime}-I\right)\right)~y
\end{eqnarray*}
The last assertion is now a direct consequence of (\ref{monge-maps-ind-2}).
\cqfd

\subsection*{Proof of (\ref{c-eta}) and (\ref{c-mu})}\label{c-mu-proofs}

We readily check (\ref{c-eta}) using the decomposition
$$
c(x,y)=\frac{1}{2}\left\Vert \tau^{-1/2}\left[(y-(\alpha+\beta m))-\beta (x-m)\right]\right\Vert_F^2+\frac{1}{2}\log{\mbox{\rm det}(2\pi \tau)}.
$$
In the same vein, we check (\ref{c-mu}) using the decomposition
 $$
c(x,y)=\frac{1}{2}\left\Vert \tau^{-1/2}\left[(y-\overline{m})+(\overline{m}-(\alpha+\beta x))\right]\right\Vert_F^2+\frac{1}{2}\log{\mbox{\rm det}(2\pi \tau)}.
$$
\cqfd

\subsection*{Proof of (\ref{c-eta-2}) and (\ref{c-mu-2})}\label{c-mu-2-proofs}
Returning to the Example~\ref{examp-gauss-sc} and using (\ref{c-eta}) we have
\begin{eqnarray*}
c_{\eta}(y)-c(x,y)&=&
\frac{1}{2}\left\Vert \tau^{-1/2}\beta (x-m)\right\Vert_F^2+
(x-m)^{\prime}\beta^{\prime}\tau^{-1}(y-(\alpha+\beta m))+\frac{1}{2}\tr(\beta^{\prime}\tau^{-1}\beta \sigma)\\
&=&-\frac{1}{2}\left\Vert \tau^{-1/2}\beta (x-m)\right\Vert_F^2+
(x-m)^{\prime}\beta^{\prime}\tau^{-1}(\overline{m}-(\alpha+\beta m))+\frac{1}{2}\tr(\beta^{\prime}\tau^{-1}\beta \sigma)\\
&&+(x-m)^{\prime}\beta^{\prime}\tau^{-1}(y-\overline{m}),
\end{eqnarray*}
which yields the formula
\begin{eqnarray*}
\log{\int~\mu(dy)~e^{c_{\eta}(y)-c(x,y)}}&=&-\frac{1}{2}\left\Vert \tau^{-1/2}\beta (x-m)\right\Vert_F^2+
(x-m)^{\prime}\beta^{\prime}\tau^{-1}(\overline{m}-(\alpha+\beta m)) \\
&& +\frac{1}{2}\tr(\beta^{\prime}\tau^{-1}\beta \sigma) + \frac{1}{2}~(x-m)^{\prime}\beta^{\prime}\tau^{-1} \overline{\sigma}~\tau^{-1}~\beta~(x-m).
\end{eqnarray*}
This completes the proof of (\ref{c-eta-2}). 

Using (\ref{c-mu}) we also have
\begin{eqnarray*}
c^{\mu}(x)-c(x,y)&=&\frac{1}{2}\left\Vert \tau^{-1/2}(\overline{m}-(\alpha+\beta x))\right\Vert_F^2
-\frac{1}{2}\left\Vert \tau^{-1/2}\left[(y-\overline{m})+(\overline{m}-(\alpha+\beta x))\right]\right\Vert_F^2 \\
&& +\frac{1}{2}\tr(\tau^{-1}\overline{\sigma})\\
&=& -\frac{1}{2}\left\Vert \tau^{-1/2}~(y-\overline{m})\right\Vert_F^2-(y-\overline{m})^{\prime}\tau^{-1}(\overline{m}-(\alpha+\beta m ))
+\frac{1}{2}\tr(\tau^{-1}\overline{\sigma})\\
&&+(y-\overline{m})^{\prime}\tau^{-1}\beta(x-m)
\end{eqnarray*}
that yields
\begin{eqnarray*}
\log{\int~\eta(dx)~e^{c^{\mu}(x)-c(x,y)}}&=& -\frac{1}{2}\left\Vert \tau^{-1/2}~(y-\overline{m})\right\Vert_F^2-(y-\overline{m})^{\prime}\tau^{-1}(\overline{m}-(\alpha+\beta m )) \\
&&+\frac{1}{2}\tr(\tau^{-1}\overline{\sigma}) +\frac{1}{2}
(y-\overline{m})^{\prime}\tau^{-1}\beta\sigma \beta^{\prime}\tau^{-1}
(y-\overline{m}).
\end{eqnarray*}
This completes the proof of (\ref{c-mu-2}).\cqfd

\bibliography{full}
\bibliographystyle{plain}

\end{document}